\documentclass[10pt,twocolumn,letterpaper]{article}

\usepackage[pagenumbers]{cvpr} 

\usepackage[dvipsnames]{xcolor}
\usepackage{graphicx}
\usepackage{amsmath}
\usepackage{amssymb}
\usepackage{booktabs}

\usepackage{multirow}
\usepackage{pifont}

\newcommand{\cmark}{\ding{51}}%
\newcommand{\xmark}{\ding{55}}%
\definecolor{cvprblue}{rgb}{0.21,0.49,0.74}
\usepackage[pagebackref,breaklinks,colorlinks,citecolor=cvprblue]{hyperref}

\usepackage[capitalize]{cleveref}
\crefname{section}{Sec.}{Secs.}
\Crefname{section}{Section}{Sections}
\Crefname{table}{Table}{Tables}
\crefname{table}{Tab.}{Tabs.}

\title{Uni3DL: Unified Model for 3D and Language Understanding}

\author{Xiang Li$^{1,*}$, Jian Ding$^{1,}$\thanks{Equal contribution}, Zhaoyang Chen$^{2}$\thanks{This work was done when Zhaoyang Chen was an intern at KAUST.}, Mohamed Elhoseiny$^{1}$\\
$^{1}$ King Abdullah University of Science and Technology\\
$^{2}$ Ecole Polytechnique\\
{\tt\small \{xiang.li.1,jian.ding,zhaoyang.chen,mohamed.elhoseiny\}@kaust.edu.sa}
}

\begin{document}

\makeatletter
\let\@oldmaketitle\@maketitle
\renewcommand{\@maketitle}{\@oldmaketitle
\myfigure\bigskip}

\makeatother
\newcommand\myfigure{%
  \makebox[0pt]
  {\hspace{17cm}\includegraphics[width=0.95\linewidth]{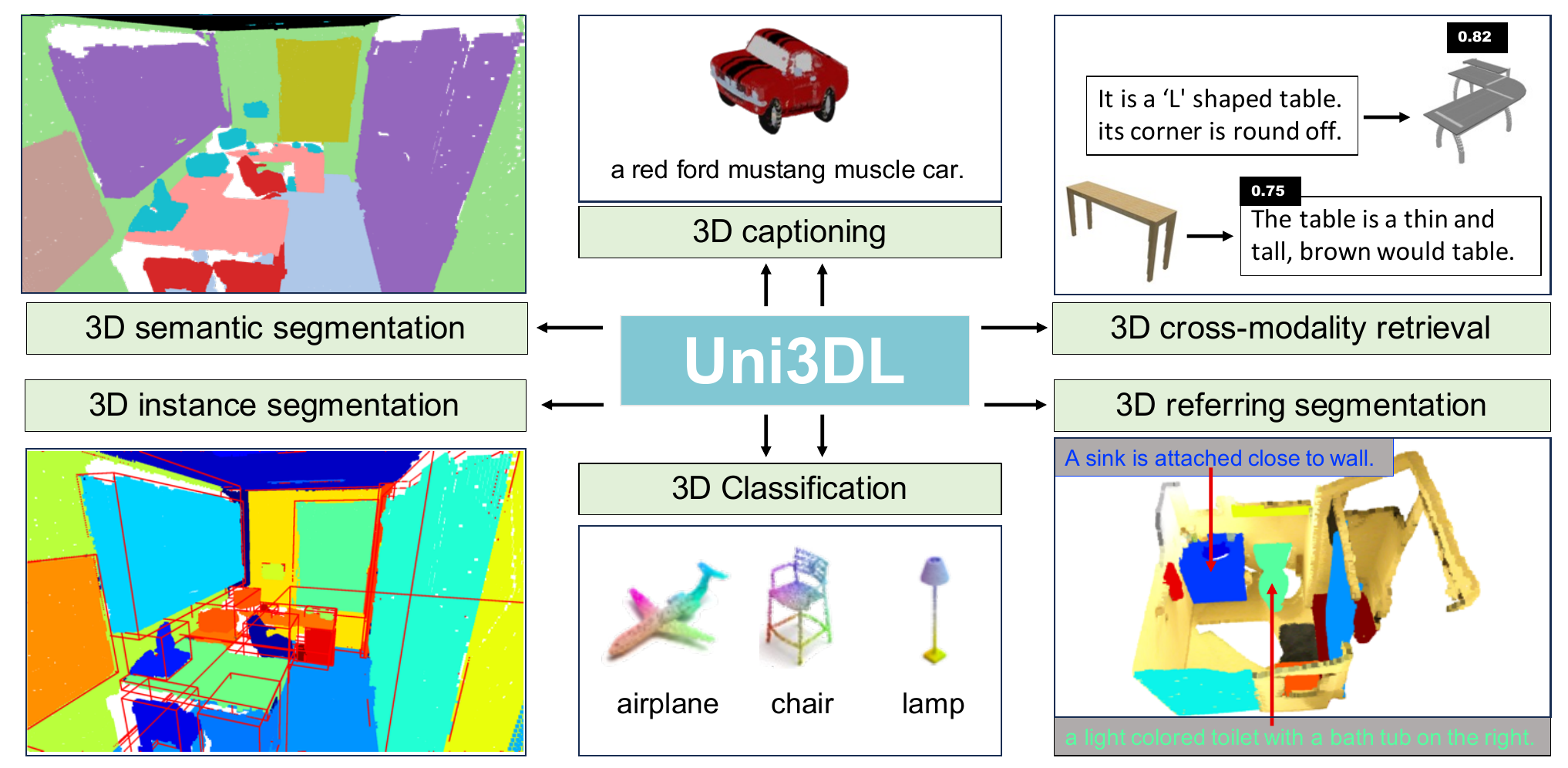}}
  \\
  \refstepcounter{figure}\normalfont{Figure~\thefigure: With a unified architecture, Uni3DL supports diverse  3D vision-language understanding tasks, including semantic segmentation, object detection, instance segmentation, grounded segmentation, captioning, text-3D cross-modality retrieval, (zero-shot) 3D object classification.}
  \label{fig:teaser}
}

\maketitle

\begin{abstract}
In this work, we present Uni3DL, a unified model for 3D and Language understanding. Distinct from existing unified vision-language models in 3D which are limited in task variety and predominantly dependent on projected multi-view images, Uni3DL operates directly on point clouds. This approach significantly expands the range of supported tasks in 3D, encompassing both vision and vision-language tasks in 3D. At the core of Uni3DL, a query transformer is designed to learn task-agnostic semantic and mask outputs by attending to 3D visual features, and a task router is employed to selectively generate task-specific outputs required for diverse tasks. With a unified architecture, our Uni3DL model enjoys seamless task decomposition and substantial parameter sharing across tasks. Uni3DL has been rigorously evaluated across diverse 3D vision-language understanding tasks, including semantic segmentation, object detection, instance segmentation, visual grounding, 3D captioning, and text-3D cross-modal retrieval. It demonstrates performance on par with or surpassing state-of-the-art (SOTA) task-specific models. We hope our benchmark and Uni3DL model will serve as a solid step to ease future research in unified models in the realm of 3D and language understanding. Project page: \url{https://uni3dl.github.io/}.
\end{abstract}

\section{Introduction}
\label{sec:intro}
3D perception technology stands as a fundamental element in the automatic understanding and operation within the physical world. It enhances various applications, including robotic navigation, object manipulation, autonomous driving, and virtual reality, thus improving machine-world interactions. 3D perception encompasses a broad spectrum of vision and vision-language tasks, such as 3D instance segmentation~\cite{hou20193d,yang2019learning,liu2020learning,yi2019gspn,jiang2020pointgroup,han2020occuseg,9710900,Liang_2021_ICCV,sun2023superpoint}, semantic segmentation~\cite{pointnet,qi2017pointnet++,qian2022pointnext,park2022fast,lai2022stratified,wu2022point,yang2023swin3d,li2020dancenet}, visual grounding~\cite{huang2021text,zhao20213dvg,cai20223djcg,bakr2022look}, object detection~\cite{yang2022unified,Lai_2023_ICCV}, retrieval~\cite{chen2019text2shape,tang2023parts2words} and captioning~\cite{cap3d,xu2023pointllm}, this technology has witnessed remarkable advancements. 

Despite these successes, task-specific models in 3D perception often lack generalizability, constraining their effectiveness across varied tasks. In contrast, the broader scientific community, as exemplified by the grand unified theory (GUT) in physics~\cite{baez2010algebra,langacker1981grand}, has consistently emphasized the importance of unification. Similarly, there is a growing trend towards unified models that integrate vision and language tasks, a concept that has demonstrated significant success in 2D domains~\cite{clip,xdecoder,uni-perceiverv2,yuan2021florence,alayrac2022flamingo,wang2022ofa,li2023rsclip}. For example, CLIP~\cite{clip} employs vision-language contrastive learning for zero-shot transfer across different classification tasks. Mask2former~\cite{maskformer,mask2former} leverages a transformer-based architecture for unifying generic segmentation tasks. Moreover, XDecoder~\cite{xdecoder} and Uni-Perceiver v2~\cite{uni-perceiverv2} adopt \textit{functionality unification modeling}~\cite{li2023multimodal}, covering both vision-only and vision-language tasks. These unified models exhibit greater versatility, efficient data utilization, and adaptability compared to task-specific models, resulting in heightened efficiency and conservation of resources during development.

Extending these successes of unified vision-language modeling for 2D tasks~\cite{xdecoder,uni-perceiverv2,clip,yuan2021florence} to 3D tasks remains a formidable challenge. This difficulty primarily stems from the substantial architectural differences between 2D and 3D models, along with the limited availability of extensive 3D datasets for pre-training purposes.
Several studies, including ~\cite{zhang2022pointclip,zhu2023pointclip,ULIP-2}, have explored adapting CLIP, originally pre-trained with 2D images, for 3D vision-language modeling. They achieve this by using projected multi-view images of point clouds in either training or testing phases. However, these methods have been mainly developed for classification tasks. 3D-VisTA~\cite{3D-vista} constructed large-scale 3D scene-text pairs dataset, and performed a vision-language pre-training for 3D without the need for 2D pre-trained models. However, it still requires fine-tuning for specific tasks, and the parameters of each task head are totally different.

Current unified vision-language models in 3D are summarized in Table~\ref{table:vlm3d_comparison}, the scope of tasks supported by current 3D vision-language models is comparatively limited, with dense prediction tasks such as semantic and instance segmentation receiving less attention. Furthermore, most existing models necessitate the use of multi-view images rather than direct training on 3D point clouds. This approach, while performing well, often results in the loss of critical information (e.g., 3D geometry) and leads to increased model complexity and overhead (multiple projected views required). 

In response to these improvement opportunities, we introduce a unified model for 3D perception that leverages both point clouds and language. Uni3DL designs a query transformer that enables latent and textural queries to softly attend to 3D visual features. The integration of these elements allows for the processing of point features, textual queries, and latent queries, and generates semantic features and mask features. Then we designed a task router with multiple \textit{functional} heads, which takes semantic features or mask features to predict diverse outputs. By combining the outputs from these functional heads, we are able to obtain the final results for various tasks.

Our contributions are summarized as:
\begin{itemize}
    \item We present Uni3DL, a unified model tailored for 3D vision and language comprehension. Its versatile architecture allows for the processing of both point clouds and textual inputs, generating diverse outputs including masks, classes, and texts. The model can be directly applied to dense prediction tasks (e.g., instance segmentation), even without task-specific finetuning.
    \item With a carefully designed query transformer decoder and task router, our model supports a wide range of vision-only and vision-language tasks within a single, unified architecture, and enjoys seamless task decomposition and substantial parameter sharing across tasks.
    \item Our results demonstrate enhanced or comparable performance against other multi-task and specialized models across a range of 3D vision-only and vision-language tasks.
\end{itemize}

\begin{table*}[htbp]
\resizebox{0.99\textwidth}{!}{
\begin{tabular}{cccccccccccc}
\toprule
Methods           & MV   & Pretrained FM & Sem Seg & Inst Seg & Gnd Seg & Gnd Loc & Class & Retr & Det  & Capt \\
\midrule
PointCLIP v2~\cite{zhu2023pointclip}      & \checkmark & CLIP~\cite{clip}, GPT-3~\cite{gpt-3}    &    \checkmark   &          &         &         & \checkmark  & \checkmark & \(\scalebox{1.6}{$\circ$}\) &       \\
UniT3D~\cite{chen2023unit3d}            & \checkmark & BERT~\cite{devlin2018bert}           &         &          &         & \checkmark    &       &      &      &     \checkmark      \\
3DJCG~\cite{cai20223djcg}           &  &  Glove~\cite{pennington2014glove}          &         &          &         & \checkmark    &       &      &      &   \checkmark    \\
ULIP~\cite{xue2023ulip}              & \checkmark & CLIP~\cite{clip}           &         &          &         &         & \checkmark  & \checkmark &      &          \\
ULIP-2~\cite{ULIP-2}            & \checkmark & CLIP~\cite{clip}           &         &          &         &         & \checkmark  & \checkmark &      &           \\
3D-VisTA~\cite{3D-vista}          &      & GPT-3~\cite{gpt-3}          &         &          &         & \(\scalebox{1.6}{$\circ$}\)    &       &      &      &    \(\scalebox{1.6}{$\circ$}\)       \\
Point-LLM~\cite{xu2023pointllm}         &      & ULIP-2~\cite{ULIP-2}, Vicuna~\cite{chiang2023vicuna}         &         &          &         &         & \checkmark  &      &      & \checkmark &     \\
Point-Bind~\cite{guo2023point}        & \checkmark & OpenCLIP~\cite{openclip}       &         &          &         &         & \checkmark  & \checkmark &      &      &       \\ 
\hline
Uni3DL (Ours) &      &                & \checkmark    & \checkmark     & \checkmark    & \checkmark    & \checkmark  & \checkmark & \checkmark & \checkmark &    \\
\bottomrule
\end{tabular}} 
\caption{Comparison of various vision-language models in 3D, highlighting their capabilities across a multitude of tasks. It specifically indicates the utilization of Multi-View (MV) images and delineates the types of Pretrained Foundation Models (FMs) employed. $\scalebox{1.6}{$\circ$}$ denotes the method is capable of doing the task but requires additional task-specific modules. The abbreviations employed in this comparison are as follows: Sem Seg for Semantic Segmentation, Inst Seg for Instance Segmentation, Gnd Seg for Grounded Segmentation, Gnd Loc for Grounded Localization, Class for Classification, Retr for Retrieval, Det for Detection, Capt for Captioning.}
\label{table:vlm3d_comparison}
\vspace{-10pt}
\end{table*}

\section{Related Work}
\label{sec:formatting}

\subsection{Unified Vision-Language Models in 2D}
The pursuit of unified architectures with shared parameters across tasks is a key goal in computer vision and machine learning. Models like CLIP~\cite{clip} and ALIGN~\cite{align} have made significant progress in merging vision and language through contrastive pre-training on extensive web-sourced image-text pairs, enabling natural language-based zero-shot transfer for various tasks. Yet, their use has predominantly been confined to classification, indicating room for broader application.

To broaden the scope, modeling in this domain has branched into two main categories: \textit{I/O unification} and \textit{functional unification}~\cite{li2023multimodal}. Inspired by sequence-to-sequence (seq2seq) modeling in NLP~\cite{radford2018improving}, I/O unification employs decoders to generate homogenous token sequences that are further processed by task-specific decoders. Prominent models such as Flamingo~\cite{alayrac2022flamingo}, OFA~\cite{wang2022ofa}, and GIT~\cite{GIT} have primarily concentrated on image-level tasks like image captioning and visual question answering (VQA). Models like Pix2Seq v2~\cite{chen2022unified}, Unitab~\cite{yang2022unitab}, and Unified-IO have extended this approach by incorporating discrete coordinate tokens in seq2seq modeling for localization tasks, with Vision-LLM~\cite{visionllm} and MiniGPT-2~\cite{chen2023minigpt} enhancing this capability using pre-trained large language models. In contrast, \textit{functional unification} models, exemplified by X-Decoder~\cite{xdecoder} and Uni-Perceiver v2~\cite{uni-perceiverv2}, predict heterogeneous outputs and utilize various routers or headers to deliver final outputs for diverse tasks. These models typically comprise a vision encoder, a text encoder, and a general decoder. Our work aligns with the \textit{functional unification} approach, but with a novel focus on 3D vision-language tasks, diverging from the conventional 2D paradigm.

\subsection{Unified Vision-Language Models in 3D}

Initial efforts in 3D vision-language modeling, such as those by PointCLIP~\cite{zhang2022pointclip}, PointCLIP v2.~\cite{zhu2023pointclip}, CLIP2Point~\cite{huang2023clip2point}, and ULIP~\cite{xue2023ulip}, focus on adapting the 2D-based CLIP~\cite{clip} model for 3D applications. These works contribute to enhancing classification tasks for point clouds but often \textit{require additional components, like 3DETR~\cite{3DETR}, for tasks such as object detection}. Furthermore, rather than directly processing point clouds, they typically rely on projected multi-view images from point clouds during training or testing.

Building on these developments, Point-LLM~\cite{xu2023pointllm}, evolving from ULIP-2~\cite{ULIP-2} and incorporating the Vicuna~\cite{chiang2023vicuna} language model, engages in a dual-stage training process of feature alignment and instruction tuning. This approach equips Point-LLM with proficiency in classification, captioning, and dialogue. UniT3D~\cite{chen2023unit3d} introduces a unified framework employing transformer technology for 3D dense captioning and visual grounding, signifying a leap in simplifying 3D vision-language tasks. Further, 3D-VisTA~\cite{3D-vista}, a pre-trained transformer adept in 3D vision and text Alignment, excels in various tasks including 3D visual grounding and question answering. A key innovation of 3D-VisTA is the Scanscribe dataset, a novel contribution to 3D-VL pre-training, which uniquely operates without the need for multi-view images. However, 3D-VisTA~\cite{3D-vista} still requires fine-tuning of task-specific heads for different applications.

In conclusion, current models face notable limitations. They only support limited tasks, \textit{and often require additional task-specific module design and fine-tuning}, as summarized in Table~\ref{table:vlm3d_comparison}. Furthermore, they generally depend on multi-view images. Our method, however, extensively extends the range of tasks it can handle, particularly emphasizing dense prediction tasks such as semantic segmentation, instance segmentation, and grounded segmentation, all within a unified architecture using shared parameters. A distinctive aspect of our approach is its direct operation on point clouds, thereby bypassing the need for multi-view images.

\section{Uni3DL}

\subsection{Method overview}
The Uni3DL is a versatile architecture tailored for diverse 3D vision-language tasks, including 3D object classification, text-to-3D retrieval, 3D captioning, 3D semantic and instance segmentation, and 3D visual grounding. This architecture encompasses four integral modules: a \textbf{Text Encoder} for textual feature extraction; a \textbf{Point Encoder} dedicated to point feature learning; a \textbf{Query Transformer Module} with a sequence of cross-attention and self-attention layers to learn relations among object and text queries and voxel features from the Point Encoder; and a \textbf{Task Router}, adaptable and comprising multiple functional heads, including a text generation head for generating text outputs, a class head for object classification, and a mask head for producing segmentation masks, a grounding head for text-to-object grounding, and a 3D-Text matching head for 3D-text cross modal matching . With the combination of these functional heads, the task router selectively combines functional heads for different tasks. For example, the instance segmentation task combines object classification and mask prediction.

Given an input point cloud $\mathbf{P}$, our Uni3DL leverages a 3D U-Net  $\mathcal{E}_{I}$ to extract hierarchy voxel features $\mathbf{V}$, along with a text encoder $\mathcal{E}_{T}$ to obtain textural features $\mathbf{F}_{T} \in \mathbb{R}^{L_T \times C}$. Voxel features, textural features, along with learnable latent queries $\mathbf{F}_Q \in \mathbb{R}^{Q \times C}$ are fed into a unified decoder network to predict mask and semantic outputs, formulated as: 
\begin{equation}
    \mathbf{O}^m, \mathbf{O}^s = \mathcal{D}(<\mathbf{F}_Q, \mathbf{F}_{T}>, \mathbf{V}),
\end{equation}
where $\mathbf{O}^m$ and $\mathbf{O}^s$ denote mask outputs and semantic outputs, $<,>$ denotes feature concatenation.

\begin{figure*}[ht]
    \centering
    \includegraphics[width=\textwidth]{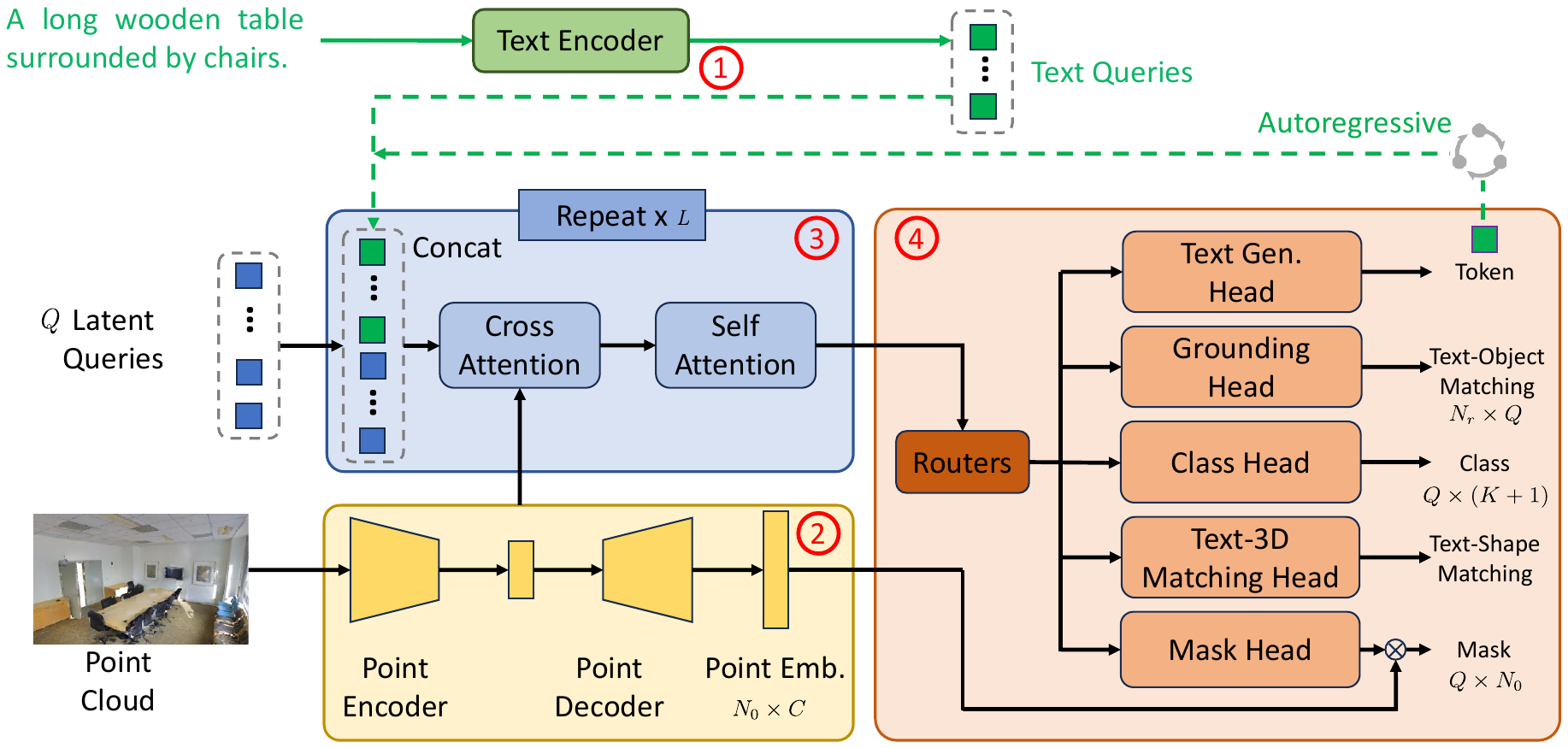}
    \caption{Overview of the Uni3DL Model. The Uni3DL is engineered for multifaceted 3D data tasks, including classification, retrieval, captioning, semantic and instance segmentation, as well as visual grounding. The architecture is composed of four principal modules: \ding{172} a \textbf{Text Encoder} for textual feature extraction; \ding{173} a \textbf{Point Encoder} for point feature learning; \ding{174} a \textbf{Query Transformer} Module, which is the cornerstone of the system with a sequence of cross-attention and self-attention operations between latent queries, text queries and voxel features derived from the Point Encoder; and \ding{175} a \textbf{Task Router} module, which comprises, as needed for the given task, text generation head for generating descriptive text, a grounding head for text-to-object grounding, a class head for object classification task, a mask head dedicated to segmentation, and a text-3D matching head for 3D-text cross modal matching. The text generation head functions in an autoregressive manner and predicts one token at each forward step.
    }
    \label{fig_overview}
\end{figure*}

\subsection{Point Cloud and Text Encoder}
The architecture of our point feature extraction network employs a sparse 3D convolutional U-net structure based on the MinkowskiEngine framework~\cite{choy20194d}, featuring both an encoder and a decoder network. A colored input point cloud, denoted as $\mathbf{P} \in \mathbb{R}^{N_0 \times 6}$, undergoes quantization into $N_0$ voxels represented as $\mathbf{V}_0 \in \mathbb{R}^{N_0 \times 3}$, with each voxel capturing the average RGB color from the points it contains as the initial voxel features. Several convolutional and downsampling layers are sequentially applied to extract high-level voxel features, followed by deconvolutional and upsampling layers to recover voxels to their original resolutions. Supposing the U-Net has $S$ stages of feature blocks, at each stage $s \in [1,..,S]$, we can get voxel features $\mathbf{V}_s \in \mathbb{R}^{N_s \times C_s}$, where $N_s$ denotes the number of valid voxels at stage $s$, and $C_s$ denotes the corresponding feature dimension. We then project all voxel features to the same dimension $D$, resulting in a set of feature maps $\{\mathbf{V}_s \in \mathbb{R}^{N_s \times C}\}_{s=1}^S$. The last feature map ($\mathbf{V}_S$) is used as point embeddings to calculate per-point mask, while the remaining feature maps $\{\mathbf{V}_s\}_{s=1}^{S-1}$ are fed into the transformer module to enhance latent and text queries.
For text inputs, we use the CLIP tokenizer~\cite{clip} along with a transformer-based network for textural feature learning. 

\subsection{Query Transformer Module}
We follow query-based transformer architecture~\cite{detr,slotatt,schult2023mask3d,xdecoder} to design our decoder network. Given voxel features $\{\mathbf{V}_s\}_{s=1}^{S-1}$, our transformer module refines latent queries $\mathbf{F}_Q$ and text queries $\mathbf{F}_T$ by a sequence of $L$ decoder layers. At each layer $l=[1...,L]$, we refine queries by cross-attending to voxel features $\{\mathbf{V}_s\}_{s=1}^{S-1}$, formulated as:
\begin{equation}
    <\hat{\mathbf{F}}_Q^{l}, \hat{\mathbf{F}}_T^{l}> = \text{Cross-Att}(<\mathbf{F}_Q^{l-1}, \mathbf{F}_T^{l-1}>, \mathbf{V}_s).
\end{equation}
We repeat this process for each feature level $s=[1,2,...,S-1]$.


\noindent \textbf{Masked Attention}.
To enhance object localization capability, we follow the attention block design in Mask2Fomer~\cite{mask2former} and use masked attention instead of vanilla cross-attention where each query only attends to masked voxels predicted by the previous layer.

\noindent \textbf{Voxel Sampling}. Point clouds in a batch usually have different numbers of points, leading to differing voxel quantities. Current transformer implementations generally require a fixed length of inputs in each batch entry. To enable efficient batch-wise training, for each feature level $s$, before feeding voxel features into the decoder network. The sampled voxel features are then utilized across all cross-attention layers following~\cite{schult2023mask3d}.

We further enhance object and text queries through self-attention layers and feed-forward layers, formulated as:
\begin{equation}
    <\hat{\mathbf{F}}_Q^{l}, \hat{\mathbf{F}}_T^{l}> = \text{Self-Att}(<\hat{\mathbf{F}}_Q^{l}, \hat{\mathbf{F}}_T^{l}>),
\end{equation}
\begin{equation}
    <\mathbf{F}_Q^l, \mathbf{F}_T^l> = \text{FFN}(<\hat{\mathbf{F}}_Q^{l}, \hat{\mathbf{F}}_T^{l}>).
\end{equation}

\subsection{Task Router}
To support diverse 3D vision-language tasks, we design multiple functional heads thus different tasks can be achieved by compositions of heads. For example, the 3D instance segmentation task includes two heads, object classification, and mask prediction. Consequently, the Uni3DL model harnesses a consistent set of parameters, while applying unique routing strategies for each specific task, ensuring efficient task decomposition and substantial parameter reuse across different tasks. We show the head composition of different tasks in Table~\ref{tab_heads}. 

\begin{table}[!h]
    \centering
    \resizebox{0.49\textwidth}{!}{
    \begin{tabular}{c|ccccc}
        \toprule
        Task & Obj Cls. & Mask & Grounding & Text Gen. & Text-3D Matching \\
        \midrule
        Semantic Segmentation & \checkmark & \checkmark &  &  &  \\
        Instance Segmentation & \checkmark & \checkmark &  &  &  \\
        Grounded Segmentation  &  & \checkmark & \checkmark &  &  \\
        Captioning &  &  &  & \checkmark &  \\
        Retrieval &  &  &  &  & \checkmark \\
        Shape Classification &  &  &  &  & \checkmark \\
        \bottomrule
    \end{tabular}}
    \caption{Head compositions of different tasks. Obj Cls denotes object classification head, Text Gen. denotes text generation head.}
    \label{tab_heads}
\end{table}

\noindent \textbf{Object Classification Head}.
We select the first $Q$ output semantic outputs for object classification. Given refined semantic queries $\mathbf{O}^s \in \mathbb{R}^{Q\times C}$, and $K$ semantic classes with additional background class. We first feed all $K+1$ class names to the text encoder to get class embeddings $\mathbf{C}_{emb} \in \mathbb{R}^{(K+1) \times C}$ , and calculate classification probabilities as $\mathbf{O}_c = \mathbf{O}^s \cdot \mathbf{C}_{emb}^T$, where $\cdot$ denotes the dot product between matrixes.

During training, we calculate cross-entropy loss between predicted classification probabilities $O_c$ and ground truth (GT) class labels $C_{gt}$ to formulate classification loss as:
\begin{equation}
    \mathcal{L}_{cls} = \lambda_{cls} \text{CE}(\mathbf{O}_c, C_{gt}),
\end{equation}
where CE denotes cross-entropy loss.

\noindent \textbf{Mask Head}.
Given mask output $\mathbf{O}^m \in \mathbb{R}^{Q\times C}$, and full-resolution voxel features $\mathbf{V}_s \in \mathbb{R}^{N_0 \times C}$, we calculate voxel mask as $\mathbf{O}_{m} =  \mathbf{O}^m \cdot \mathbf{V}_s^T$. The output voxel mask $\mathbf{O}_{m} \in \mathbb{R}^{Q\times N_0}$, where each row denotes a mask for the corresponding latent query.

During training, given ground truth object mask $\mathbf{M}_{gt}$, we calculate mask loss as:
\begin{equation}
    \mathcal{L}_{mask} = \lambda_{bce} \text{BCE}(\mathbf{O}_m, \mathbf{M}_{gt}) + \lambda_{dice} \text{DICE}(\mathbf{O}_m, \mathbf{M}_{gt}),
\end{equation}
where BCE and DICE denote binary cross-entropy loss and dice loss respectively.

\noindent \textbf{Grounding Head}.
Visual grounding requires matching text descriptions to visual objects. We first generate textural embeddings $\mathbf{T}_{emb} \in \mathbb{R}^{N_r \times C}$ by feeding all referring sentences to the text encoder. We select the first $Q$ output semantic queries $\mathbf{O}^s \in \mathbb{R}^{Q \times C}$ as object embeddings. Then, we calculate object-text similarity by
\begin{equation}
    \mathbf{S}_{t} = \text{Softmax}(e^{\eta} \; \mathbf{T}_{emb} \cdot (\mathbf{O}^s)^T),
\end{equation}
where $\mathbf{S}_{t} \in \mathbb{R}^{N_r \times Q}$ and $\eta$ denotes a learnable scaling parameter. Softmax operation is applied on the last dimension.

Following DETR~\cite{detr}, we use Hungarian matching to get ground truth matching labels $T_{gt} \in \mathbb{R}^{N_r}$. We modified the original mask matching module in DETR to adapt it for voxel masks. We then calculate cross-entropy loss as:
\begin{equation}
    \mathcal{L}_{gc} = \text{CE} (\mathbf{S}_{t}, T_{gt}).
\end{equation}

Following the common practice of 3D visual grounding practice ~\cite{chen2020scanrefer,cai20223djcg}, we design a lightweight classification that takes textural embeddings as inputs and predicts the existence of all $K$ candidate object categories. Given input textural embeddings $\mathbf{T}_{emb} \in \mathbb{R}^{N_r \times C}$, we use a single-layer MLP network to calculate the probabilities matrix $\mathbf{T}_{cls} \in \mathbb{R}^{N_r \times K}$ over $K$ candidate object categories and calculate multi-label classification loss as:
\begin{equation}
    \mathcal{L}_{gtxt} = \text{BCE}(\mathbf{T}_{cls}, \mathbf{T}_{cls}^{gt}),
\end{equation}
where $\mathbf{T}_{cls}^{gt} \in \mathbb{R}^{N_r\times K}$ denotes the ground truth labels of category existence.

Additional grounding mask $\mathcal{L}_{gmask}$ is calculated similarly to the mask head. The overall grounding loss is calculated as:
\begin{equation}
    \mathcal{L}_{grd} = \lambda_{gc} \mathcal{L}_{gc} + \mathcal{L}_{gtxt} + \mathcal{L}_{gmask}.
\end{equation}

\noindent \textbf{Text Generation Head}.
In the context of 3D captioning, our method begins by generating textural embeddings for each token within the vocabulary, which comprises $V$ tokens, utilizing the text encoder. Subsequently, we use the last $L_T$ semantic outputs generated by the decoder network and calculate the dot product against the token embeddings, resulting in an affinity matrix $\mathbf{S}_{cap} \in \mathbb{R}^{L_T \times V}$. The cross-entropy loss is calculated as:
\begin{equation}
    \mathcal{L}_{cap} = \lambda_{cap} \text{CE}(\mathbf{S}_{cap}, y_{cap}),
\end{equation}
where $y_{cap}$ is the ground truth token indices. 

During training, a causal masking strategy is adopted in all self-attention layers of the decoder network. During inference, our model predicts one token at each time and gets 3D captions in an autoregressive way.

\noindent \textbf{Text-3D Matching Head}. Our Uni3DL uses decoupled point and text encoder networks. To predict text-3D matching, the last output semantic token is used as the shape embedding with a dimension of $\mathbb{R}^{1 \times C}$. Given a batch of $B$ text-shape pairs, the retrieval head computes the similarities between 3D shape embeddings and corresponding text embeddings as $\mathbf{S}_{ret} \in \mathbb{R}^{B \times B}$, and calculates retrieval loss as: 
\begin{equation}
    \mathcal{L}_{ret} = \lambda_{ret} \text{CL}(\mathbf{S}_{ret}, y_{ret}),
\end{equation}
where $y_{cap} \in \mathbb{R}^{1\times B}$ denotes the ground truth matching indices. CL denotes contrastive loss defined in CLIP~\cite{clip}.

\noindent \textbf{Multi-Task Training}.
During pretraining, we simultaneously train the whole network with both object classification head, mask head, grounding head, text generation head, and text-3D matching head. The overall loss is formulated as:
\begin{equation}
    \mathcal{L} = \mathcal{L}_{cls} + \mathcal{L}_{mask}+ \mathcal{L}_{grd}+ \mathcal{L}_{cap}+ \mathcal{L}_{ret}.
\end{equation}

\begin{table*}[!ht]
\centering
\resizebox{\textwidth}{!}{
\begin{tabular}{l|ccc|cc|cccc|ccc|ccc|cc}
\toprule
\multirow{3}{*}{Method} & \multicolumn{3}{c|}{\underline{Semantic Segmentation}} & \multicolumn{2}{c|}{\underline{Object Detection}} & \multicolumn{4}{c|}{\underline{Instance Segmentation}} & \multicolumn{3}{c|}{\underline{Grounded Segmentation}} & \multicolumn{3}{c|}{\underline{3D Captioning}} & \multicolumn{2}{c}{\underline{3D Retrieval}} \\
& \multicolumn{2}{c}{S3DIS (Area 5)} & SN Val & \multicolumn{2}{c|}{SN Val} & \multicolumn{2}{c}{SN Val} & \multicolumn{2}{c|}{S3DIS (Area 5)} & \multicolumn{3}{c|}{ScanRefer} & \multicolumn{3}{c|}{Cap3D}  & \multicolumn{2}{c}{Text2Shape} \\
& mIoU & mAcc & mIoU & bAP\(_{50}\) & bAP\(_{25}\) & mAP & mAP\(_{50}\) & mAP\(_{50}\) & mAP\(_{25}\) & mIoU & Acc@0.25 & Acc@0.5 & B-1 & R & M & R@1 & R@5 \\
\hline
MinkowskiNet42\cite{choy20194d} & 67.1 & 74.4 & 72.2 &  - & - & - & - & - & - & - & - & - & - & -  & - & - & - \\
FastPointTransformer\cite{park2022fast} & 68.5 & 76.5 & 72.1 & - & - & - & - & - & - & - & - & - & - & - & - & -  \\
PointNeXt-XL\cite{qian2022pointnext} & 71.1 & 77.2 & 71.5 & - & - & - & - & - & - & - & - & - & -  & - & - & - & -  \\
StratifiedTransformer~\cite{lai2022stratified} & 72.0 & \underline{78.1} & 73.1 & - & -  & - & - & - & - & - & - & - & - & - & - & - & -  \\
PointTransformerV2~\cite{wu2022point} & 71.6 & 77.9 & 74.4 & - & - & - & - & - & - & - & - & - & - & - & -  & - & -  \\
EQ-Net\cite{yang2022unified} & 71.3 & * & \underline{75.3} & - & - & - & - & - & - & - & - & - & & - & - & - & -  \\
Swin3D\cite{yang2023swin3d} & \underline{72.5} & * & 75.2 & - & - & - & - & - & - & - & - & - & - & - & - & - & - \\
\textcolor{gray!70}{Swin3D$^\dagger$\cite{yang2023swin3d}} & \textcolor{gray!70}{\textbf{73.0}} & * & \textcolor{gray!70}{75.6} & - & - & - & - & - & - & - & - & - & - & - & - & - & - \\
\hline
VoteNet~\cite{xie2021venet} & & - & -  & 33.5 & 58.6 &  - & - & - & - & - & - & - & - & - & - & - & - \\
3DETR~\cite{misra2021end} & - & - & -  & 47.0 & 65.0 &  - & - & - & - & - & - & - & - & - & - & - & - \\
CAGroup3D~\cite{wang2022cagroup3d} & -  & - & - & {61.3} & \underline{75.1} & - & - & - & - & - & - & - & -  & - & - & - & -\\
\hline
PointGroup\cite{jiang2020pointgroup} & *  & * & * & * & * & 34.8 & 56.7 &  57.8 & * & - & - & - & - & - & - & -  & - \\
MaskGroup\cite{zhong2022maskgroup} & *  & * & *  & * & * & 42.0 & 63.3 & 65.0 & * & - & - & - & - & - & - & - & -  \\
SSTNet\cite{Liang_2021_ICCV} & *  & * & * & * & *  & 49.4 & 64.3 &  59.3 & * & - & - & - & - & - & - & - & - \\
SoftGroup\cite{softgroup} & *  & * & * & 59.4 & 71.6 & 50.4 & \underline{76.1} & {66.1} & *  & - & - & - & - & - & - & - & -  \\
Mask3D\cite{schult2023mask3d} & * & * & * & 56.2 & 70.2  & 55.2 & 73.7 & \underline{68.4} & \underline{75.2}  & - & - & - & - & - & - & - & -  \\
{Mask-Att-Free$^\dagger$\cite{Lai_2023_ICCV}} & * & * & * & \underline{63.9} & {73.5}  & {\underline{58.4}} & {{75.9}} & {\textbf{69.1}} & {\textbf{75.7}}  & - & - & - & - & - & - & - & - \\
\hline
TGNN (GRU)~\cite{huang2021text} & - & - & - & - & - & - & - & - & -  & 26.1 & 35.0 & 29.0  & - & - & - & - & -  \\
TGNN (BERT)~\cite{huang2021text} & - & - & - & - & - & - & - & - & -  & \underline{27.8} & \underline{37.5} & \underline{31.4}  & - & - & - & - & -  \\
\hline
InstructBLIP-7B~\cite{dai2023instructblip} & -& - & - & - & - & - & - & - & - & - & - & - & 11.2 & 13.9 & 14.9 & * & * \\
InstructBLIP-13B~\cite{dai2023instructblip}  & -& - & - & - & - & - & - & - & - & - & - & - & \underline{12.6} & \underline{15.0} & \textbf{16.0} & *  & * \\
PointLLM-7B~\cite{xu2023pointllm} & -& - & - & - & - & - & - & - & - & - & - & - & 8.0 & 11.1 & 15.2 & *  & *  \\
PointLLM-13B~\cite{xu2023pointllm} & -& - & - & - & - & - & - & - & - & - & - & - & 9.7 & 12.8 & 15.3 & * & *  \\
\hline
FTST~\cite{chen2019text2shape} & -& - & - & - & - & - & - & - & - & - & - & - & - & - & - & 0.2 & 1.6 \\
FMM~\cite{chen2019text2shape}  & -& - & - & - & - & - & - & - & - & - & - & - & - & - & -  & 0.2 & 2.4  \\
Y2S~\cite{han2019y2seq2seq} & -& - & - & - & - & - & - & - & - & - & - & - & * & * & *  &  2.9 & 9.2 \\
Parts2Words (no parts)~\cite{tang2023parts2words} & -& - & - & - & - & - & - & - & - & - & - & - & - & -  & - & \underline{5.1} & \underline{17.2} \\
\textcolor{gray!70}{Parts2Words$^\dagger$~\cite{tang2023parts2words}} & -& - & - & - & - & - & - & - & - & - & - & - & - & -  & - &  \textcolor{gray!70}{\textbf{12.7}} & \textcolor{gray!70}{\underline{33.0}} \\
\hline
Ours & \textbf{72.7} & \textbf{79.3} & \textbf{76.2} & \textbf{67.7} & \textbf{77.1} & \textbf{60.9} & \textbf{80.9} & 65.3 & {74.3} & \textbf{32.3} & \textbf{39.4} & \textbf{36.4} & \textbf{31.6} & \textbf{33.1} & \underline{14.4} & \textbf{5.8} & \textbf{19.7} \\
\bottomrule
\end{tabular}}
\caption{Performance of our Uni3DL on different segmentation and VL tasks. `SN' denotes the ScanNet (v2) dataset. `*' indicates the model is capable of the task without a reported metric, and `-' signifies the model lacks this specific capability. The results highlighted in \textbf{bold} and \underline{underline} denote the best and second-best outcomes, respectively, for each column. Note that Swin3D$^\dagger$ uses extra training data (Structure3D~\cite{zheng2020structured3d}), and Parts2Words$^\dagger$~\cite{tang2023parts2words} uses additional part labels. Mask-Att-Free$^\dagger$ uses several task-specific designs, including a center regression module and position-aware components, to improve the performance.}
\label{tab_all_results}
\vspace{-5pt}
\end{table*}

\section{Experiments}

\subsection{Dataset}
We pretrain our Uni3DL on three datasets, including ScanNet (v2)~\cite{dai2017scannet} for instance segmentation, ScanRefer~\cite{chen2020scanrefer} for visual grounding, and Cap3D Objaverse~\cite{cap3d} dataset for 3D captioning and text-3D cross-modal retrieval. 

\noindent \textbf{ScanNet (v2)~\cite{dai2017scannet}} captures RGB-D videos with 2.5 million views from more than 1,500 3D scans. Following the official benchmark, we use 1,201 scenes for training, 312 for validation, and 100 indoor scenes for online evaluation. There are in total 20 semantic labels, 18 of which are instance classes. 

\noindent \textbf{ScanRefer~\cite{chen2020scanrefer}} dataset contains 51,583 referring descriptions of 11,046 objects from 800 ScanNet scenes. We use 562 scenes for training and 141 scenes for evaluation.

\noindent \textbf{Cap3D Objaverse~\cite{cap3d}} dataset, is derived from Objaverse, one of the largest 3D datasets with around 800K objects. It features 660K 3D-text pairs, created using an automated captioning process. We randomly select 80\% for training and the remaining 20\% for evaluation\footnote{To ensure a fair comparison with PointLLM, we filter out 200 objects used for benchmark evaluation from our training set and report the performance on the same 200 objects.}. 

For model evaluation, other than ScanNet (v2), ScanRefer, Cap3D, we use additional S3DIS~\cite{armeni20163d} to evaluate both semantic and instance segmentation, Text2Shape~\cite{chen2019text2shape} to evaluate text-to-3D retrieval.

\noindent \textbf{S3DIS} dataset contains 6 large-scale areas with 271 scenes, and 13 semantic categories are annotated. Following previous works, we use 68 scenes in Area 5 for validation and the others for model training.

\noindent \textbf{Text2Shape~\cite{chen2019text2shape}} contains 8,447 table instances and 6,591 chair instances from the ShapeNet dataset, along with 75,344 natural language descriptions. We use the same training/test split as~\cite{chen2019text2shape}.

\subsection{Implementation Details}
In this work, we employ 150 latent queries and an additional latent query for scene-level tasks. The point encoder-decoder network is based on Minkowski Res16UNet34C~\cite{choy20194d} and pretraiend from Mask3D~\cite{schult2023mask3d}, and we use 12 transformer layers for the language encoder. Our Query Transformer module consists of 15 ($L=15$) transformer layers. The segmentation weights $\lambda_{\text{cls}}, \lambda_{\text{bce}}, \lambda_{\text{dice}}$ are set 2.0, 5.0, 5.0, grounding classification weight to $\lambda_{\text{gc}}$ to 0.4, captioning and retrieval weight $\lambda_{\text{cap}}, \lambda_{\text{ret}}$ are set to 2.0.

During pretraining, the voxel size is set to 0.02m for 3D scans (e.g., ScanNet (v2)) and 0.01 for normalized 3D shapes (e.g., Cap3D Objaverse), with a batch size of 8 for 3D scans and 12 for 3D-text pairs. Input scenes are augmented by random flips along the X and Y axes, and rotations along the X, Y, and Z axes. Color augmentations, including jittering, brightness, and contrast adjustments, are also applied. The training process spans 50 epochs using the AdamW optimizer~\cite{loshchilov2018decoupled}, taking approximately 20 hours on four NVIDIA A100 GPUs. Details about pertaining and task-specific finetuning can be found in Appendix \ref{supp_settings}.

During inference, the top 200 (for S3DIS) and 500 (for ScanNet (v2)) instances with the highest classification scores are retained for the instance segmentation task.

\subsection{3D Semantic/Instance Sementation}
We compare 3D semantic segmentation, object detection, and instance segmentation performance with previous STOA methods in Table~\ref{tab_all_results}. From the table, our Uni3DL method achieves comparable or superior performance than previous STOA methods on semantic/instance segmentation on S3DIS and ScanNet (v2) datasets. Specifically, our method archives the best performance on ScanNet (v2) semantic segmentation, with a mIoU of 76.2. Figure~\ref{fig_seg} shows qualitative results of our method. Additional qualitative results are presented in the Appendix \ref{supp_vis_seg}.

\begin{figure}[ht]
  \centering
  \begin{minipage}[t]{0.11\textwidth}
    \includegraphics[width=\linewidth]{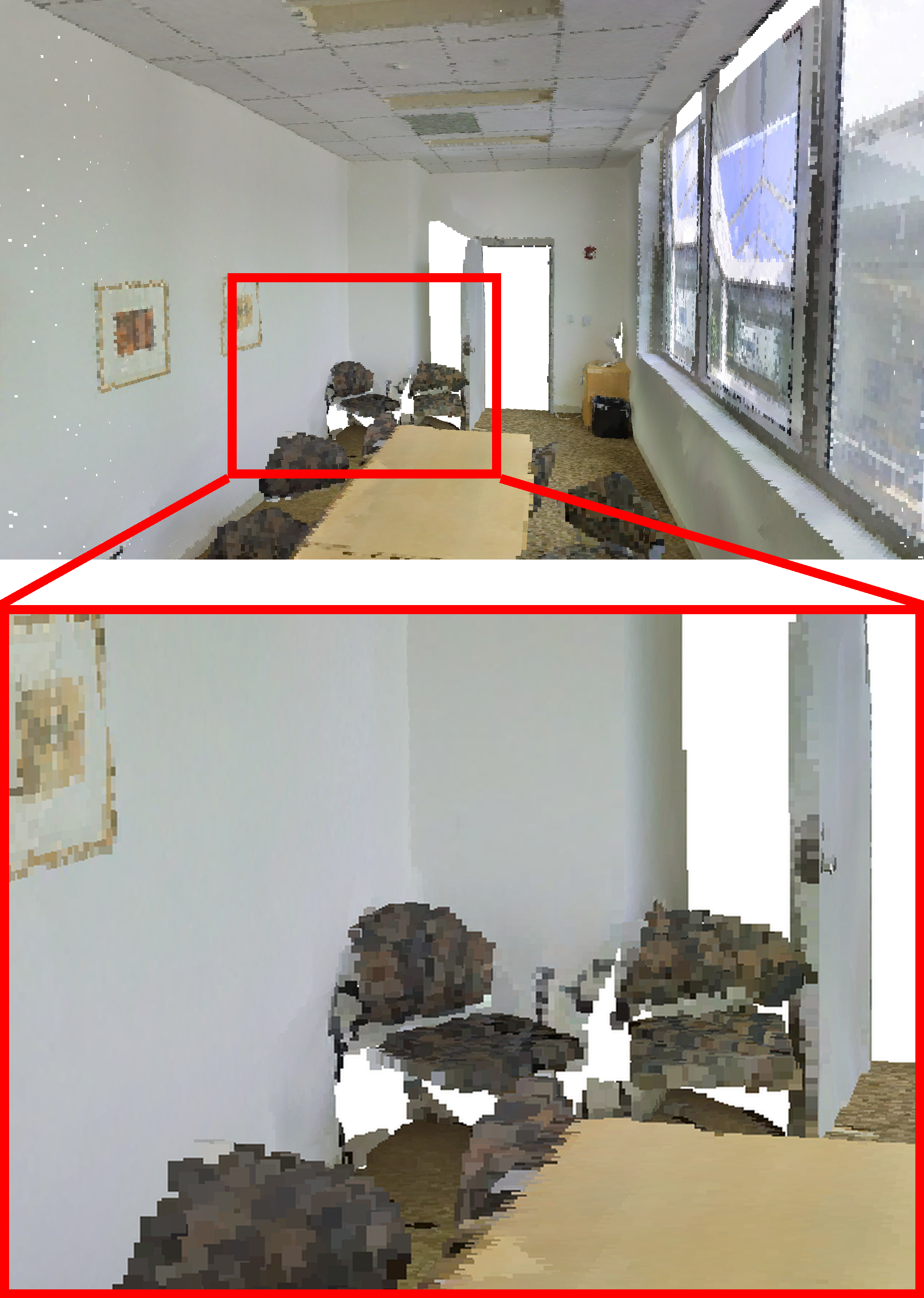}
    \includegraphics[width=\linewidth]{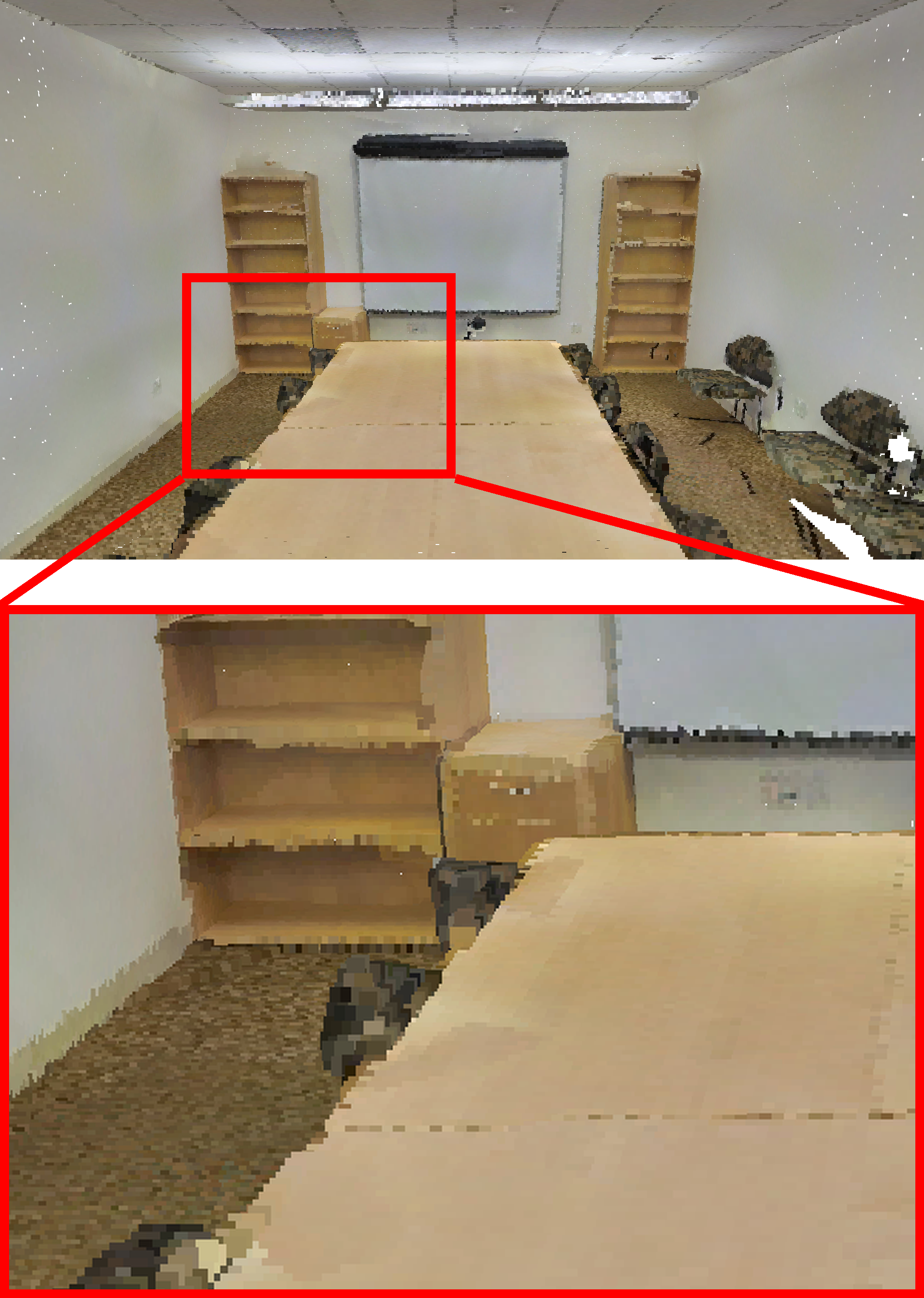}
    \caption*{Input}
  \end{minipage}
  \hfill
  \begin{minipage}[t]{0.11\textwidth}
    \includegraphics[width=\linewidth]{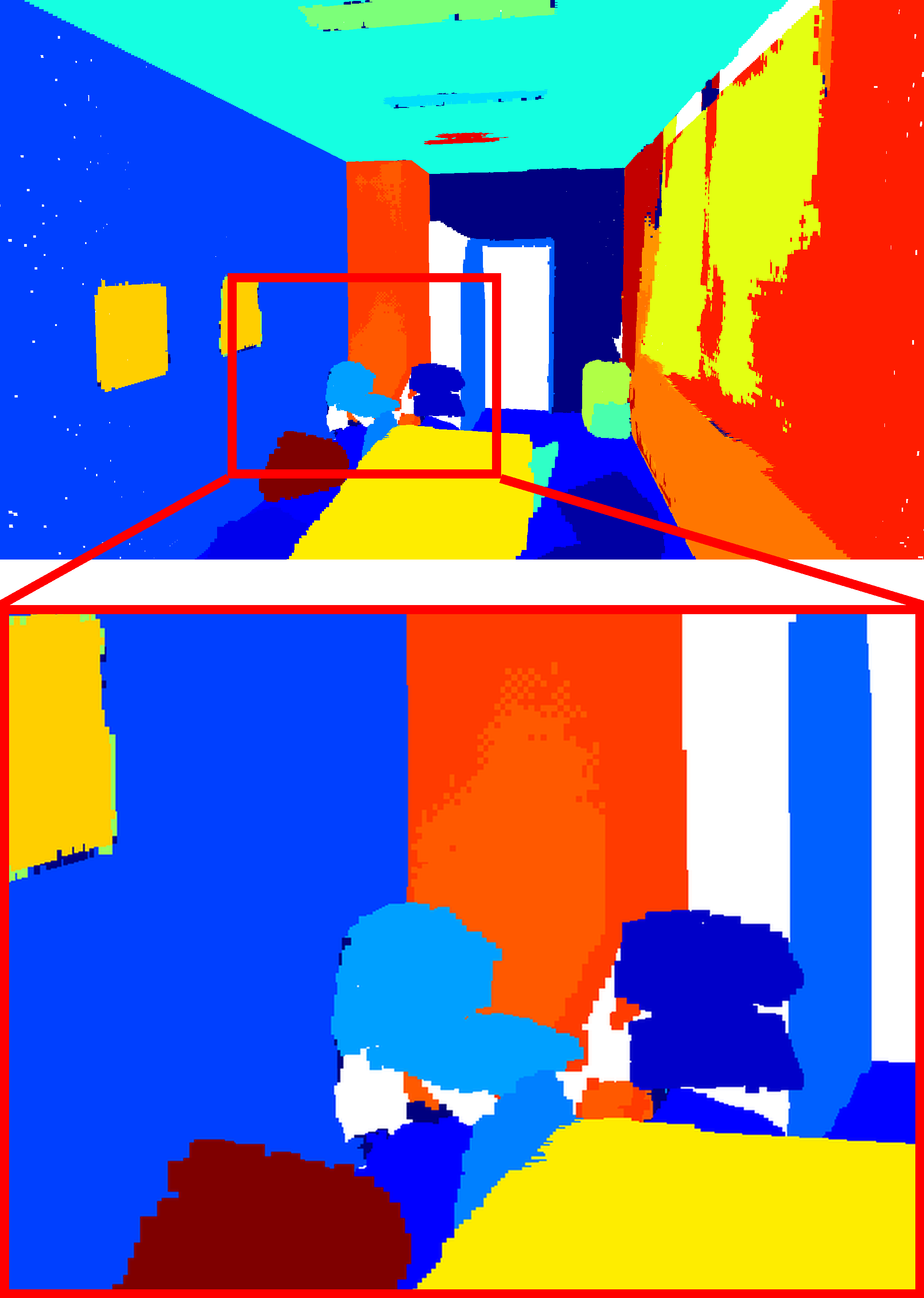}
    \includegraphics[width=\linewidth]{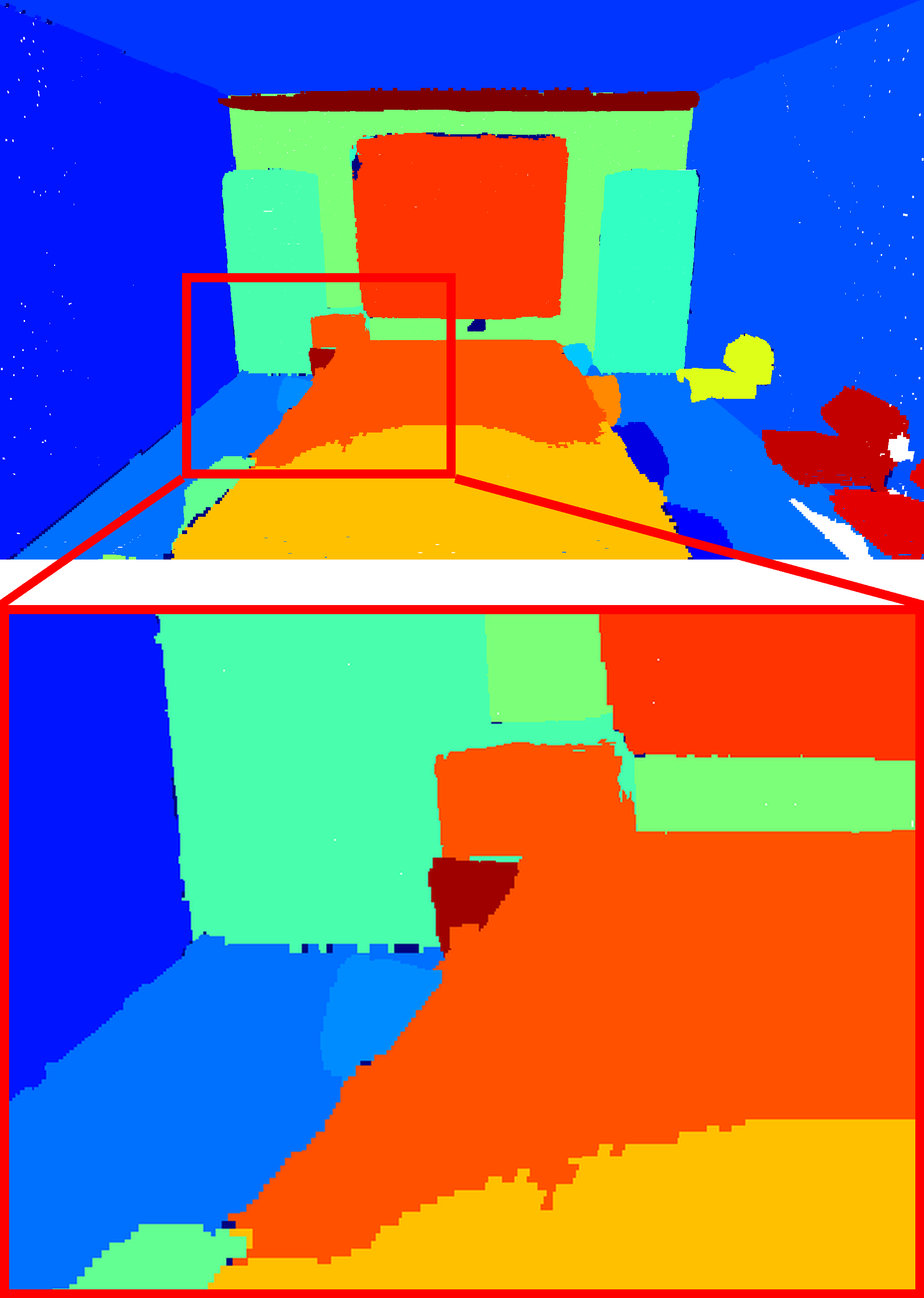}
    \caption*{Baseline Inst.}
  \end{minipage}
  \hfill
  \begin{minipage}[t]{0.11\textwidth}
    \includegraphics[width=\linewidth]{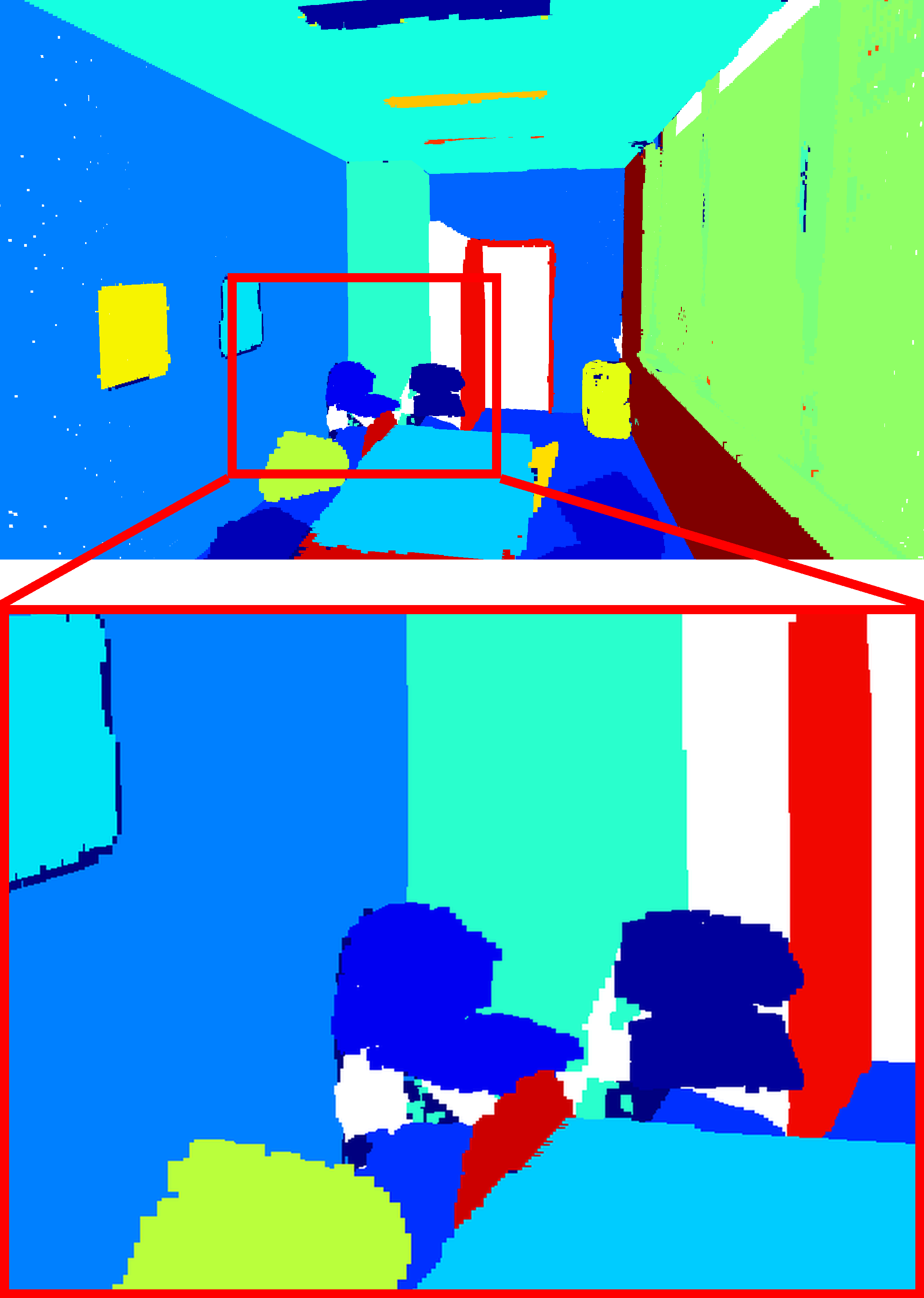}
    \includegraphics[width=\linewidth]{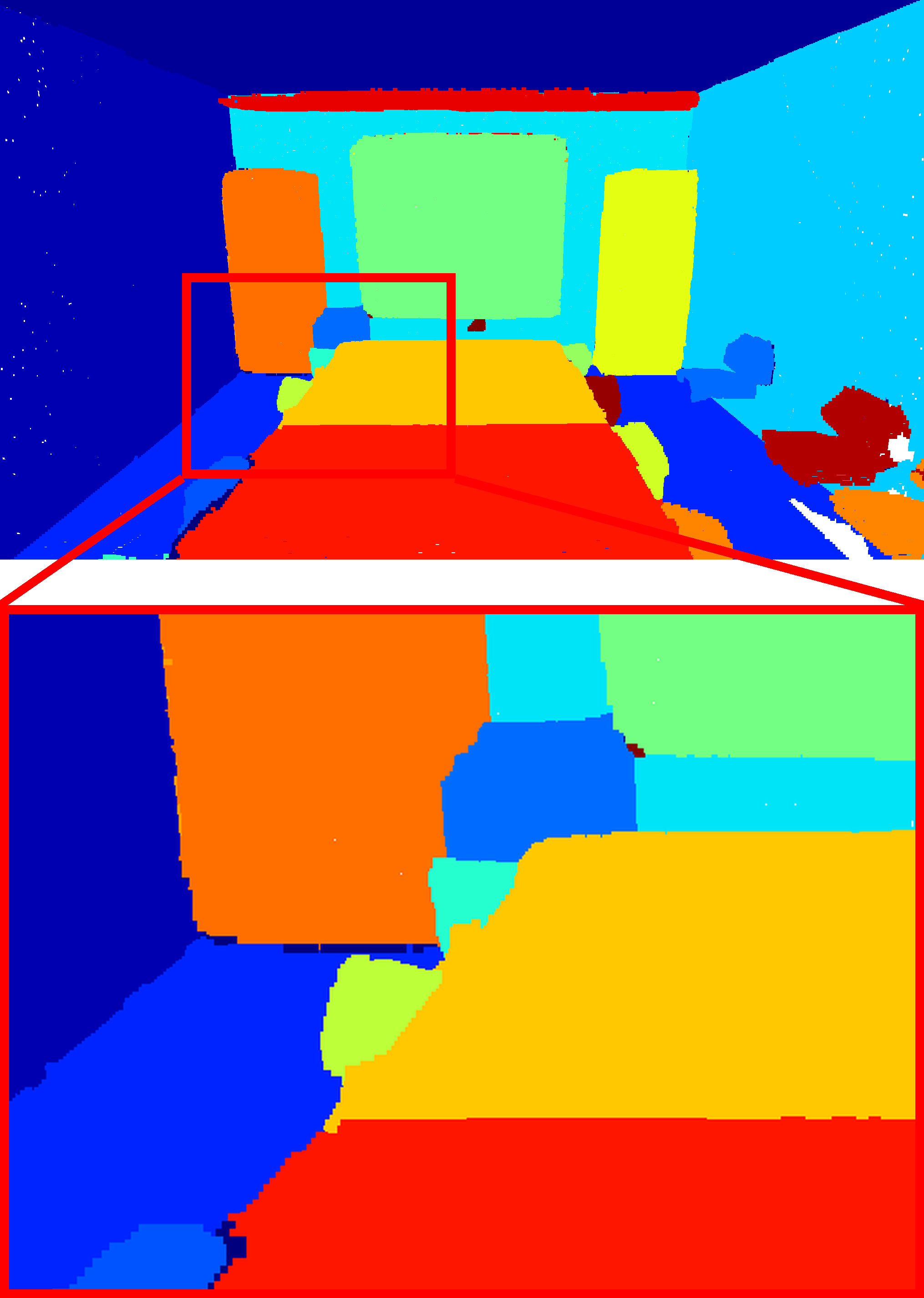}
    \caption*{Ours Inst.}
  \end{minipage}
  \hfill
  \begin{minipage}[t]{0.11\textwidth}
    \includegraphics[width=\linewidth]{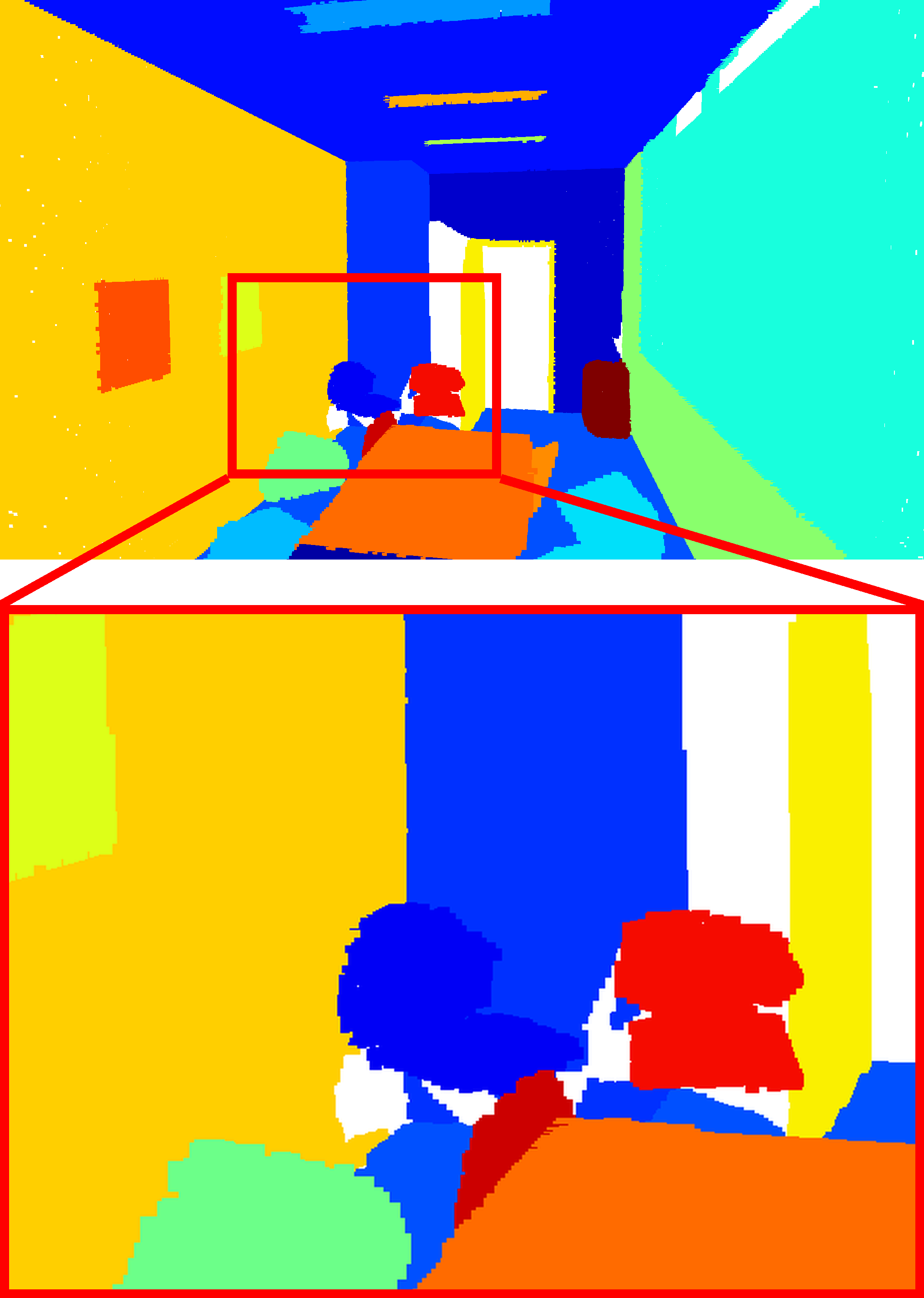}
    \includegraphics[width=\linewidth]{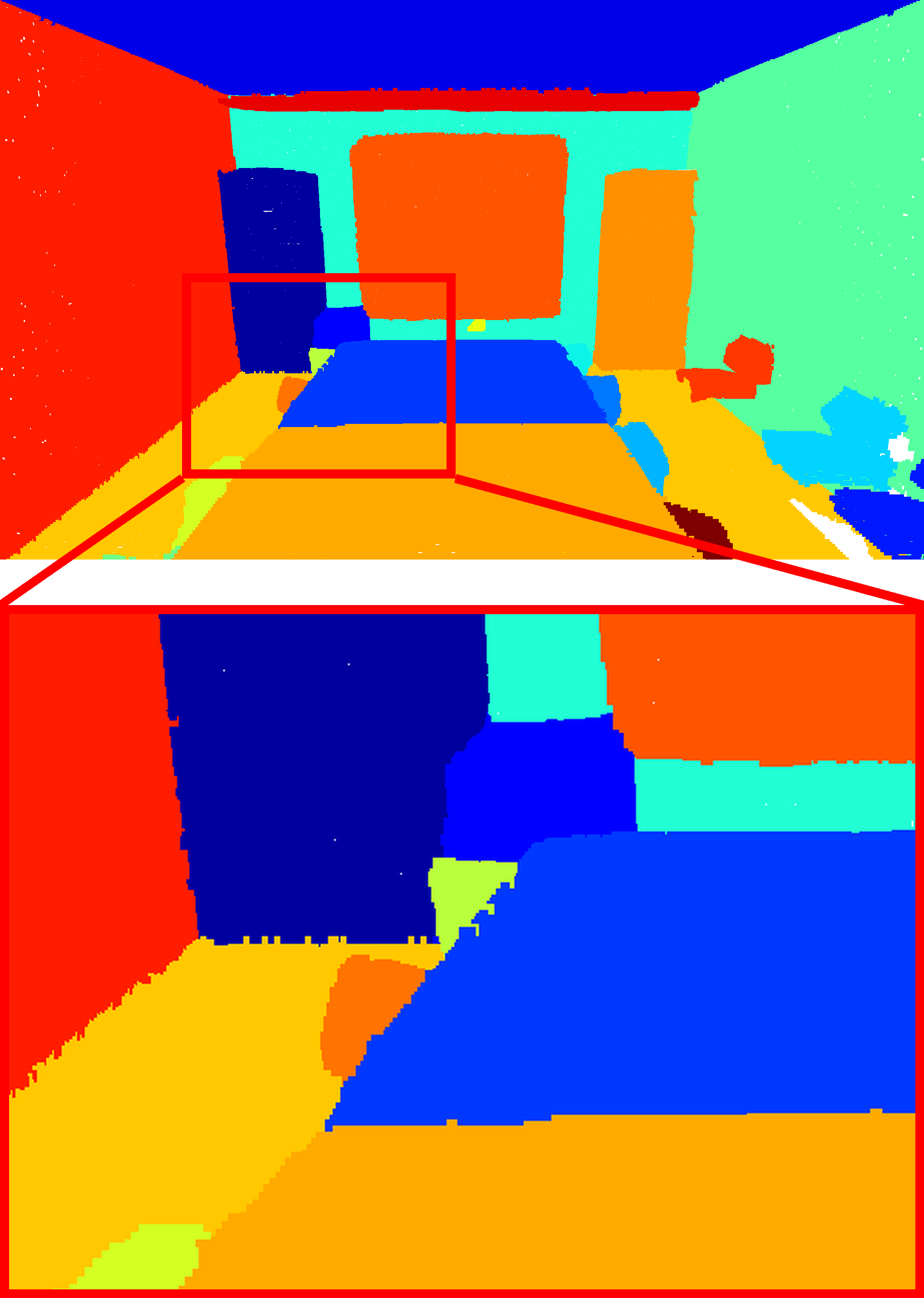}
    \caption*{GT Inst.}
  \end{minipage}
  \hfill
  \caption{Instance (Inst.) segmentation results on S3DIS dataset. We show results of the baseline method trained from scratch and our finetuned model.}
  \label{fig_seg}
\end{figure}

\subsection{3D Visual Grounding}
We compare the 3D grounded segmentation performance of our Uni3DL with previous STOA methods TGNN (GRU)~\cite{huang2021text} and TGNN (BERT)~\cite{huang2021text} in Table~\ref{tab_all_results}. Our method achieves significantly better performance than previous SOTA methods as indicated by instance-average IoU, and accuracy at the IoU thresholds of 0.25 and 0.5. Figure ~\ref{fig_refer} shows qualitative results of our method. More qualitative results are presented in Appendix \ref{supp_vg_seg}. Grounded localization performance can be found in Appendix \ref{supp_grounded_loc}.

\begin{figure}[ht]
  \centering
  \begin{subfigure}[t]{0.45\textwidth}
    \includegraphics[width=0.49\linewidth]{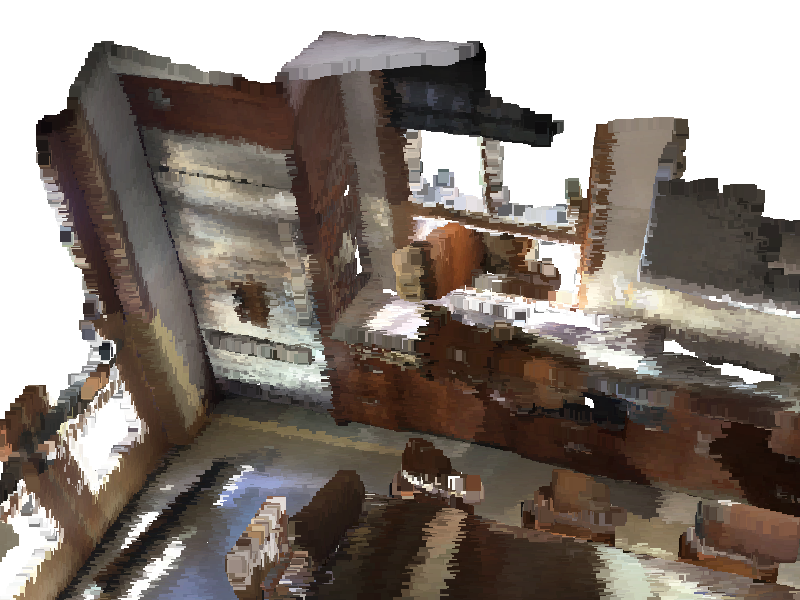}
    \hfill
    \includegraphics[width=0.49\linewidth]{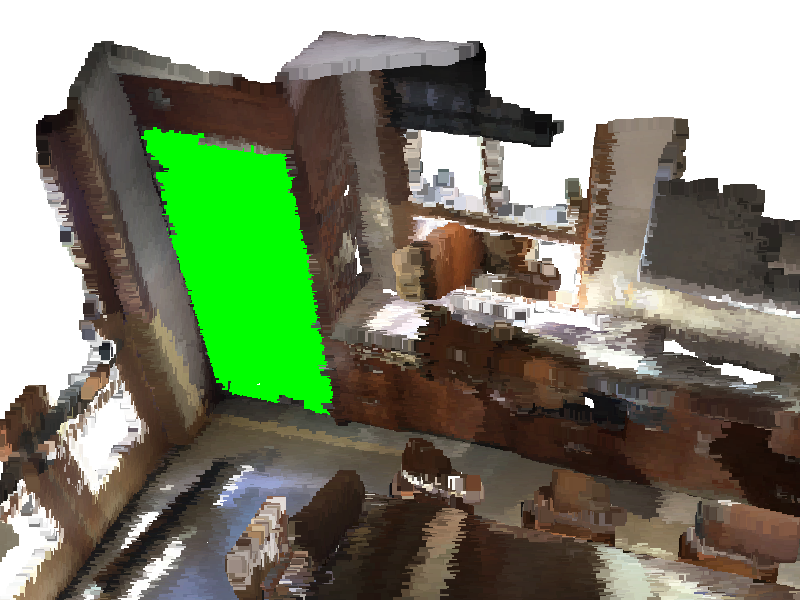}
    \caption*{Refer: It is a stainless steel refrigerator. the refrigerator sits to the left of the windows above the counter.}
  \end{subfigure}
  \hfill
  \begin{subfigure}[t]{0.45\textwidth}
    \includegraphics[width=0.49\linewidth]{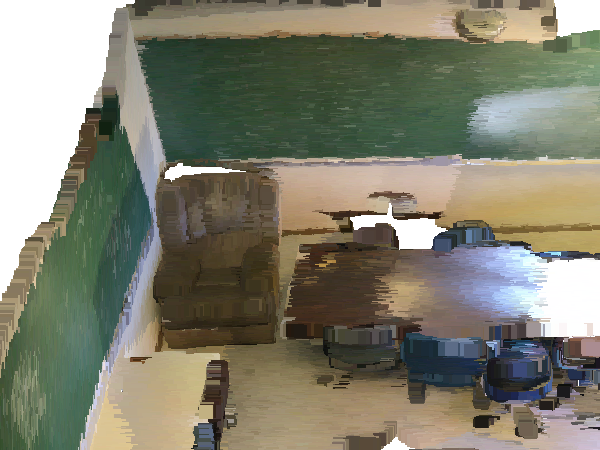}
    \hfill
    \includegraphics[width=0.49\linewidth]{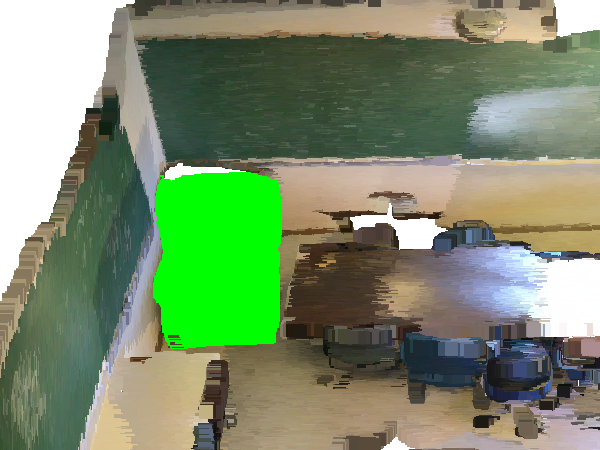}
    \caption*{Refer: This is the brown easy chair at the back of the room where the two chalk boards meet. it is a brown easy chair.}
  \end{subfigure}
  \hfill
  \caption{Results of grounded segmentation on the ScanRefer dataset. Grounded masks are shown in green.}
  \label{fig_refer}
\end{figure}

\subsection{3D Captioning}
From Table~\ref{tab_all_results}, our Uni3DL model outperforms existing methods in 3D captioning on the Cap3D Objaverse dataset, as evidenced by its superior BLEU-1 (B-1)~\cite{papineni2002bleu}, ROUGE-L(R)~\cite{lin2004rouge}, and METEOR (M)~\cite{banerjee2005meteor} scores. Specifically, on the BLEU-1 and ROUGE-L scores, our method beats precious STOA methods by a large margin (more than 20\%). Qualitative analyses, illustrated in Figure~\ref{fig_captioning}, demonstrate our caption predictions closely align with the ground truth. Additional qualitative results are presented in the Appendix \ref{supp_vis_cap}. We use this finetuned model to evaluate zero-shot 3D classification performance on ModelNet40/10 dataset and results are provided in \ref{supp_zeroshot_cls}.

\begin{figure}[!h]
  \centering
  \begin{subfigure}[t]{0.23\textwidth}
    \includegraphics[width=0.32\linewidth]{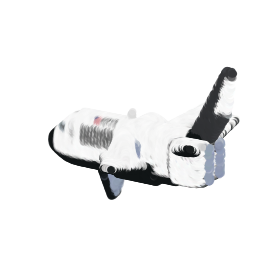}
    \includegraphics[width=0.32\linewidth]{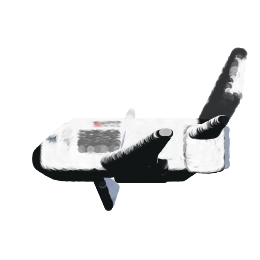}
    \includegraphics[width=0.32\linewidth]{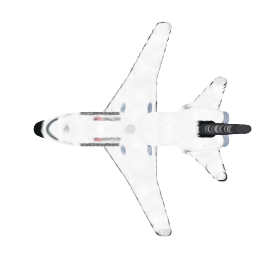}
    \caption*{
        \textcolor{blue}{\textit{GT: a small white nasa space shuttle airplane flying in the sky.}}
        \newline
        \textcolor{red}{Ours: a small white airplane flying in the air}
    }
  \end{subfigure}
  \hfill
  \begin{subfigure}[t]{0.23\textwidth}
    \includegraphics[width=0.32\linewidth]{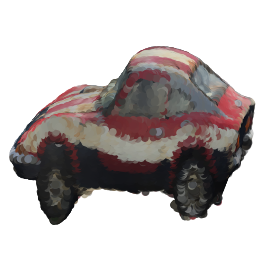}
    \includegraphics[width=0.32\linewidth]{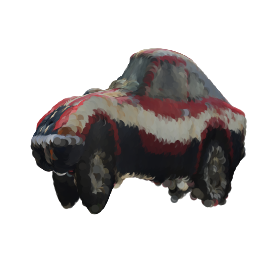}
    \includegraphics[width=0.32\linewidth]{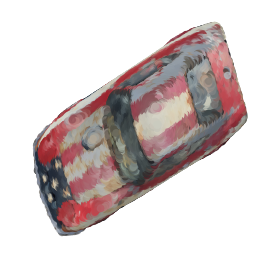}
    \caption*{
        \textcolor{blue}{\textit{GT: an old red and white car with an American flag painted on it.}}
        \newline
        \textcolor{red}{Ours: an old red and white race car with its rear paintings featuring stickers}
    }
  \end{subfigure}
  \hfill
  \begin{subfigure}[t]{0.23\textwidth}
    \includegraphics[width=0.32\linewidth]{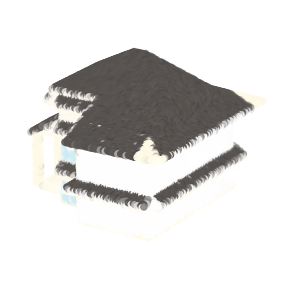}
    \includegraphics[width=0.32\linewidth]{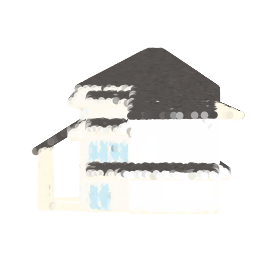}
    \includegraphics[width=0.32\linewidth]{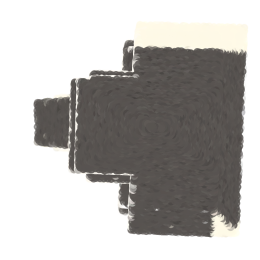}
    \caption*{
        \textcolor{blue}{\textit{GT: a white house with a roof.}}
        \newline
        \textcolor{red}{Ours: a white house with a roof and stairs}
    }
  \end{subfigure}
  \hfill
  \begin{subfigure}[t]{0.23\textwidth}
    \includegraphics[width=0.32\linewidth]{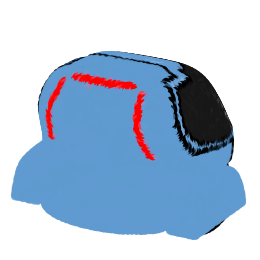}
    \includegraphics[width=0.32\linewidth]{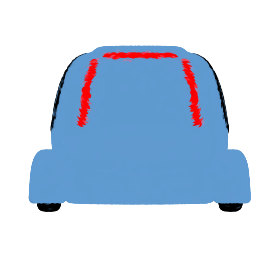}
    \includegraphics[width=0.32\linewidth]{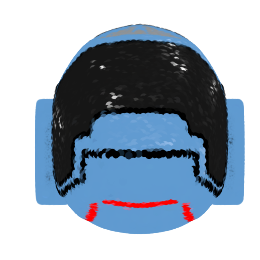}
    \caption*{
        \textcolor{blue}{\textit{GT: a small blue toy car with red accents and a helmet on top.}}
        \newline
        \textcolor{red}{Ours: a small blue toy vehicle, resembling a car with wheels}
    }
  \end{subfigure}
  \hfill
  \caption{3D captioning results on Cap3D Objaverse dataset.}
  \label{fig_captioning}
\end{figure}

\subsection{Text-to-3D Retrieval}
We evaluate text-to-3D retrieval performance on the Text2Shape ShapeNet subset. From Table~\ref{tab_all_results}, our Uni3DL model achieves comparable text-to-3D retrieval performance with STOA task-specific methods, including 
FTST~\cite{chen2019text2shape}, FMM~\cite{chen2019text2shape}, Y2S~\cite{han2019y2seq2seq}, and Parts2Words~\cite{tang2023parts2words}, as indicated by recall scores R@1 and R@5. For the Parts2Words method, we primarily compare its performance without using part information for a fair comparison. Qualitative results are provided in Appendix \ref{supp_fig_ret}.

\subsection{Ablation Study}

\noindent \textbf{Effect of Pretraining}. We assess the impact of pretraining on downstream tasks. Ablation experiments are conducted by training separate models from scratch for various tasks, including ScanNet (v2) semantic segmentation, S3DIS instance segmentation, ScanRefer grounded segmentation, and Text2Shape retrieval. As evidenced in Table~\ref{tab_finetune}, the pretraining stage significantly enhances performance across all downstream tasks. We show the qualitative comparison of the baseline model trained from scratch and our finetuned model on S3DIS instance segmentation in Figure~\ref{fig_seg}.

\noindent \textbf{Effect of pertaining tasks}. 
We further investigate the effect of each pertaining task, including instance/grounded segmentation, 3D captioning, and text-to-3D retrieval. In Table \ref{tab_multi}, we keep grounded segmentation while evaluating the significance of remaining pretraining tasks. From Table~\ref{tab_multi}, we have the following findings: 1) Instance segmentation benefits both grounded segmentation and text-3D cross-modal retrieval. Without instance segmentation task, the grounded segmentation Acc@0.25 drops from 37.8\% to 33.8\%. This is because the grounding task itself is based on instance identification. Instance segmentation also helps to better learn object-text alignment and benefits text-3D cross-modal retrieval. 2) Caption and retrieval benefit each other. Without pertaining on the captioning task, the text-3D cross-modal retrieval accuracy drops from 8.0\% 5o 3.5\% in terms of shape-to-text R@1 on the Cap3D retrieval task. Without pertaining on the retrieval task, the captioning performance drops from 18.6\% to 15.8\% in terms of ROUGE scores on the Cap3D captioning task.

\begin{table}[]
    \centering
    \resizebox{0.49\textwidth}{!}{
    \begin{tabular}{l|cccc}
        \toprule
        \multirow{3}{*}{Task} & \underline{Sem Seg} & \underline{Inst Seg} & \underline{Gnd Seg} & \underline{Ret} \\
         & SN Val & S3DIS (Area 5) & ScanRefer & Text2Shape \\
         & mIoU/mAcc & mAP\(_{50}\) / mAP\(_{25}\) & Acc@0.25/Acc@0.5 & R@1/R@5 \\
        \hline
        From scratch & 72.3/81.8 & 61.7/71.7 & 33.8/31.4 & 2.4/7.7 \\
        Ours & \textbf{76.2}/\textbf{84.8} & \textbf{65.3}/\textbf{74.3} & \textbf{39.4}/\textbf{36.4} & \textbf{4.6}/\textbf{18.0} \\
         \bottomrule
    \end{tabular}}
    \caption{Ablation of pertaining.}
    \label{tab_finetune}
\end{table}

\begin{table}[]
    \centering
    \resizebox{0.48\textwidth}{!}{
    \begin{tabular}{l|ccc}
        \toprule
        \multirow{3}{*}{Task} & \underline{Gnd Seg} & \underline{Captioning} & \underline{Retrieval}\\
         & ScanRefer & Cap3D & Cap3D\\
         & Acc@0.25/Acc@0.5 & B-1/R & T2S R@1/S2T R@1\\
        \hline
        Ours & \underline{37.8}/34.2 & 15.4/\textbf{18.6} & \textbf{5.5}/\textbf{8.0}\\
        \hline
        \; - Inst Seg & 33.8/31.3 & \textbf{20.9}/\underline{17.8} & 3.0/\underline{4.0} \\ 
        \; - Retrieval & 37.7/\underline{34.5} & \underline{19.6}/15.8 & n/a \\ 
        \; - Captioning & \textbf{37.9}/\textbf{35.9} & n/a & \underline{5.0}/3.5 \\ 
         \bottomrule
    \end{tabular}}
    \caption{Ablation of pertaining tasks. T2S for Text-to-Shape retrieval, and S2T for Shape-to-Text retrieval.}
    \label{tab_multi}
\end{table}

\section{Conclusion}

In this work, we introduce a unified model named Uni3DL for generalized 3D vision and language understanding tasks. We design a query transformer module to attentively align 3D features with latent and textural queries. A task router module with multiple functional heads is faithfully designed to support diverse vision-language tasks, including 3D object classification, 3D semantic/instance segmentation, 3D object detection, 3D grounded segmentation, 3D captioning, and text-3D cross-modal retrieval. Experiments on multiple benchmark datasets show comparable or even superior performance of our Uni3DL model compared to the previous STOA method. 

\section{Acknowledgement}
We extend our sincere gratitude to Xueyan Zou from the University of Wisconsin-Madison for the helpful and insightful discussions that contributed to our work.

{\small
\bibliographystyle{ieeenat_fullname}
\bibliography{main}
}


\setcounter{section}{1}
\maketitlesupplementary

\appendix


\section{Experimental Settings}\label{supp_settings}

\subsection{Pretraining}\label{supp_pretraining}

As described in the main paper, we use the ScanNet (v2), ScanRefer, and Cap3D Objaverse datasets for joint pretraining. For the Cap3D Objaverse caption dataset, we only include objects whose captions contain any object name from the ScanNet, S3DIS, or ModelNet categories. The Uni3DL model is pretrained for 50 epochs. We set the initial learning rate to 1e-4 and reduce it by 0.1 after 50\% and 70\% of the total training steps. A linear warmup is applied for the first 10 iterations.

\subsection{Finetuning}\label{supp_finetuning}
\noindent \textbf{Finetuning for 3D semantic/instance Segmentation}. For the S3DIS dataset, we randomly crop $5m\times5m\times5m$ blocks from each scene, ensuring a minimum of 25,000 points per scene. The Uni3DL model is finetuned for 25 epochs with an initial learning rate of 2e-5, which is multiplied by 0.1 after 50\% and 70\% of the total training steps. For ScanNet Segmentation, we finetune our Uni3DL model for 30 epochs on ScanNet semantic/instance segmentation with the same learning rate strategy as in S3DIS segmentation.

Current state-of-the-art instance segmentation methods, including Mask3D and Mask-Att-Free, use additional segment labels obtained from an unsupervised graph-based segmentation method~\cite{felzenszwalb2004efficient} during training and evaluation. To ensure a fair comparison, we report the performance of our Uni3DL model using segment information.

\noindent \textbf{Finetuning for Grounded Segmentation}. For 20 epochs, we finetune the Uni3DL model on Grounded Segmentation with an initial learning rate of 1e-5, decaying it by 0.1 after reaching 50\% and 70\% of the total training steps.

\noindent \textbf{Finetuning for 3D Captioning}. The Uni3DL model is finetuned for 30 epochs on the Cap3D Objaverse dataset. The learning rate starts at 1e-4 and is reduced by 0.2 after 50\% and 70\% of the training steps.

\noindent \textbf{Finetuning for Text-3D Cross-Modal Retrieval}. We finetune the Uni3DL model for 30 epochs on the Text2Shape retrieval task, following a similar learning rate strategy as in 3D Captioning.
For data augmentation, we applied random scaling to the training shapes, using a scale factor uniformly sampled from the range [0.8, 1.2]. Additionally, we randomly rotated the shapes along the z-axis, selecting rotation angles within the range $[-\pi/2, \pi/2]$.

\section{More quantative results}

\subsection{Zero-Shot 3D Classification}\label{supp_zeroshot_cls}
We use our Uni3DL model fine-tuned on the Cap3D Objaverse dataset to evaluate zero-shot 3D classification performance on ModelNet40 and ModelNet10 datasets. ModelNet40 includes 40 different categories with 12, 311 CAD models, while ModelNet10, a smaller subset, consists of 10 categories with 4, 899 models. We use the same validation set as~\cite{wu20153d} for performance evaluation.

Table~\ref{tab_zeroshot_cls} summarizes the performance on ModelNet10 and ModelNet40 test datasets. From this Table, we can see that our method achieves a competitive performance on both datasets, with a classification accuracy of 70.4\% on ModelNet10 and 57.0\% on ModelNet40. Specifically, our Uni3DL trained achieves the best top-5 classification accuracy. \textit{It should be noted that all compared methods rely on projecting 3D data to multiview 2D images and use a pretrained CLIP for image-text alignment; while our method does not require view projection}.

\begin{table*}[!ht]
\centering
\resizebox{0.7\textwidth}{!}{%
\begin{tabular}{lcccccc}
\toprule
\multirow{2}{*}{Method}          & \multirow{2}{*}{Input} & \multirow{2}{*}{Pretraing dataset} & \multirow{2}{*}{Pretrained FM} & ModelNet10 & \multicolumn{2}{c}{ModelNet40} \\
\cline{6-7}
& & & & top-1 & top-1 & top-5 \\ 
\midrule
PointCLIP\cite{zhang2022pointclip}  & MV Images & ShapeNet   & Yes (CLIP)        & 30.2      & 23.8   & -   \\
CLIP2Point\cite{huang2023clip2point} & MV Images & ShapeNet  & Yes (CLIP)   & 66.6      & 49.4   & -   \\
PointCLIP V2\cite{zhu2023pointclip}    & MV Images & ShapeNet   & Yes (CLIP+GPT3)     & \textbf{73.1}      & \underline{64.2}   & -   \\
ULIP~\cite{xue2023ulip} & MV Images & ShapeNet & Yes (CLIP) & -  & 60.4 & \underline{84.0} \\
ULIP~\cite{xue2023ulip} & MV Images & Cap3D Objaverse & Yes (CLIP) & -  & \textbf{67.2} & 83.1 \\
\midrule
Ours & Point Cloud & Cap3D Objaverse & No & \underline{70.4} & 57.0 & \textbf{88.8} \\
\bottomrule
\end{tabular}
}
\caption{Zero-shot 3D shape classification performance on ModelNet10 and ModelNet40 datasets. We show input types, pretrained datasets, and foundation model (FM) requirements for detailed comparison. Our method does not require projected multiview images as inputs and does not require pretrained foundation models. The results highlighted in \textbf{bold} and \underline{underline} denote the best and second-best performance, respectively.}
\label{tab_zeroshot_cls}
\end{table*}

\begin{table*}[!ht]
\centering
\resizebox{0.6\textwidth}{!}{
\begin{tabular}{lcccc}
\toprule
\multirow{2}{*}{Model} & \multirow{2}{*}{Single Stage} & \multirow{2}{*}{Detector} & \multicolumn{2}{c}{\underline{Overall}} \\
 & & & Acc@0.25 & Acc@0.5 \\ 
\midrule
ScanRefer \cite{chen2020scanrefer} & \xmark & VoteNet & 39.0 & 26.1 \\
InstanceRefer \cite{yuan2021instancerefer} & \xmark & PointGroup & 38.2 & 31.4 \\
3DVG-Transformer \cite{zhao20213dvg} & \xmark & VoteNet & 45.9 & 34.5 \\
3DJCG~\cite{cai20223djcg} & \xmark & VoteNet & \textbf{47.6} & 36.1 \\
D3Net~\cite{chen2022d3net} & \xmark & PointGroup & - & 35.6 \\
UniT3D~\cite{chen2023unit3d} & \xmark & PointGroup & - & 36.5 \\
M3DRef~\cite{zhang2023multi3drefer} & \xmark & PointGroup &  - & \textbf{40.4} \\
\midrule
TGNN~\cite{huang2021text} & \cmark & N/A & 37.4 & 29.7 \\
Uni3DL (Ours) & \cmark & N/A & \textbf{37.8} & \textbf{33.7} \\
\bottomrule
\end{tabular}
}
\caption{Comparative analysis of grounded localization performance on the ScanRefer~\cite{chen2020scanrefer} dataset. We report the ratios of correctly predicted bounding boxes with IoU thresholds of 0.25 and 0.5. We report the performance of all comparing methods with only 3D point clouds as inputs.}
\label{supp_scanref}
\end{table*}

\subsection{Grounded Localization}\label{supp_grounded_loc}
In the main paper, we report the performance of our Uni3DL model for grounded \textit{segmentation}. Previous methods have also explored the grounded \textit{localization} task. To produce grounded object location, we directly use grounded object masks to calculate their bounding boxes.
Table~\ref{supp_scanref} summarizes the performance of Uni3DL and previous state-of-the-art methods for grounded localization. It should be noted that all compared methods except TGNN~\cite{huang2021text} employ a dual-stage process, where a 3D object detector identifies potential bounding box candidates, followed by a disambiguation module employed to fuse visual and textural features and determine the precise target bounding box. \textit{In contrast, our Uni3DL model is a single-stage model, without using second-stage object-text fusion modules.} Specifically, our Uni3DL model achieves better performance than another single-stage model TGNN~\cite{huang2021text} which also generates bounding boxes from object segmentation masks.



\section{More qualitative results}
\subsection{3D Captioning}\label{supp_vis_cap}
We show more qualitative results of 3D captioning on the Cap3D objaverse dataset in Figure~\ref{fig_captioning}. As shown in the figure, our Uni3DL can generate text descriptions well aligned with ground truth captions.

\begin{figure}[!h]
  \centering
  \begin{subfigure}[t]{0.23\textwidth}
    \includegraphics[width=0.32\linewidth]{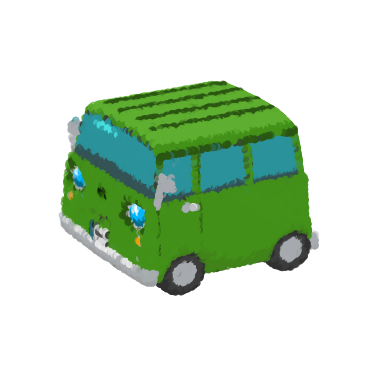}
    \includegraphics[width=0.32\linewidth]{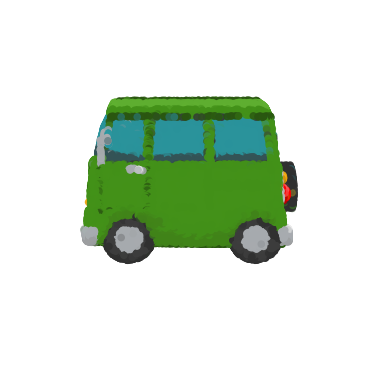}
    \includegraphics[width=0.32\linewidth]{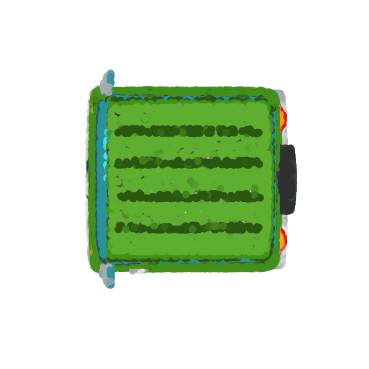}
    \caption*{
        \textcolor{blue}{\textit{GT: a small green cartoon car with blue eyes.}}
        \newline
        \textcolor{red}{Ours: a small green cartoon car with blue eyes}
    }
  \end{subfigure}
  \hfill
  \begin{subfigure}[t]{0.23\textwidth}
    \includegraphics[width=0.32\linewidth]{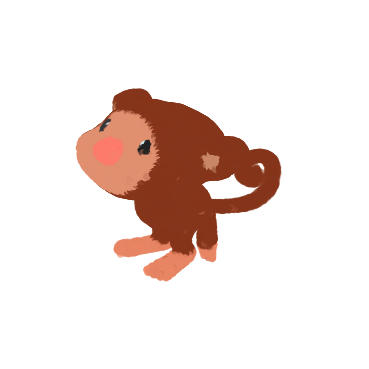}
    \includegraphics[width=0.32\linewidth]{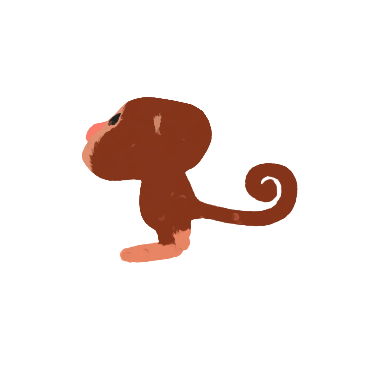}
    \includegraphics[width=0.32\linewidth]{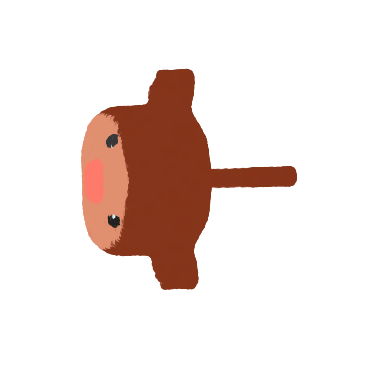}
    \caption*{
        \textcolor{blue}{\textit{GT: a cartoon monkey with a tail.}}
        \newline
        \textcolor{red}{Ours: a cartoon monkey with a tail}
    }
  \end{subfigure}
  \hfill
  \begin{subfigure}[t]{0.23\textwidth}
    \includegraphics[width=0.32\linewidth]{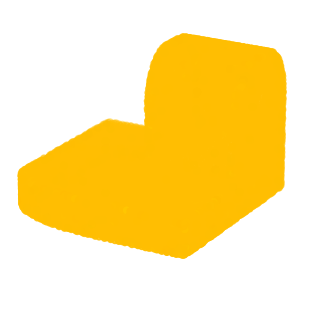}
    \includegraphics[width=0.32\linewidth]{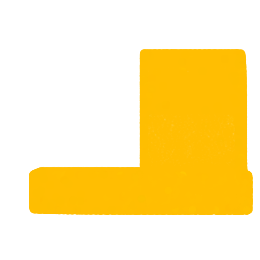}
    \includegraphics[width=0.32\linewidth]{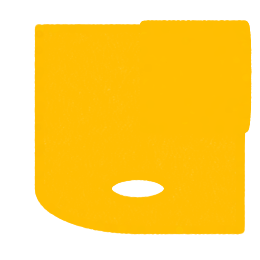}
    \caption*{
        \textcolor{blue}{\textit{GT: a yellow plastic block with two holes.}}
        \newline
        \textcolor{red}{Ours: a yellow plastic box with a hole, resembling a wooden bench}
    }
  \end{subfigure}
  \hfill
  \begin{subfigure}[t]{0.23\textwidth}
    \includegraphics[width=0.32\linewidth]{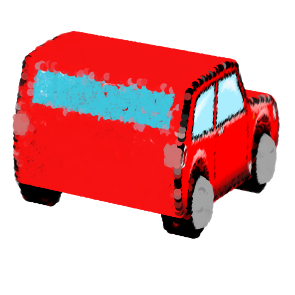}
    \includegraphics[width=0.32\linewidth]{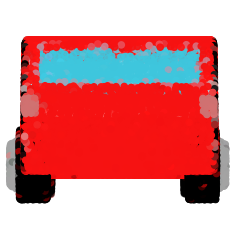}
    \includegraphics[width=0.32\linewidth]{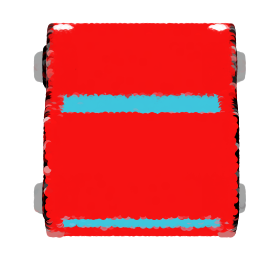}
    \caption*{
        \textcolor{blue}{\textit{GT: a small red car.}}
        \newline
        \textcolor{red}{Ours: a small red car}
    }
  \end{subfigure}
  \hfill

  \caption{3D captioning results on Cap3D Objaverse dataset.}
  \label{fig_captioning}
\end{figure}

\subsection{3D Segmentation}\label{supp_vis_seg}
We show more instance segmentation results on both S3DIS and ScanNet validation set in Figure~\ref{fig_seg}. From the figure, we can see that our Uni3DL model produces satisfying results for both semantic and instance segmentation tasks. 

\begin{figure*}[ht]
  \centering
  \begin{minipage}[t]{0.19\textwidth}
    \includegraphics[width=\linewidth]{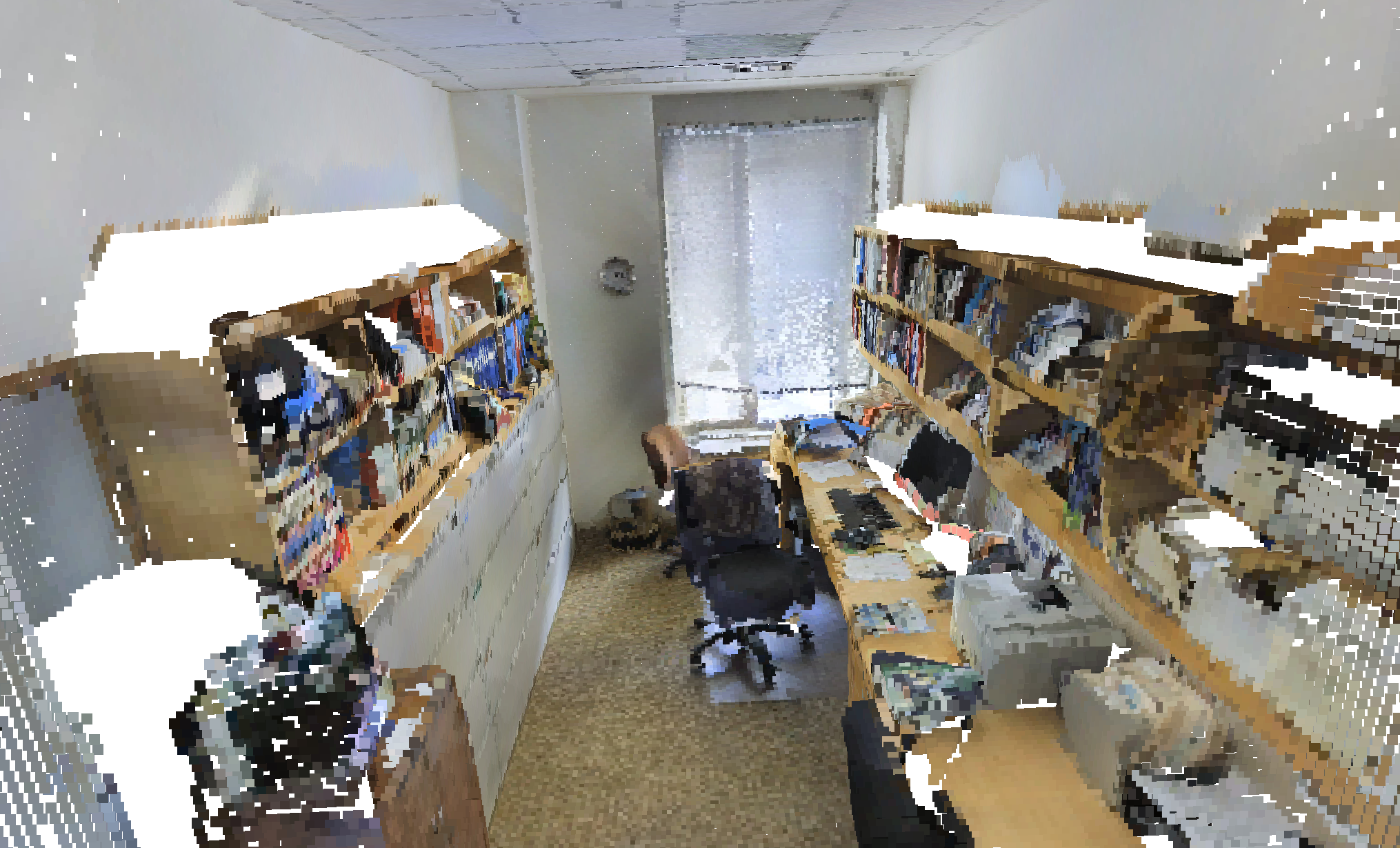}
    \includegraphics[width=\linewidth]{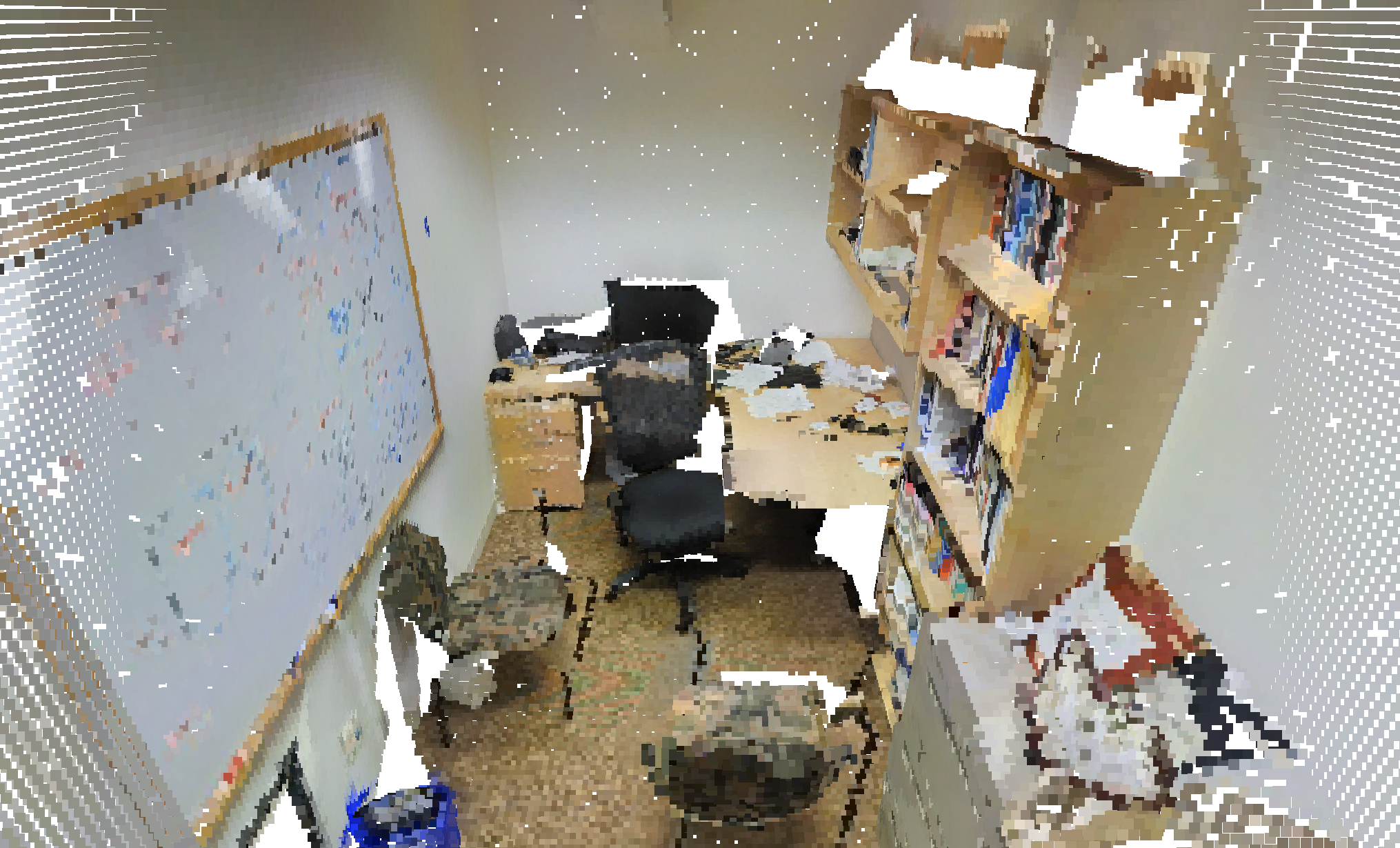}
    \includegraphics[width=\linewidth]{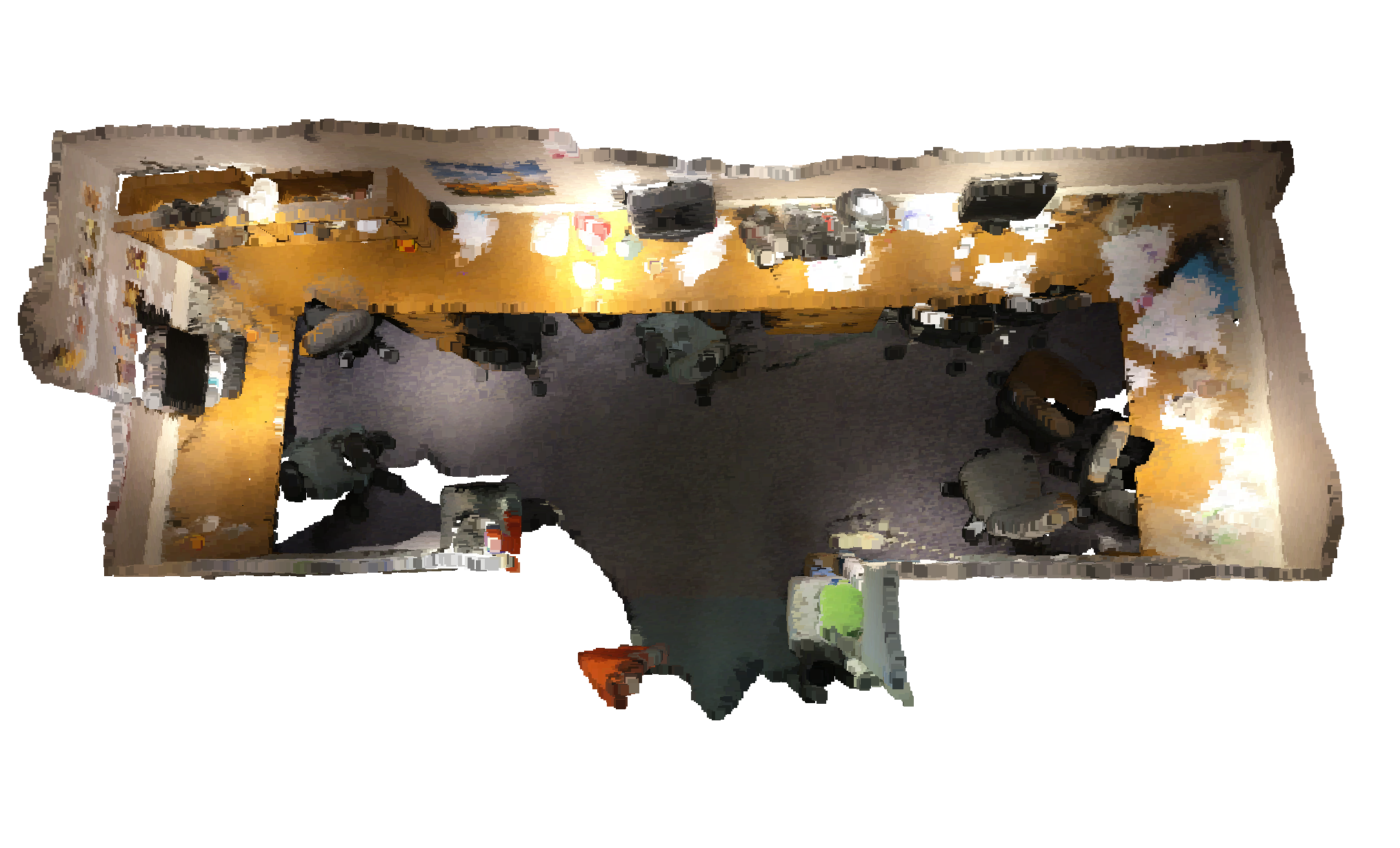}
    \includegraphics[width=\linewidth]{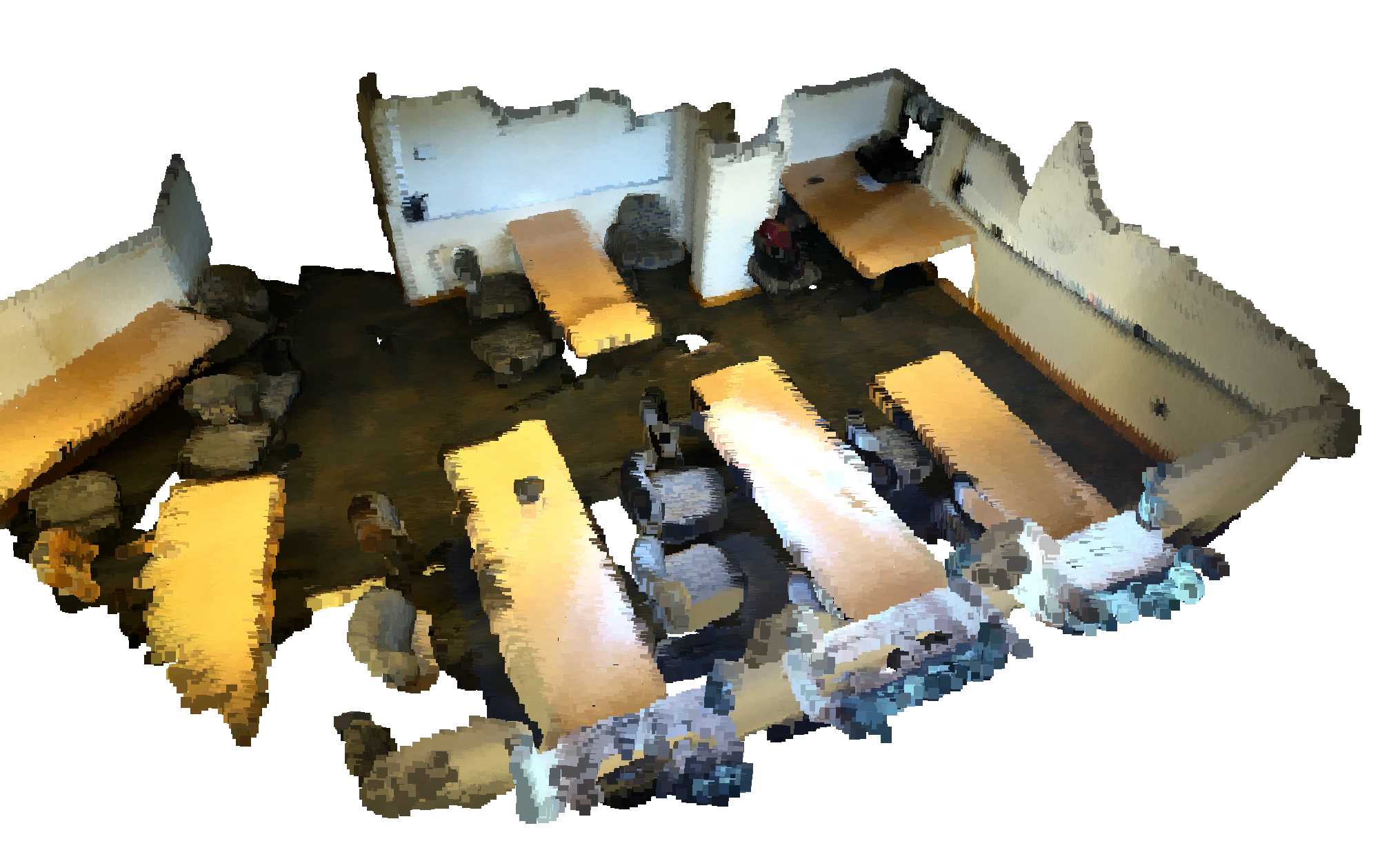}
    \caption*{Input}
  \end{minipage}
  \hfill
  \begin{minipage}[t]{0.19\textwidth}
    \includegraphics[width=\linewidth]{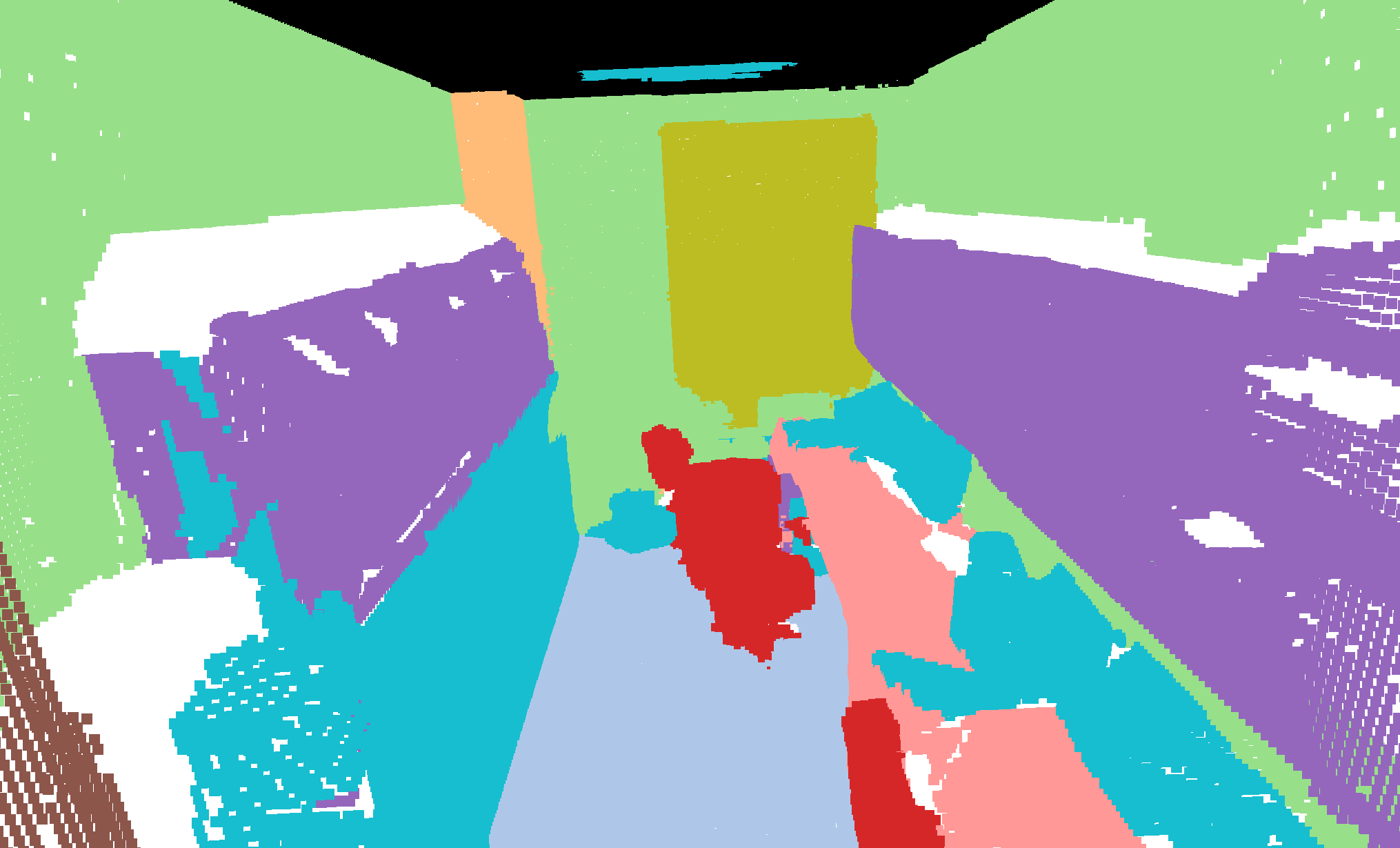}
    \includegraphics[width=\linewidth]{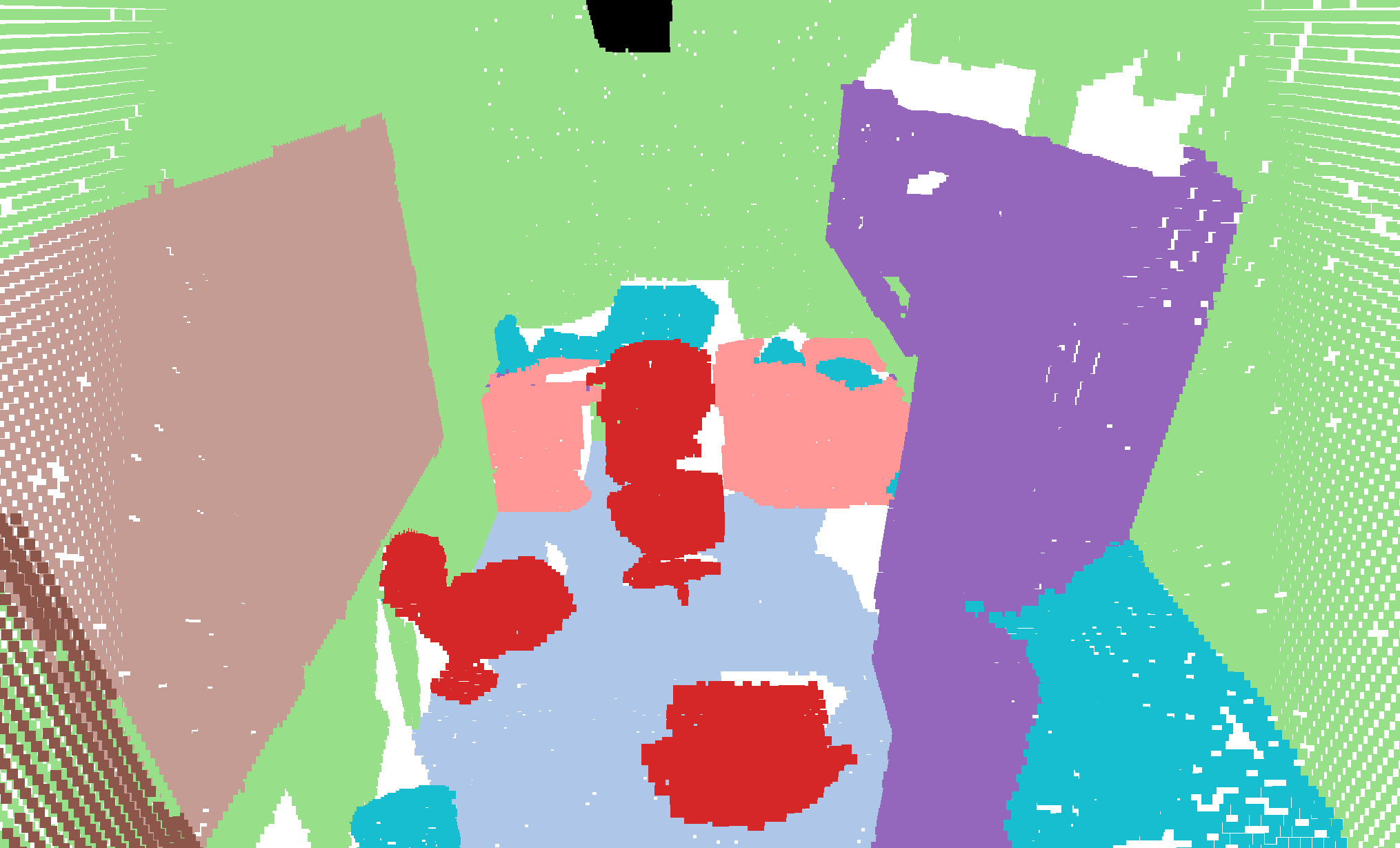}
    \includegraphics[width=\linewidth]{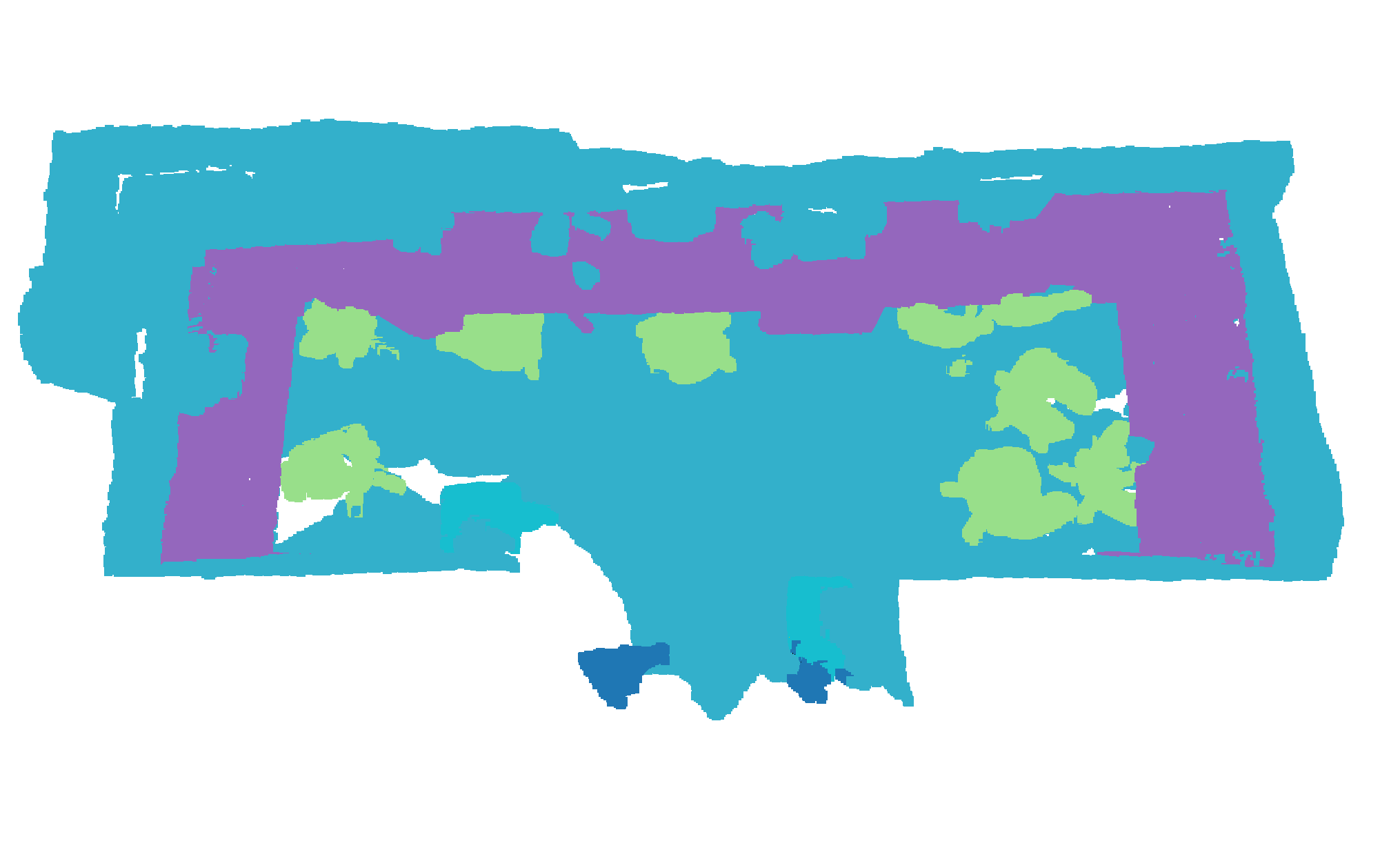}
    \includegraphics[width=\linewidth]{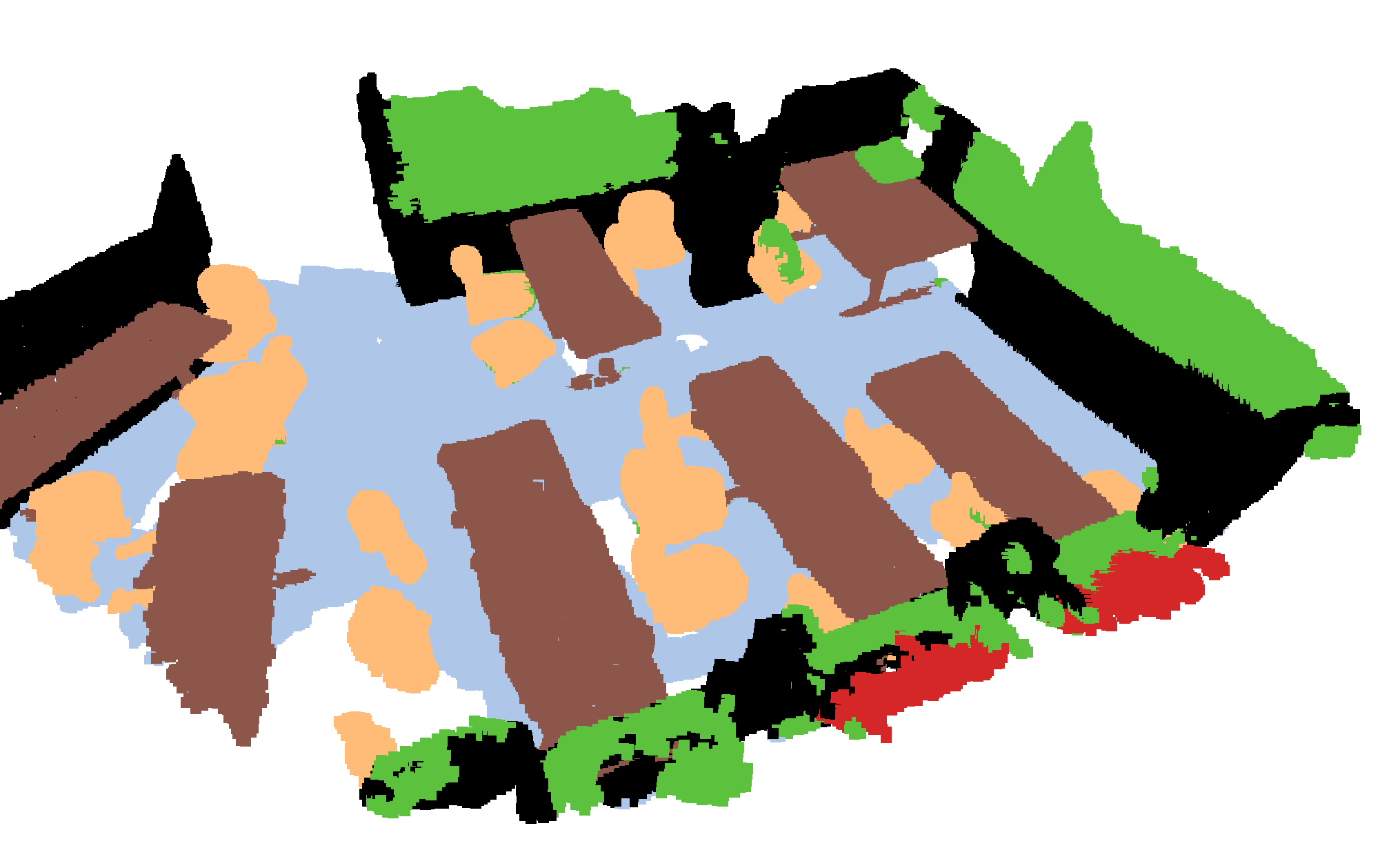}
    \caption*{Ours Sem.}
  \end{minipage}
  \hfill
  \begin{minipage}[t]{0.19\textwidth}
    \includegraphics[width=\linewidth]{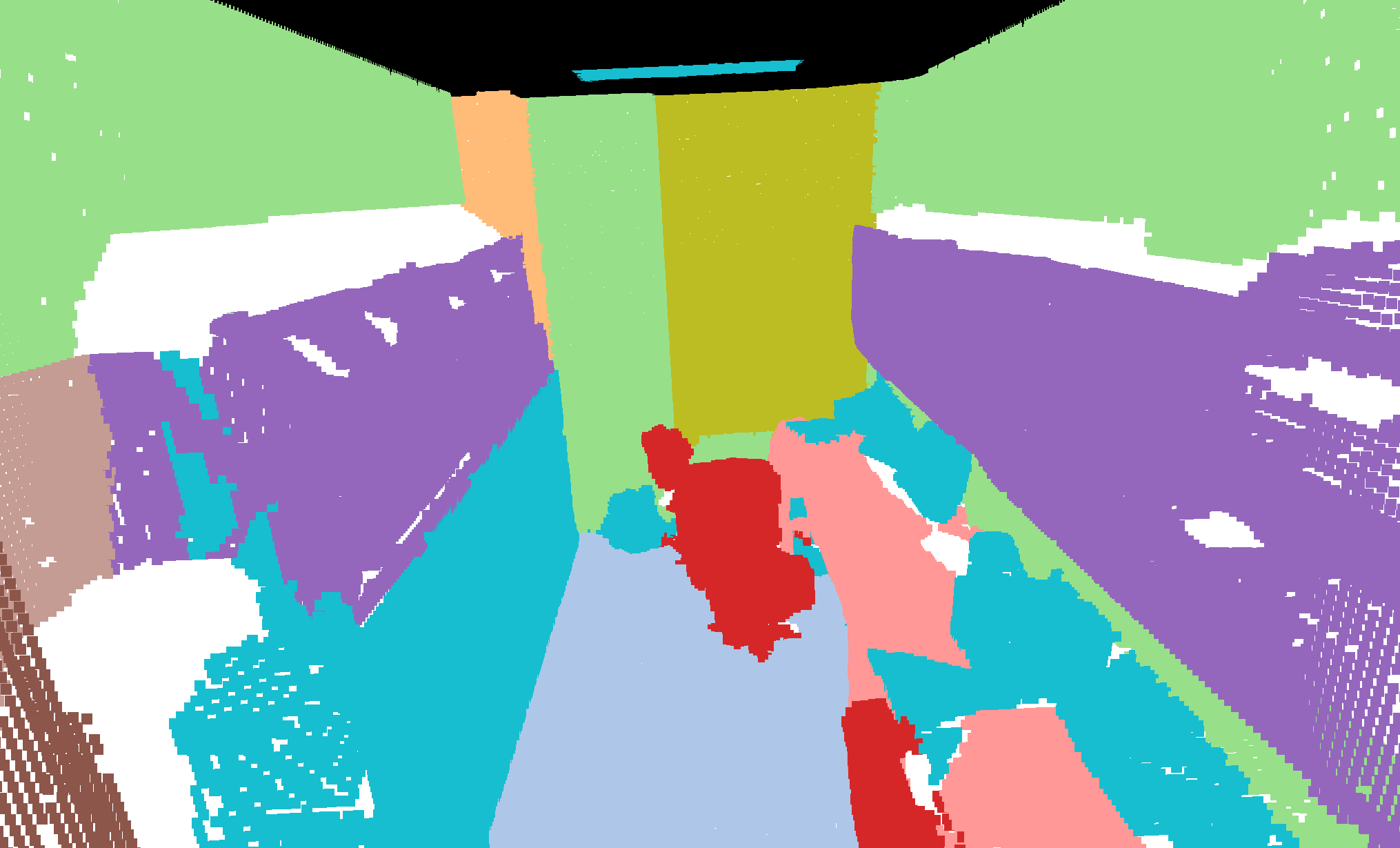}
    \includegraphics[width=\linewidth]{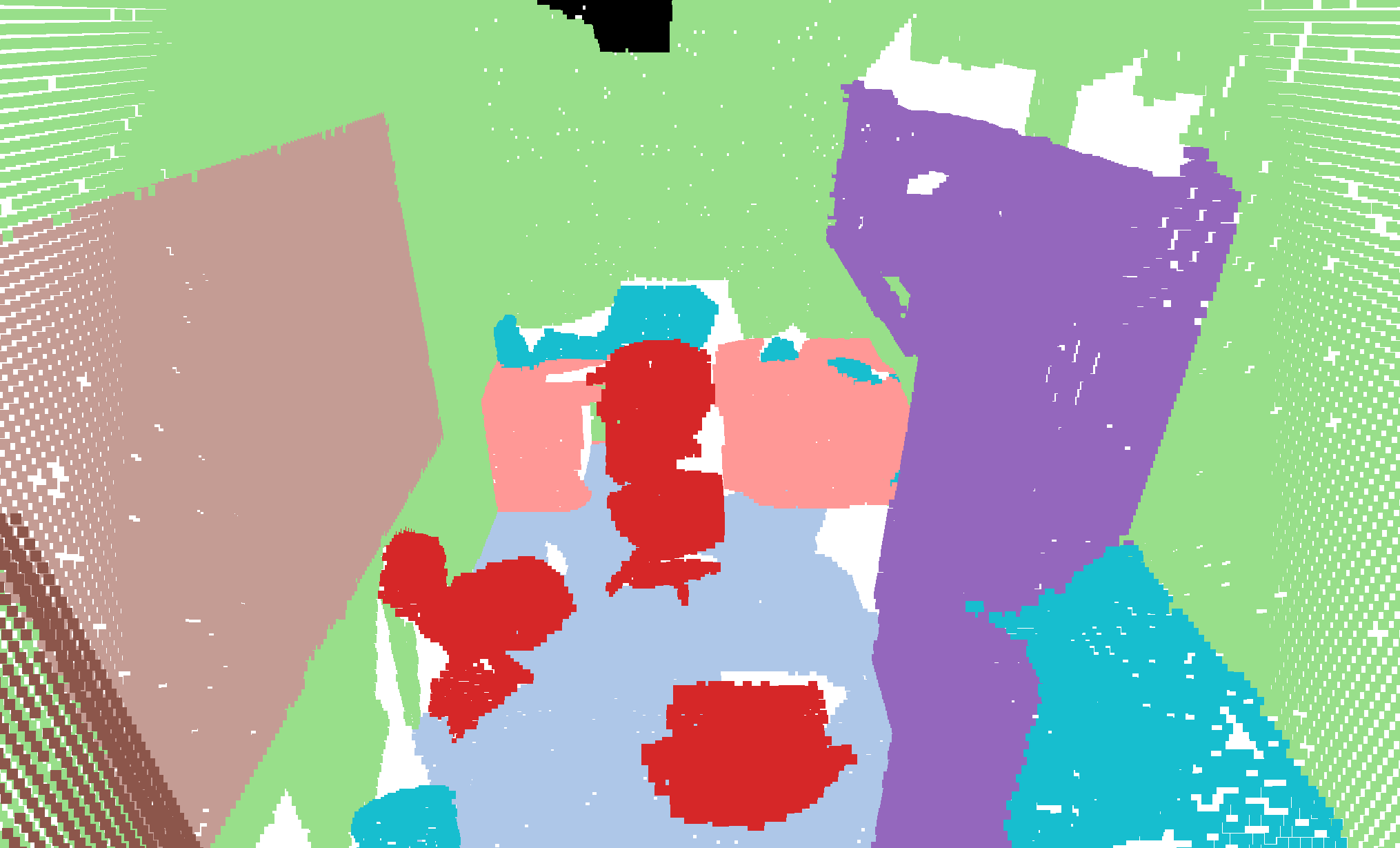}
    \includegraphics[width=\linewidth]{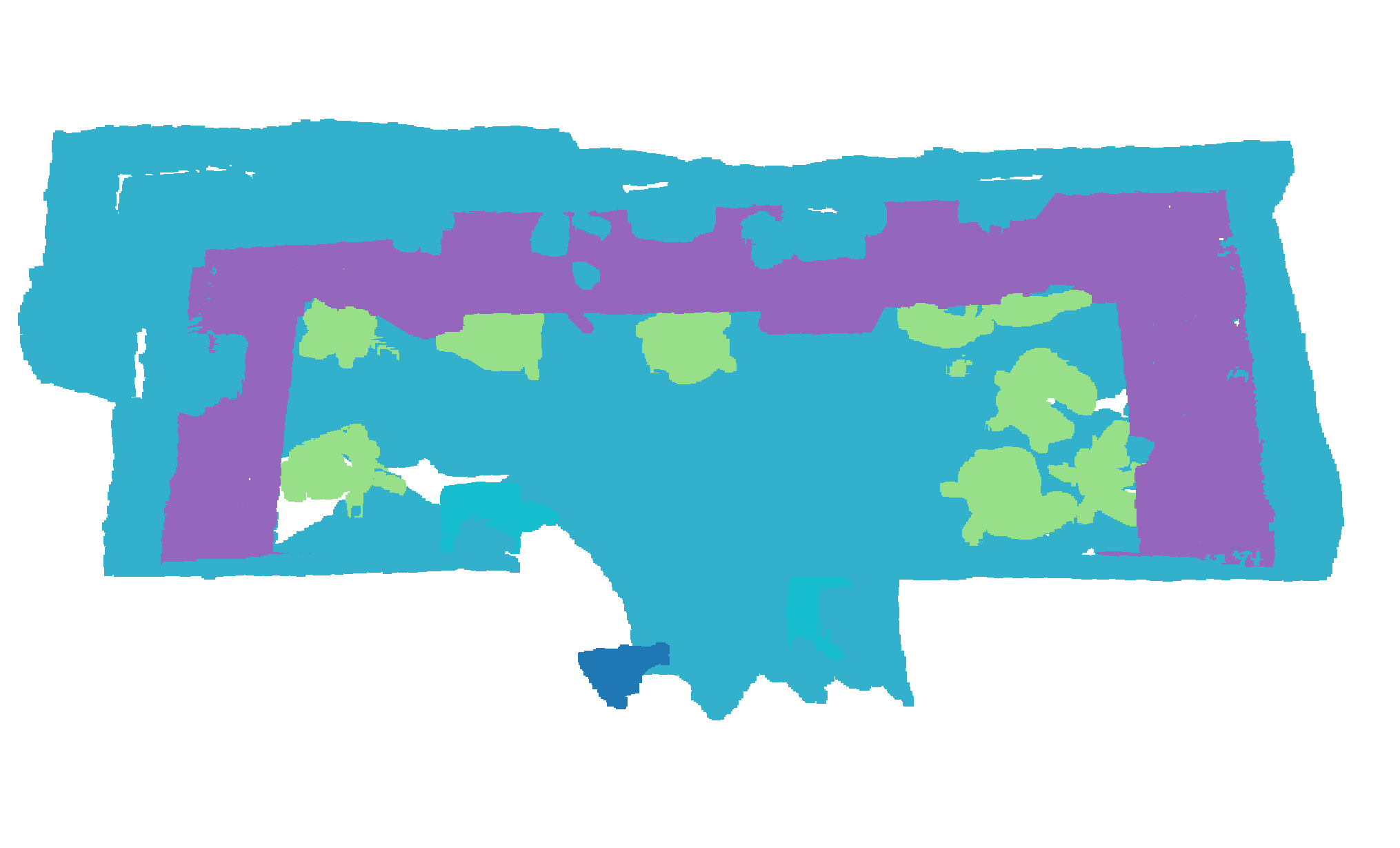}
    \includegraphics[width=\linewidth]{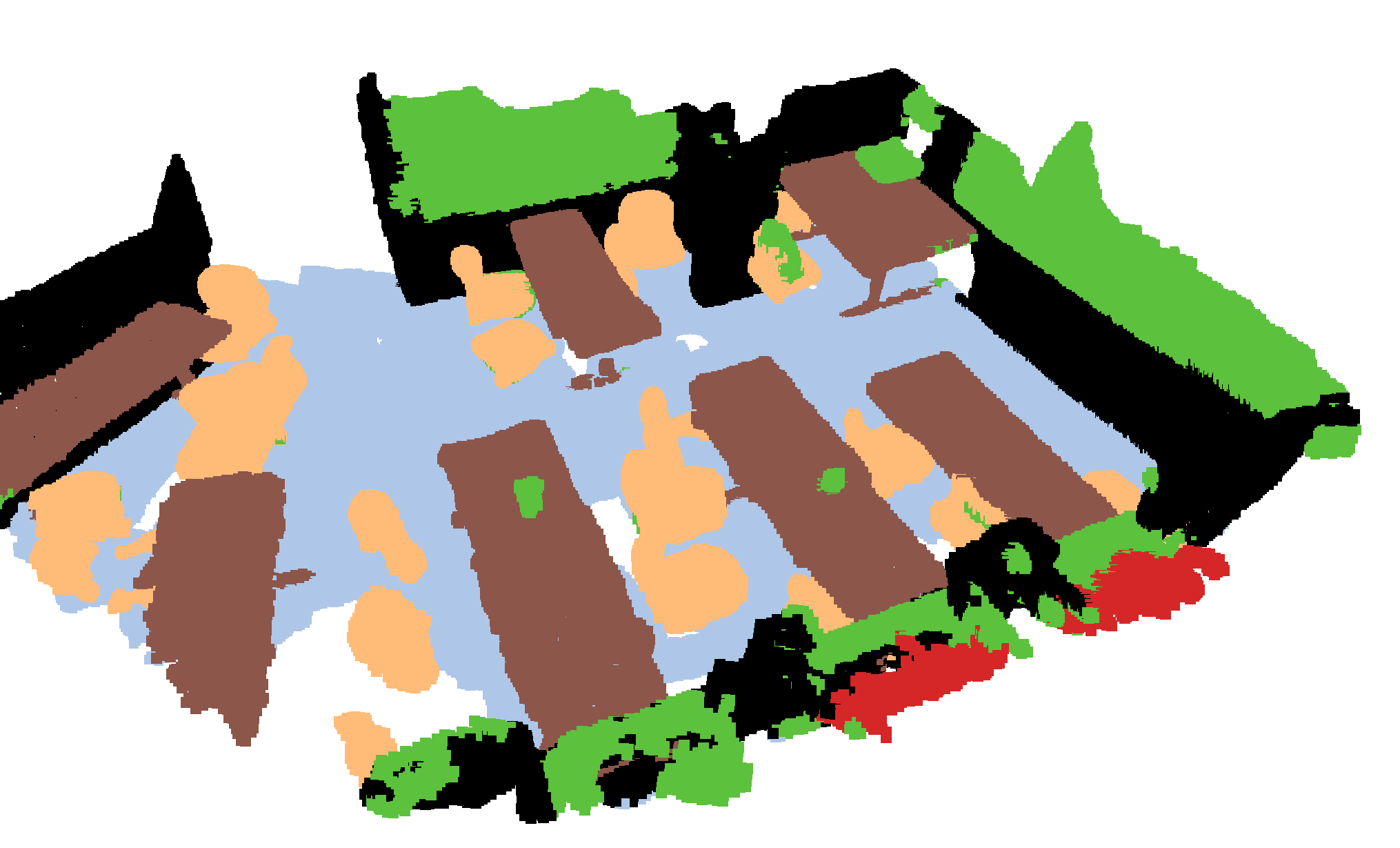}
    \caption*{GT Sem.}
  \end{minipage}
  \hfill
  \begin{minipage}[t]{0.19\textwidth}
    \includegraphics[width=\linewidth]{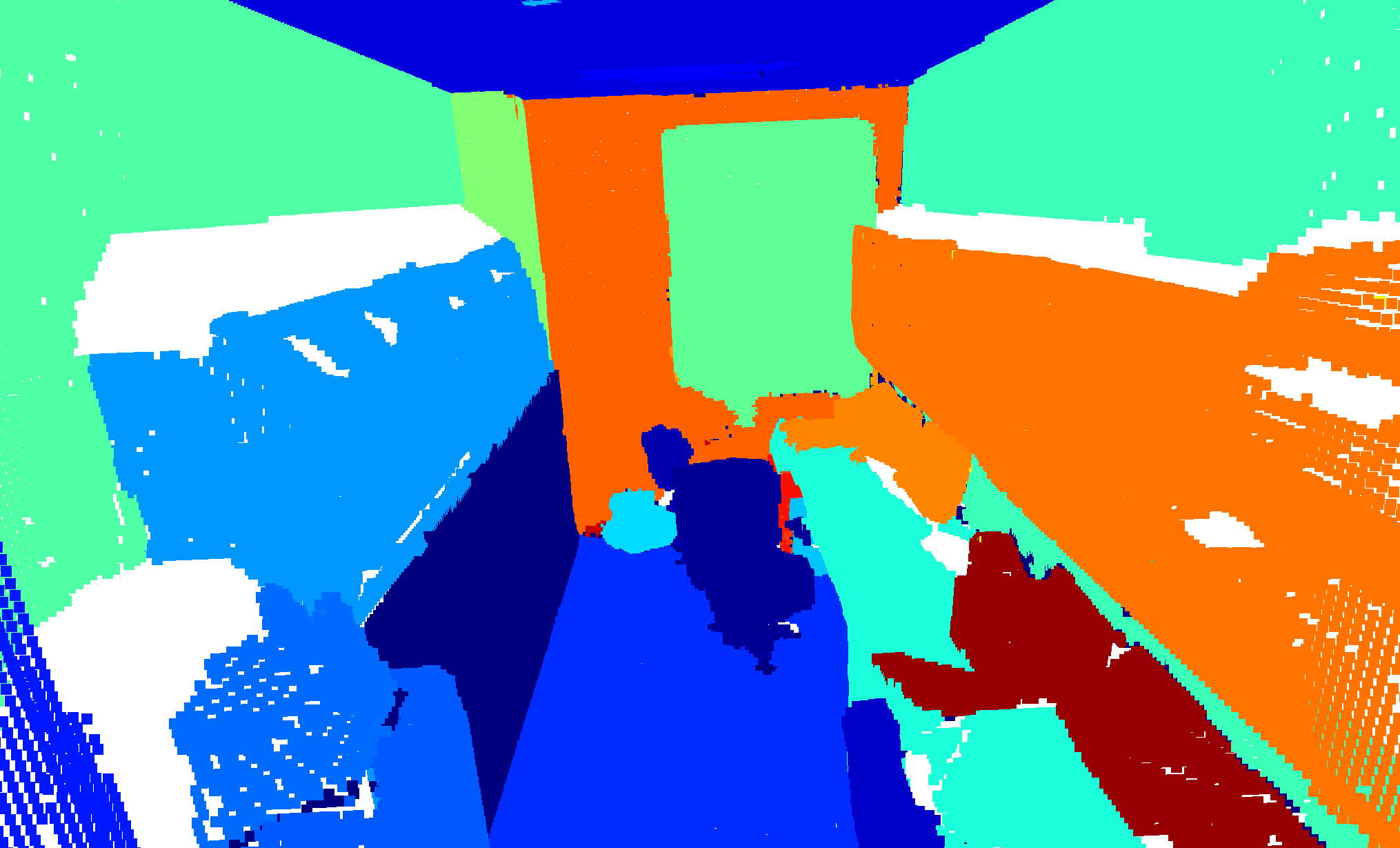}
    \includegraphics[width=\linewidth]{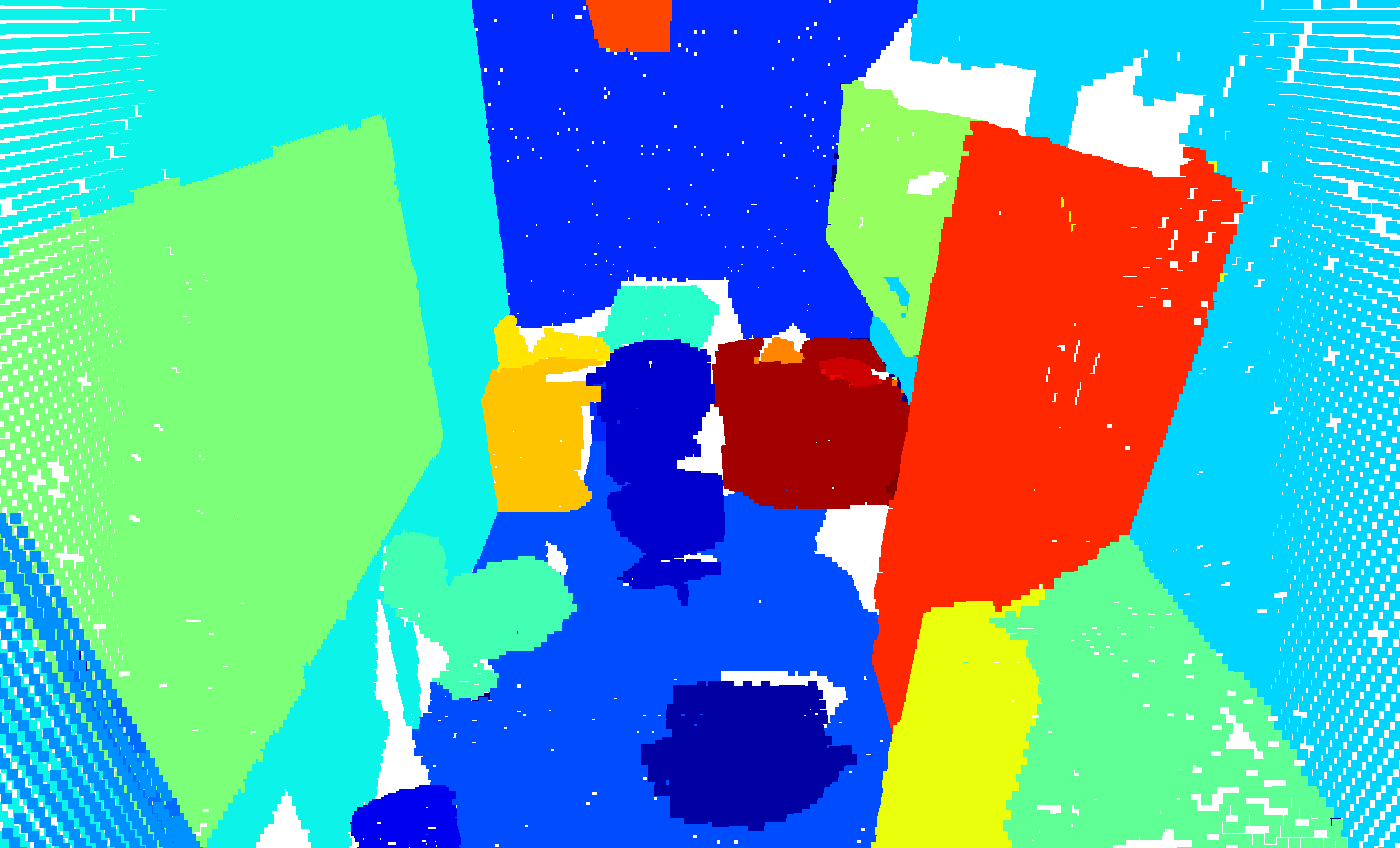}
    \includegraphics[width=\linewidth]{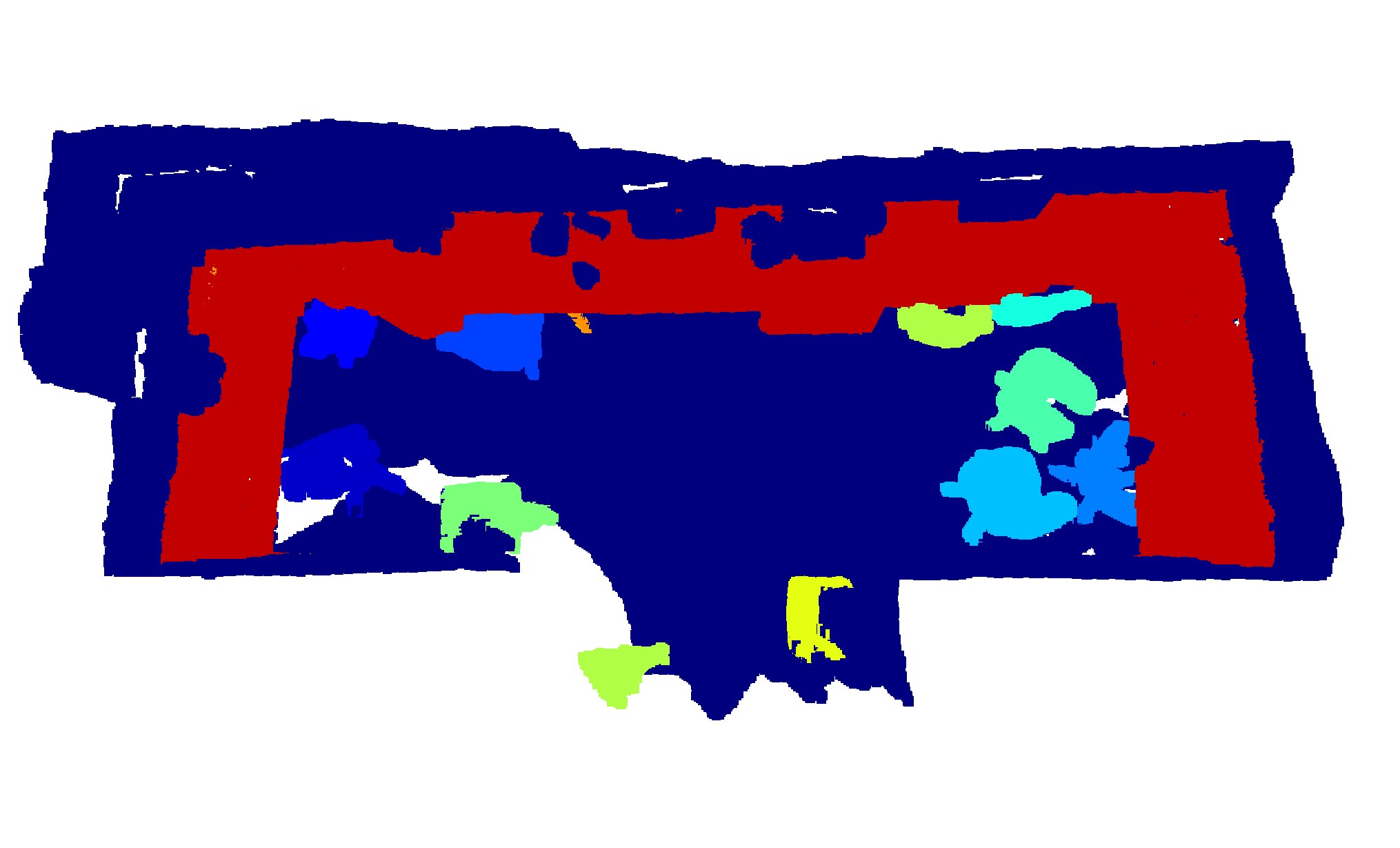}
    \includegraphics[width=\linewidth]{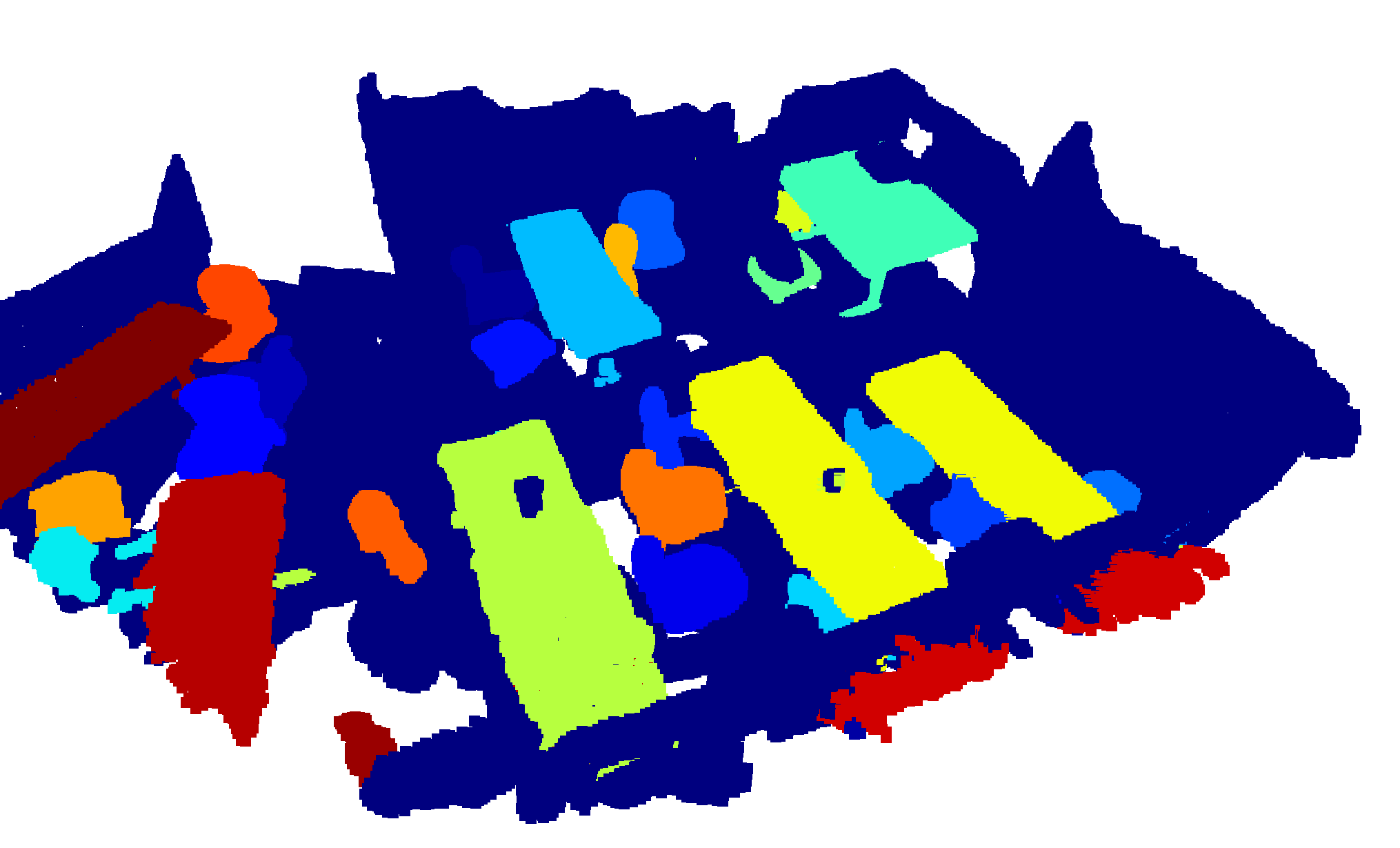}
    \caption*{Ours Inst.}
  \end{minipage}
  \hfill
  \begin{minipage}[t]{0.19\textwidth}
    \includegraphics[width=\linewidth]{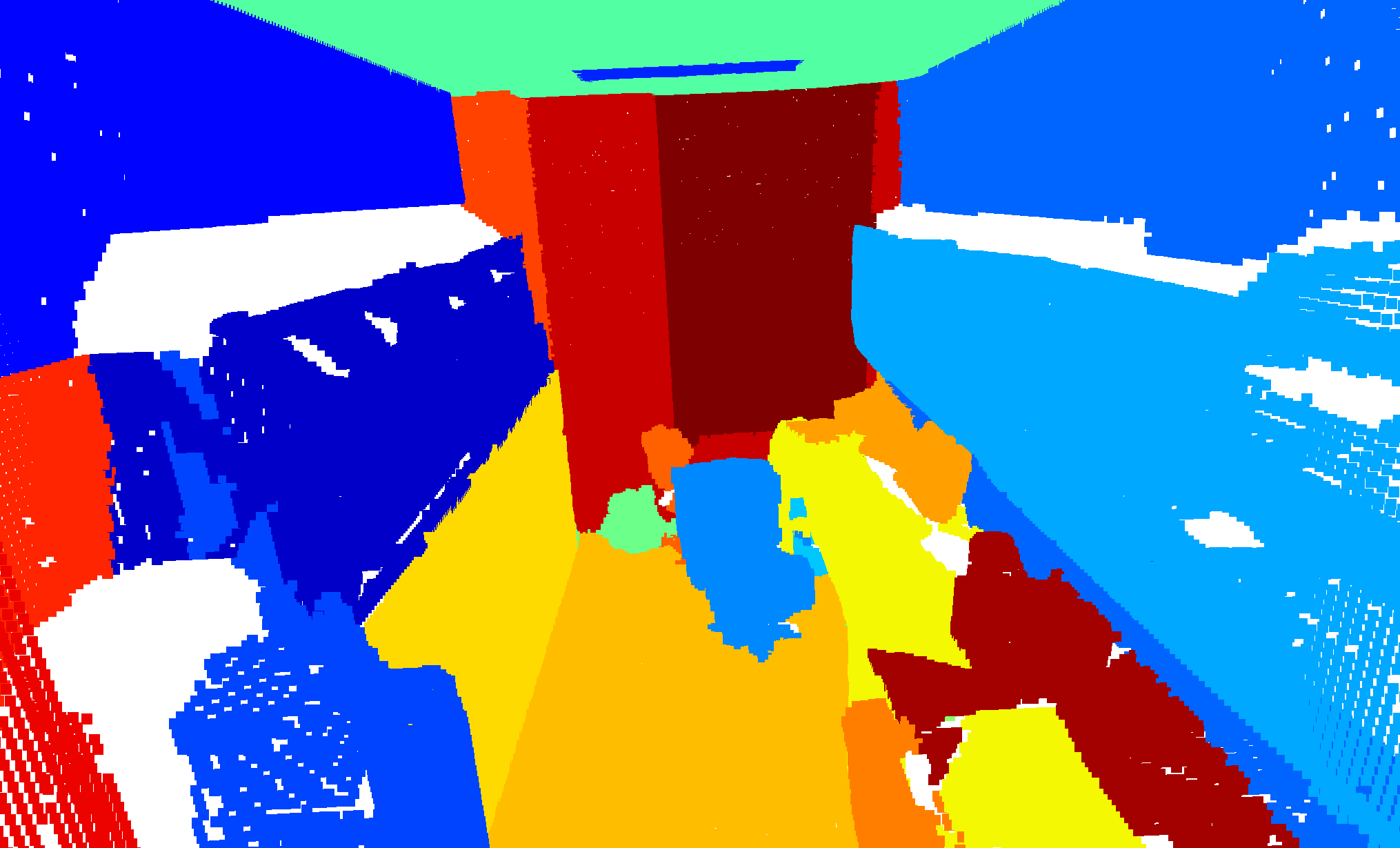}
    \includegraphics[width=\linewidth]{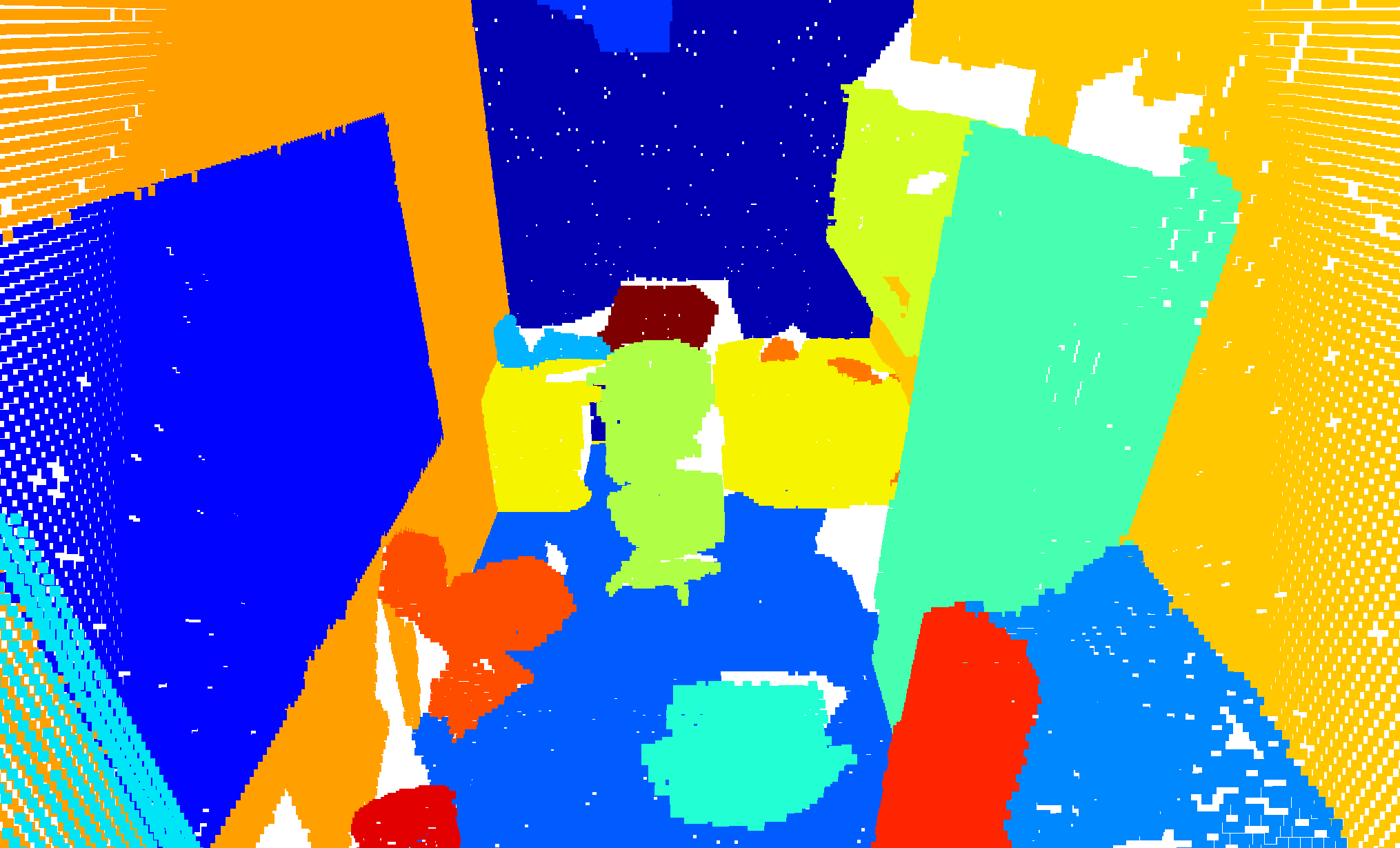}
    \includegraphics[width=\linewidth]{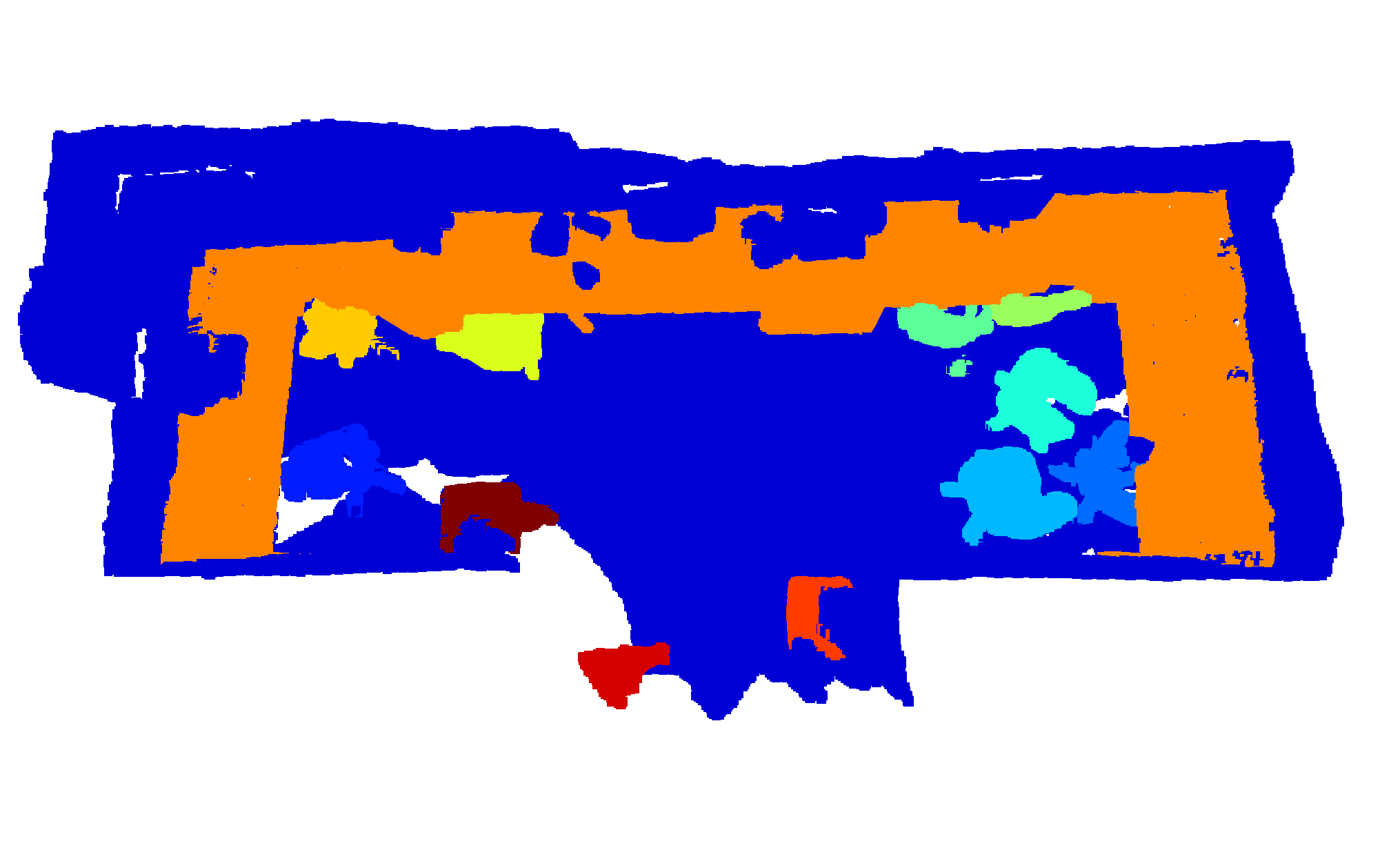}
    \includegraphics[width=\linewidth]{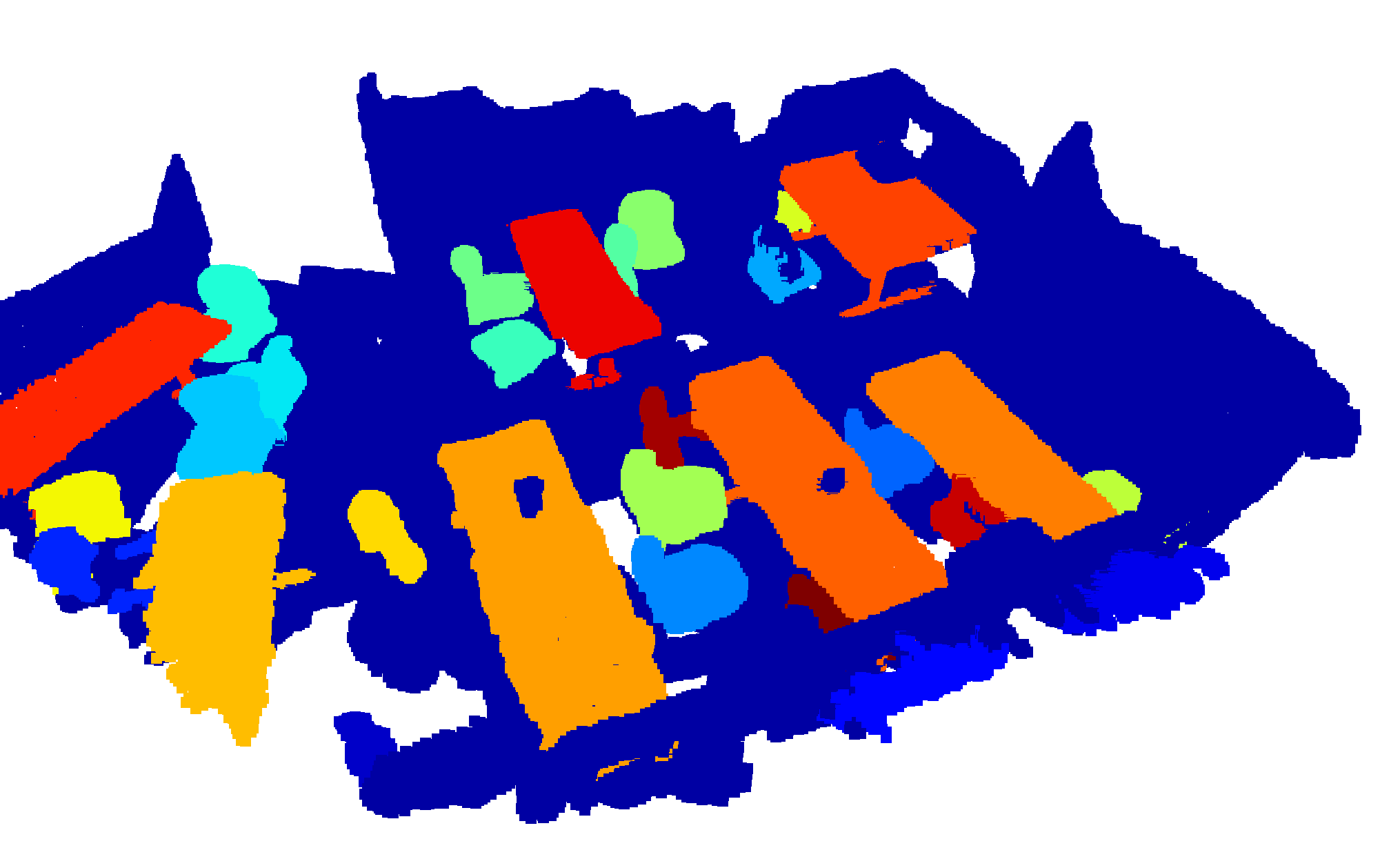}
    \caption*{GT Inst.}
  \end{minipage}
  \hfill
  \caption{3D Segmentation results on S3DIS (top) and ScanNet (bottom) datasets.}
  \label{fig_seg}
\end{figure*}

\subsection{Grounded Segmentation}\label{supp_vg_seg}
\begin{figure*}[!h]
    \centering
    \begin{subfigure}[t]{0.49\textwidth}
        \includegraphics[width=0.32\linewidth]{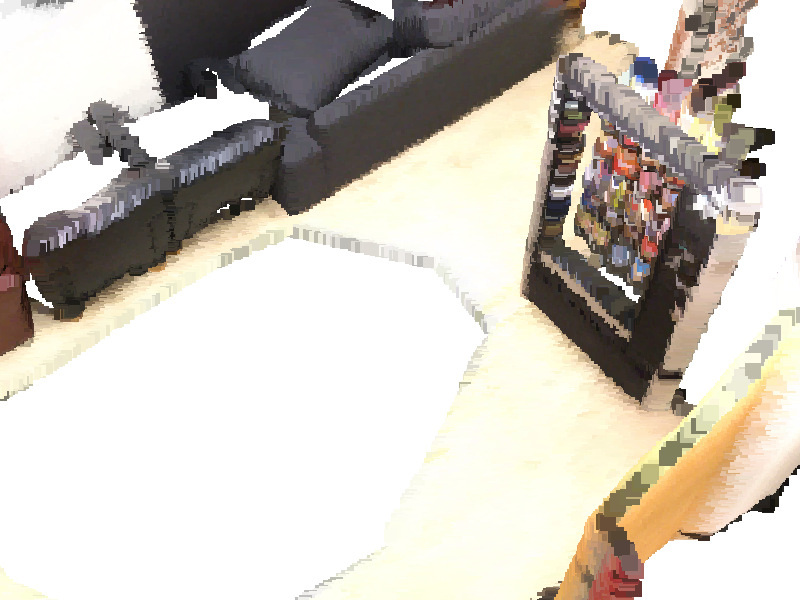}
        \hfill
        \includegraphics[width=0.32\linewidth]{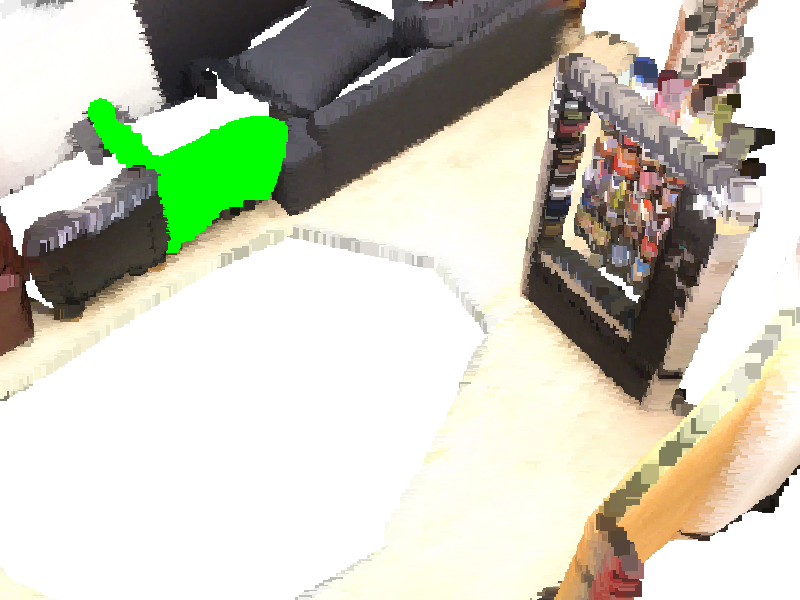}
        \hfill
        \includegraphics[width=0.32\linewidth]{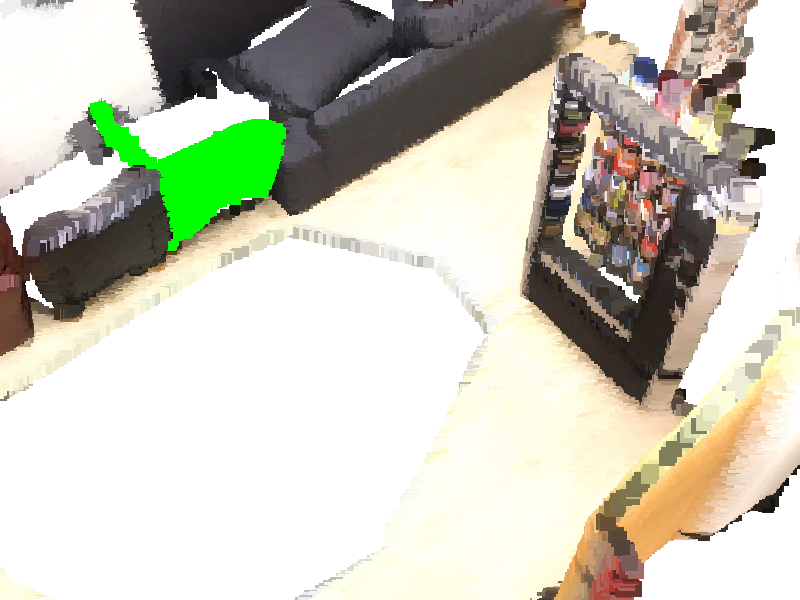}
        \caption*{Input\hspace{2.5cm}GT\hspace{2.5cm}Ours}
        \caption*{this black chair is next to the black couch. it appears to be leather. it is black. there is a snack machine on the opposite wall.}
    \end{subfigure}
    \hfill
    \vline
    \hfill
    \begin{subfigure}[t]{0.49\textwidth}
        \includegraphics[width=0.32\linewidth]{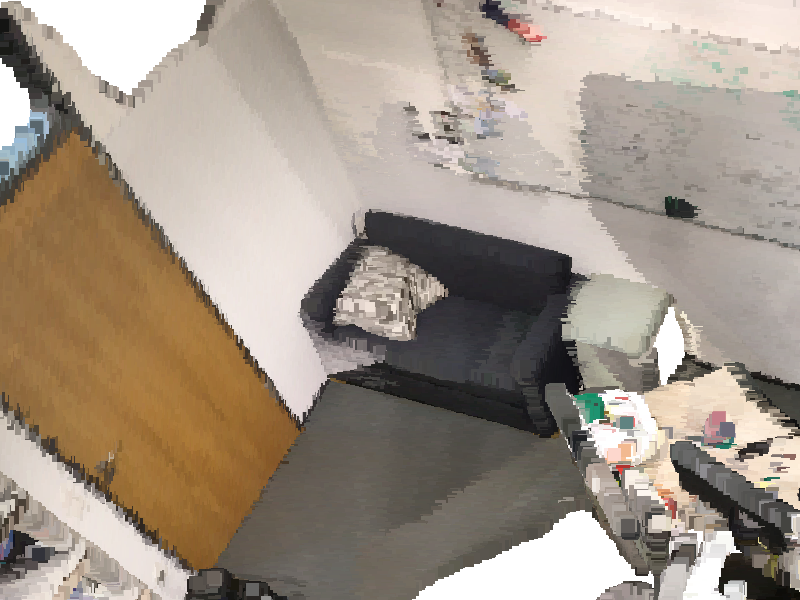}
        \hfill
        \includegraphics[width=0.32\linewidth]{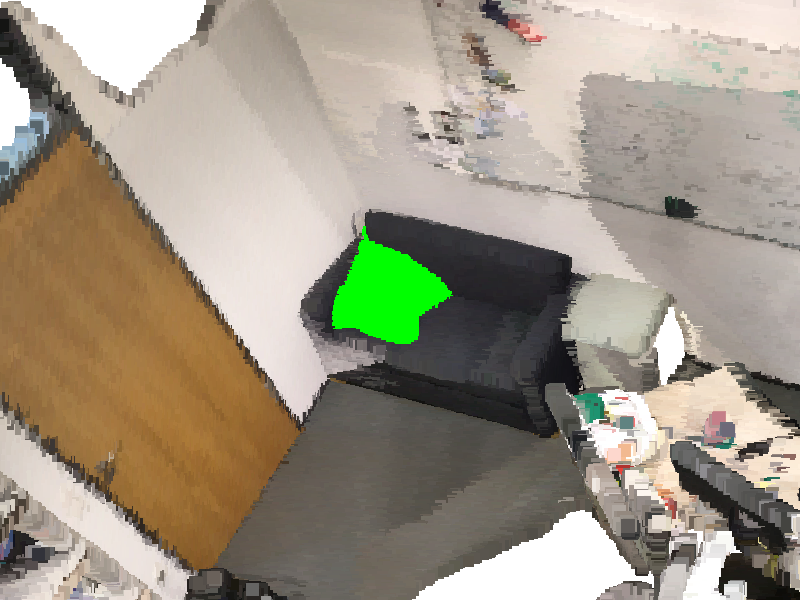}
        \hfill
        \includegraphics[width=0.32\linewidth]{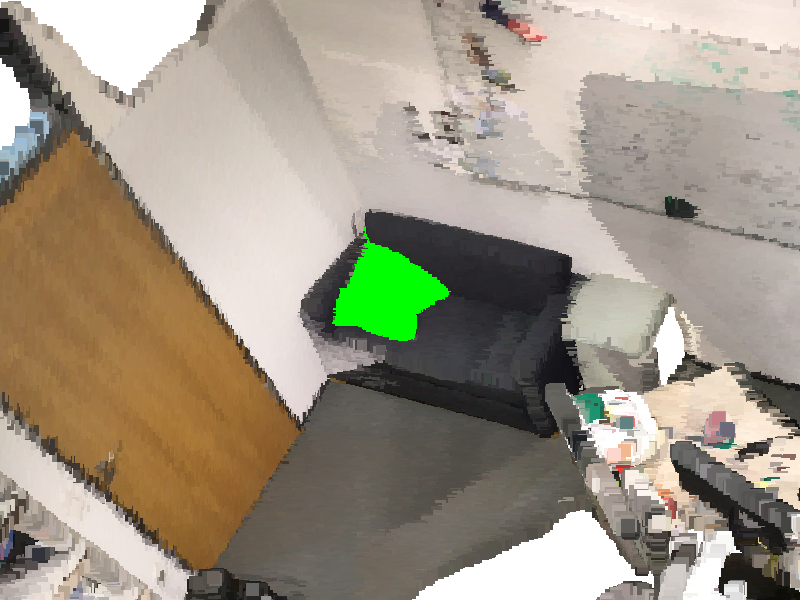}
        \caption*{Input\hspace{2.5cm}GT\hspace{2.5cm}Ours}
        \caption*{it is a small pillow located on the coutch.  you can notice it directly on your left when walking through the door into the room.}
    \end{subfigure}
    \hfill
    \begin{subfigure}[t]{0.49\textwidth}
        \includegraphics[width=0.32\linewidth]{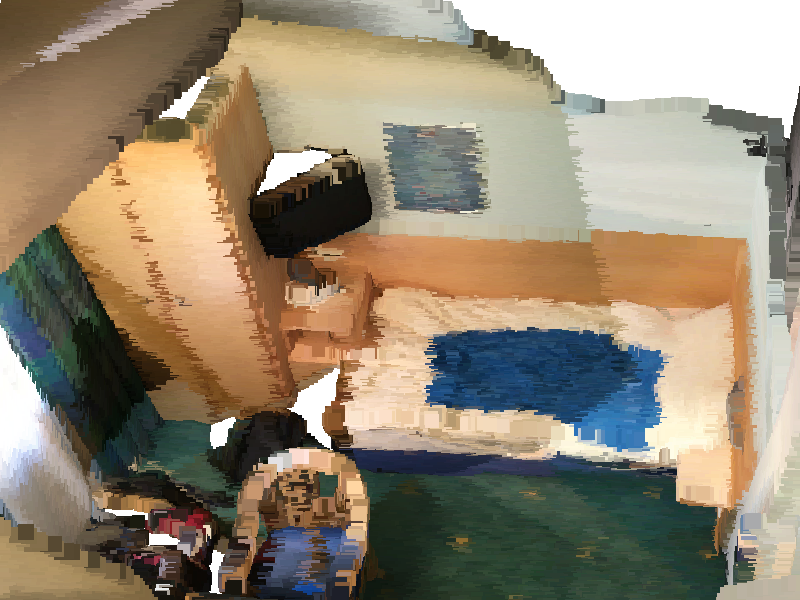}
        \hfill
        \includegraphics[width=0.32\linewidth]{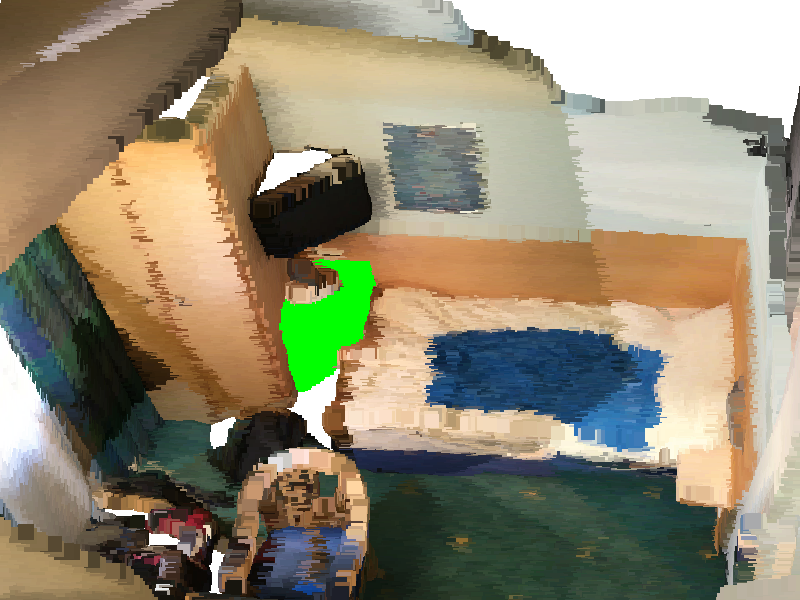}
        \hfill
        \includegraphics[width=0.32\linewidth]{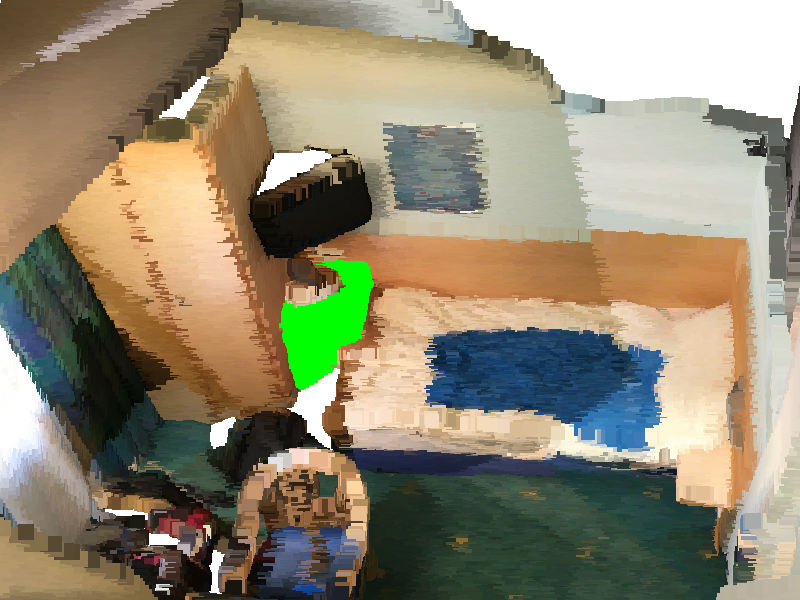}
        \caption*{Input\hspace{2.5cm}GT\hspace{2.5cm}Ours}
        \caption*{a brown wooden nightstand.  it's between the end of the bed and close to the wall.}
    \end{subfigure}
    \hfill
    \vline
    \hfill
    \begin{subfigure}[t]{0.49\textwidth}
        \includegraphics[width=0.32\linewidth]{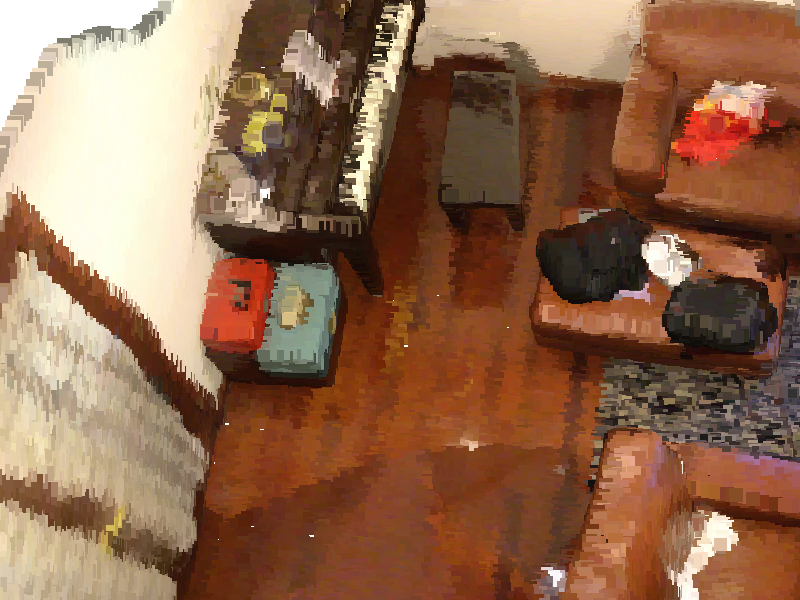}
        \hfill
        \includegraphics[width=0.32\linewidth]{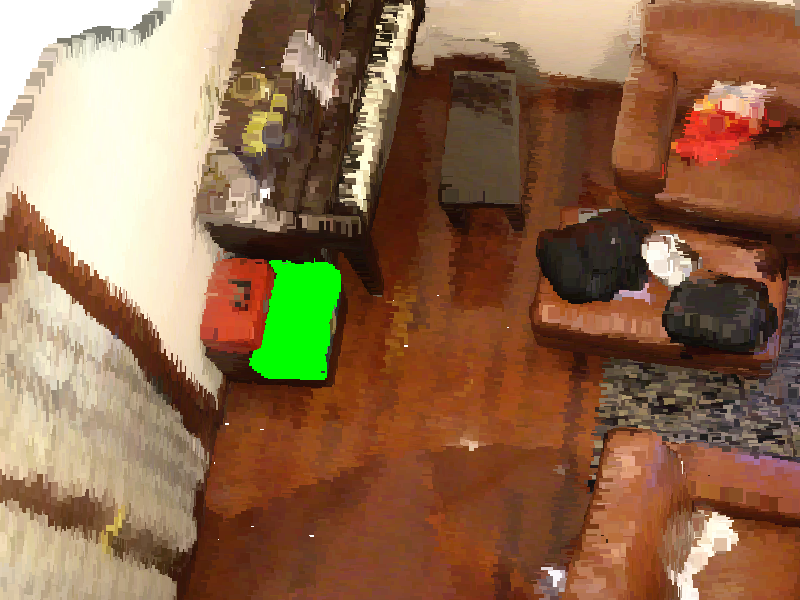}
        \hfill
        \includegraphics[width=0.32\linewidth]{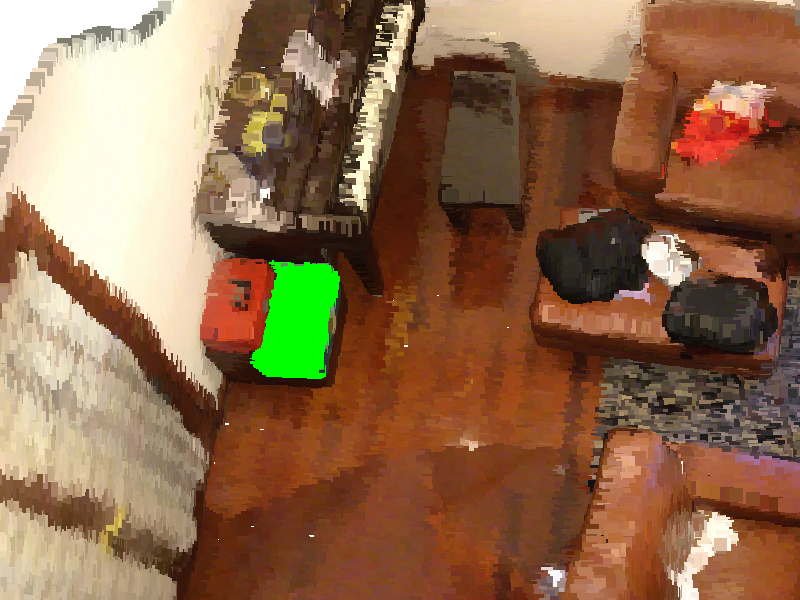}
        \caption*{Input\hspace{2.5cm}GT\hspace{2.5cm}Ours}
        \caption*{this is a green tool box. the green tool box is in front of a red tool box on the floor next to a piano.}
    \end{subfigure}
    \hfill
    \begin{subfigure}[t]{0.49\textwidth}
        \includegraphics[width=0.32\linewidth]{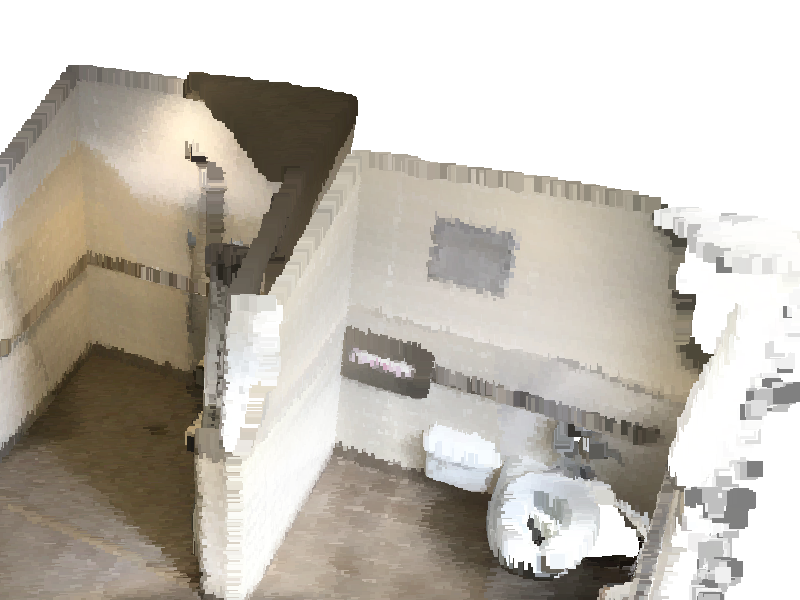}
        \hfill
        \includegraphics[width=0.32\linewidth]{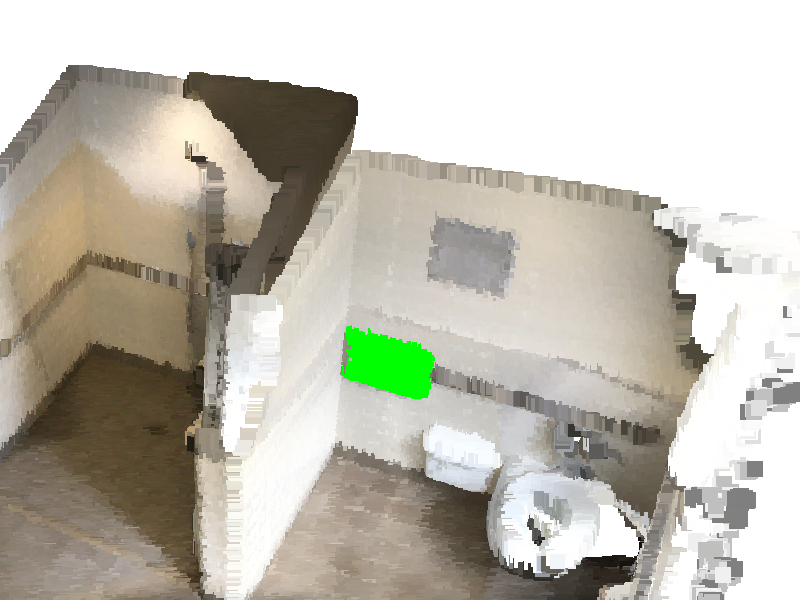}
        \hfill
        \includegraphics[width=0.32\linewidth]{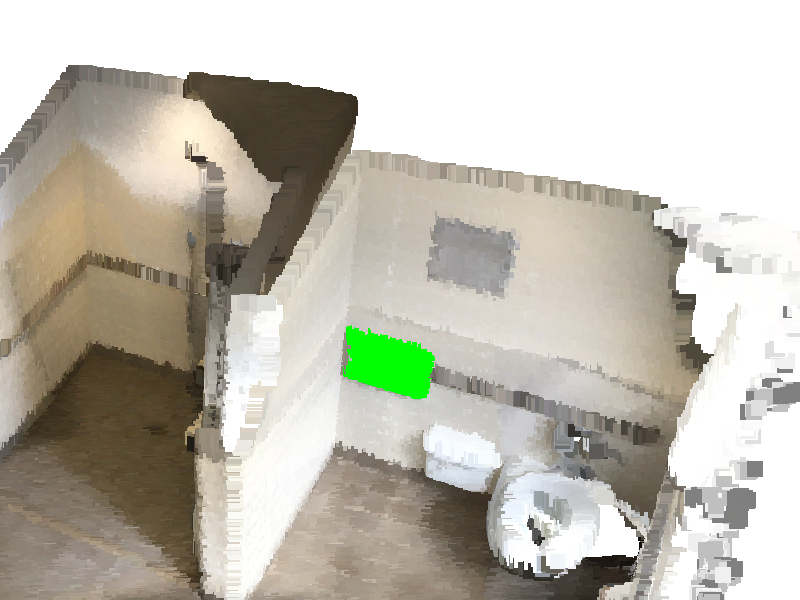}
        \caption*{Input\hspace{2.5cm}GT\hspace{2.5cm}Ours}
        \caption*{this is a rectangular toilet seat cover dispenser. it is to the left of a silver bar on the wall.}
    \end{subfigure}
    \hfill
    \vline
    \hfill
    \begin{subfigure}[t]{0.49\textwidth}
        \includegraphics[width=0.32\linewidth]{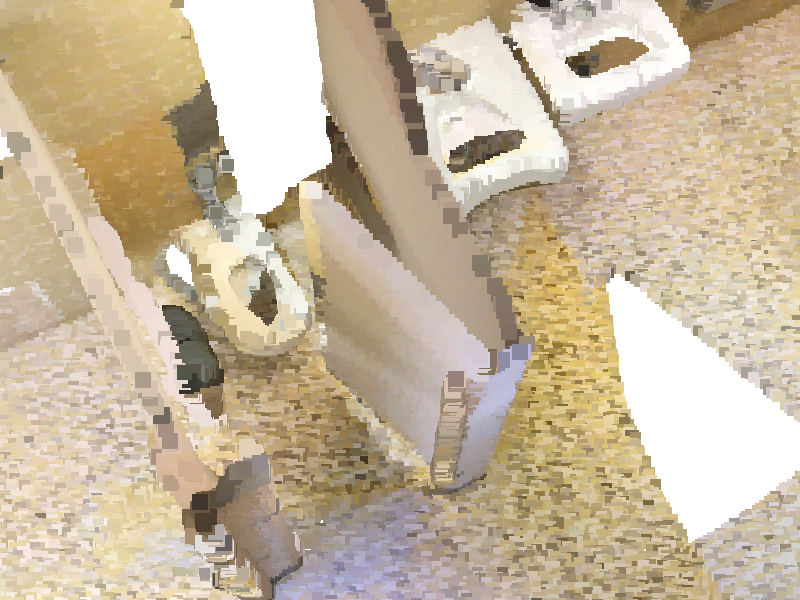}
        \hfill
        \includegraphics[width=0.32\linewidth]{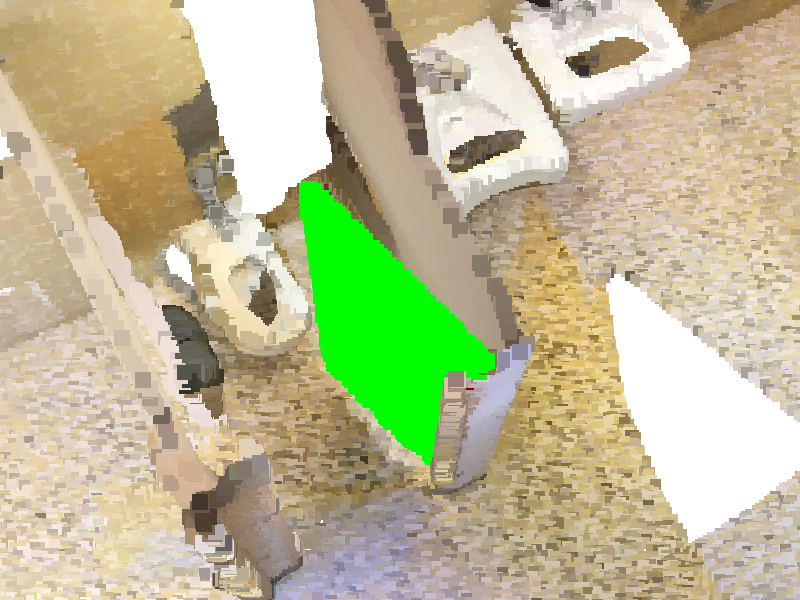}
        \hfill
        \includegraphics[width=0.32\linewidth]{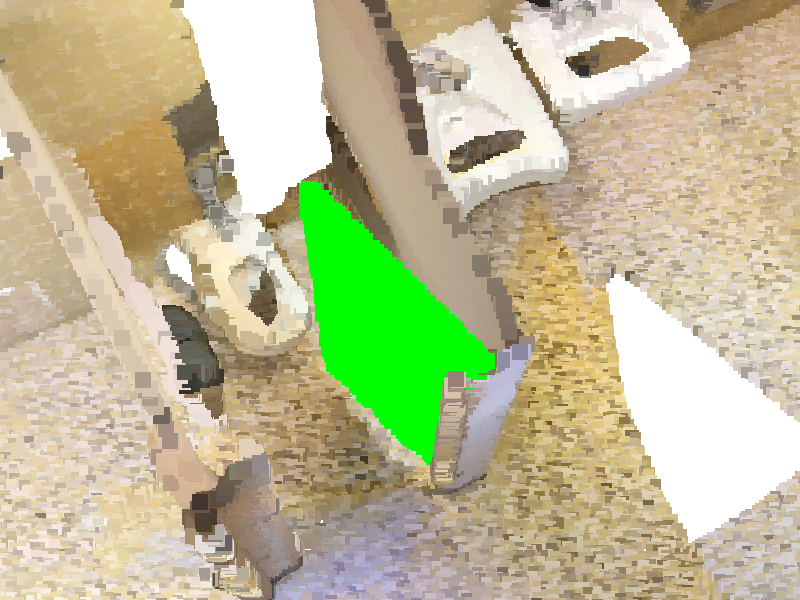}
        \caption*{Input\hspace{2.5cm}GT\hspace{2.5cm}Ours}
        \caption*{a bathrom stall door sits to the left of two sinks. just behind it is a toilet with a toilet-paper dispenser to its left, hangin on the wall.}
    \end{subfigure}
    \caption{Results of grounded segmentation on ScanRefer dataset.}
    \label{supp_refer}
    \hfill
\end{figure*}

Figure~\ref{supp_refer} presents additional grounded segmentation results obtained using the ScanRefer dataset. As illustrated, our Uni3DL model accurately predicts the grounded masks corresponding to each referring sentence.

\subsection{Text-3D cross-modal retrieval}\label{supp_fig_ret}
\begin{figure*}[!ht]
    \rule{\linewidth}{0.4pt}
    \caption*{\hspace{250pt}top1\hspace{1.2cm}top2\hspace{1.2cm}top3\hspace{1.2cm}top4\hspace{1.2cm}top5}
    \rule{\linewidth}{0.4pt}
    \begin{minipage}[t]{0.49\textwidth}
        \vspace{-1cm}
        a round table with differnt type of look and is good 
    \end{minipage}
    \hfill
    \vline 
    \begin{subfigure}[t]{0.09\textwidth}
        \includegraphics[width=\linewidth]{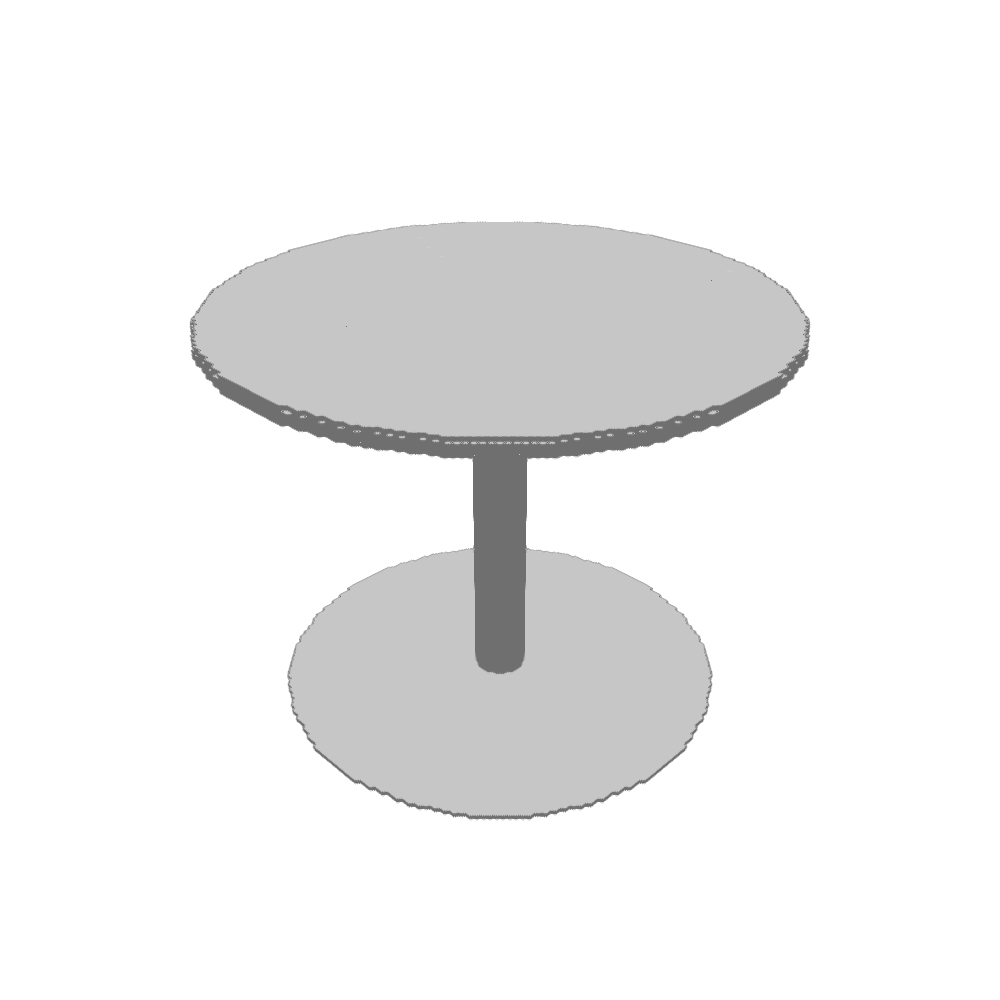}
        \caption*{\textcolor{red}{\textbf{0.91(GT)}}}
    \end{subfigure}
    \hfill    
    \begin{subfigure}[t]{0.09\textwidth}
        \includegraphics[width=\linewidth]{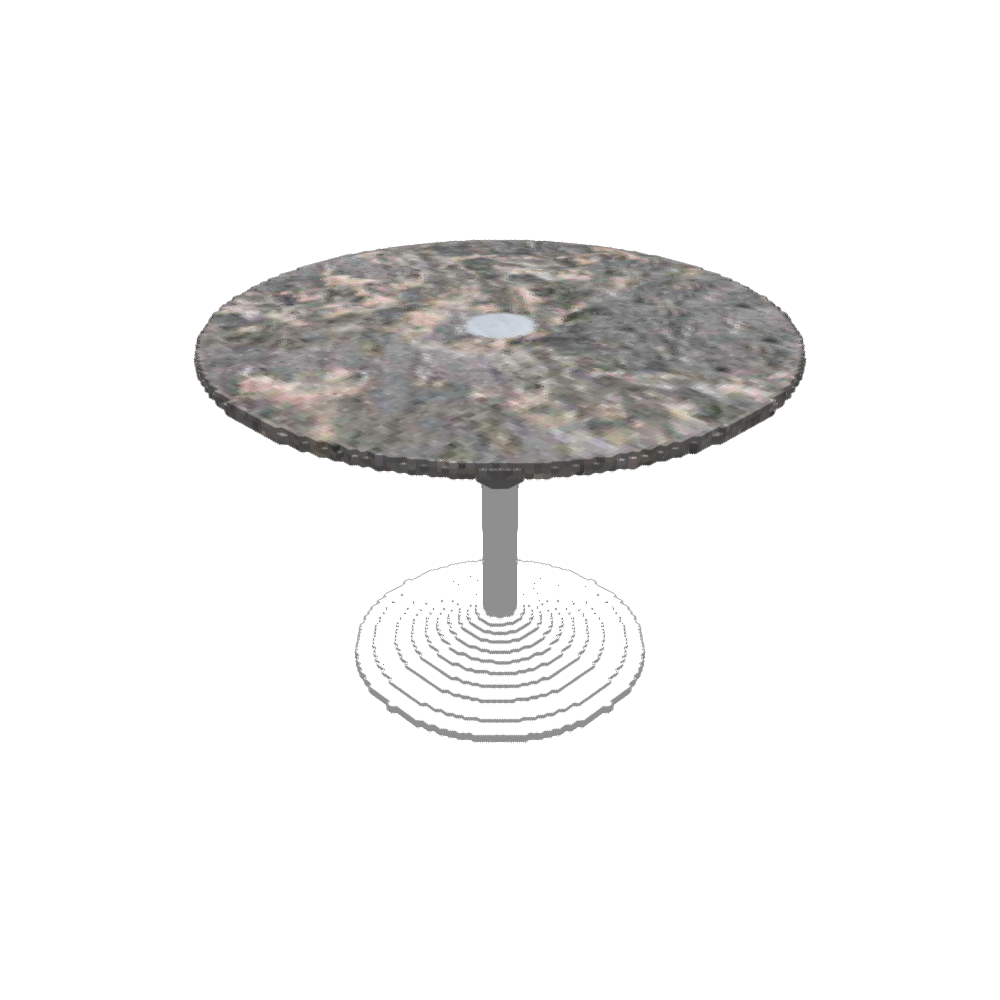}
        \caption*{0.90}
    \end{subfigure}
    \hfill    
    \begin{subfigure}[t]{0.09\textwidth}
        \includegraphics[width=\linewidth]{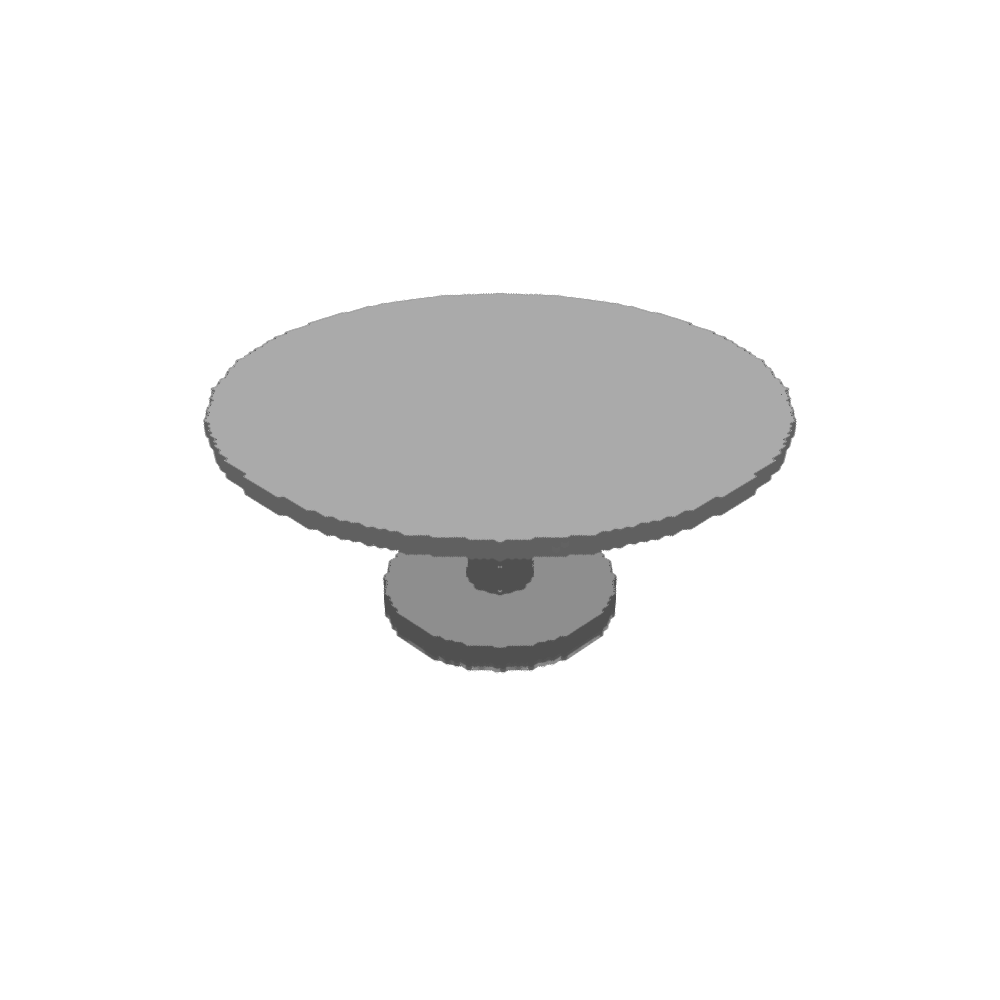}
        \caption*{0.90}
    \end{subfigure}
    \hfill    
    \begin{subfigure}[t]{0.09\textwidth}
        \includegraphics[width=\linewidth]{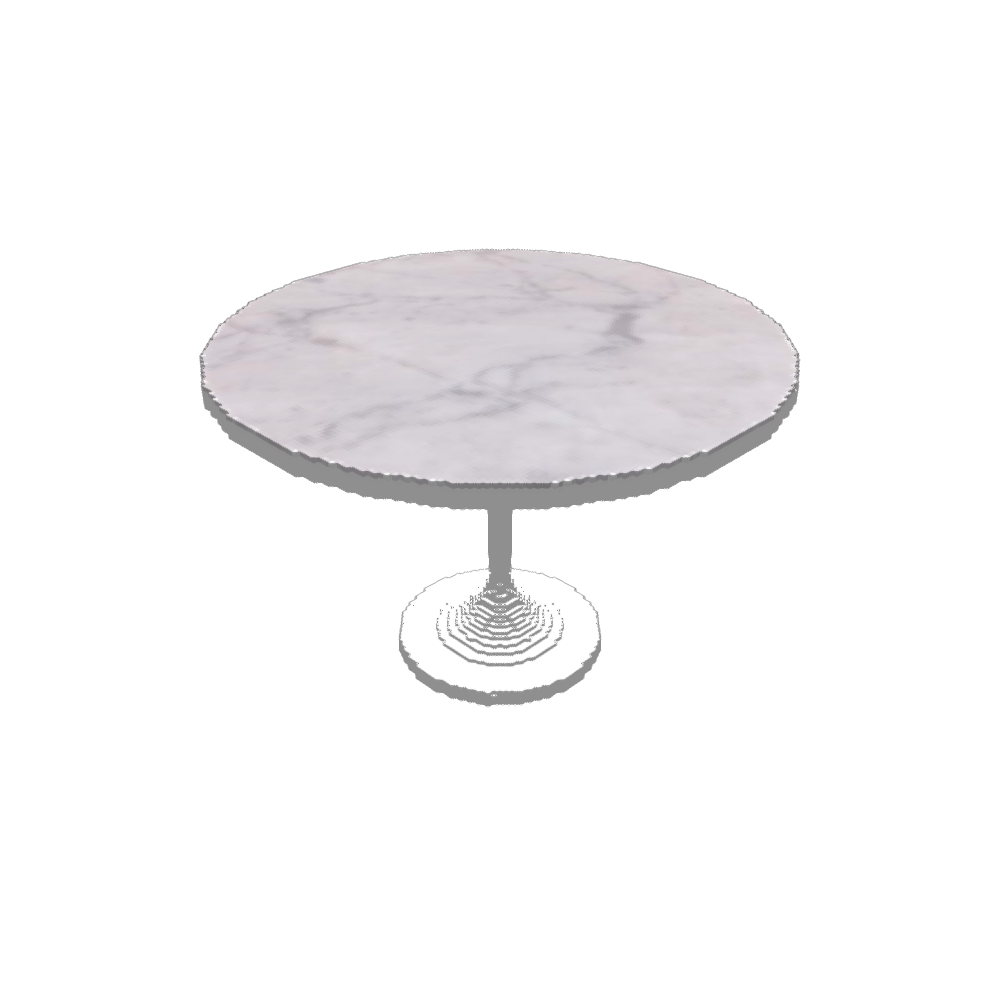}
        \caption*{0.90}
    \end{subfigure}
    \hfill    
    \begin{subfigure}[t]{0.09\textwidth}
        \includegraphics[width=\linewidth]{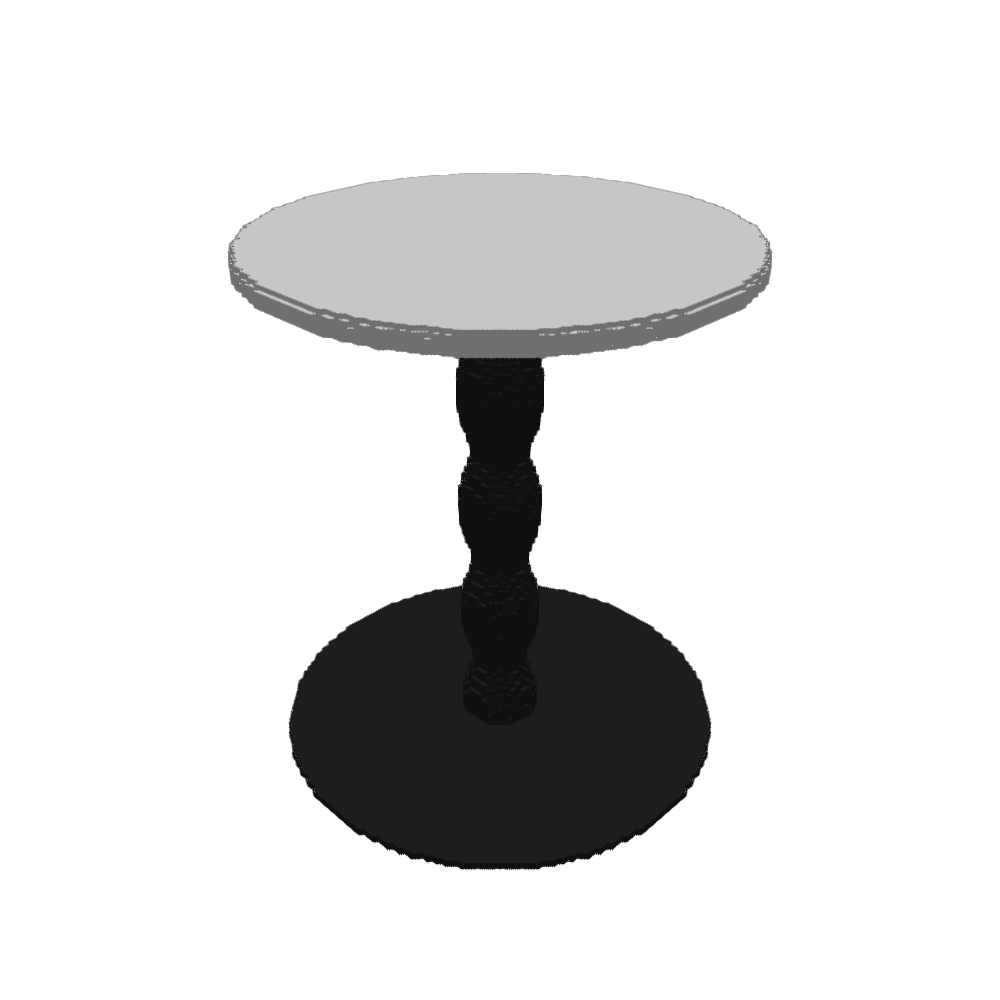}
        \caption*{0.88}
    \end{subfigure}
    \hfill
    
    \rule{\linewidth}{0.4pt}
    
    \begin{minipage}[t]{0.49\textwidth}
        \vspace{-1cm}
        it is an oblong table with distressed wooden top and six spindle shaped legs. 
    \end{minipage}
    \hfill
    \vline
    \begin{subfigure}[t]{0.09\textwidth}
        \includegraphics[width=\linewidth]{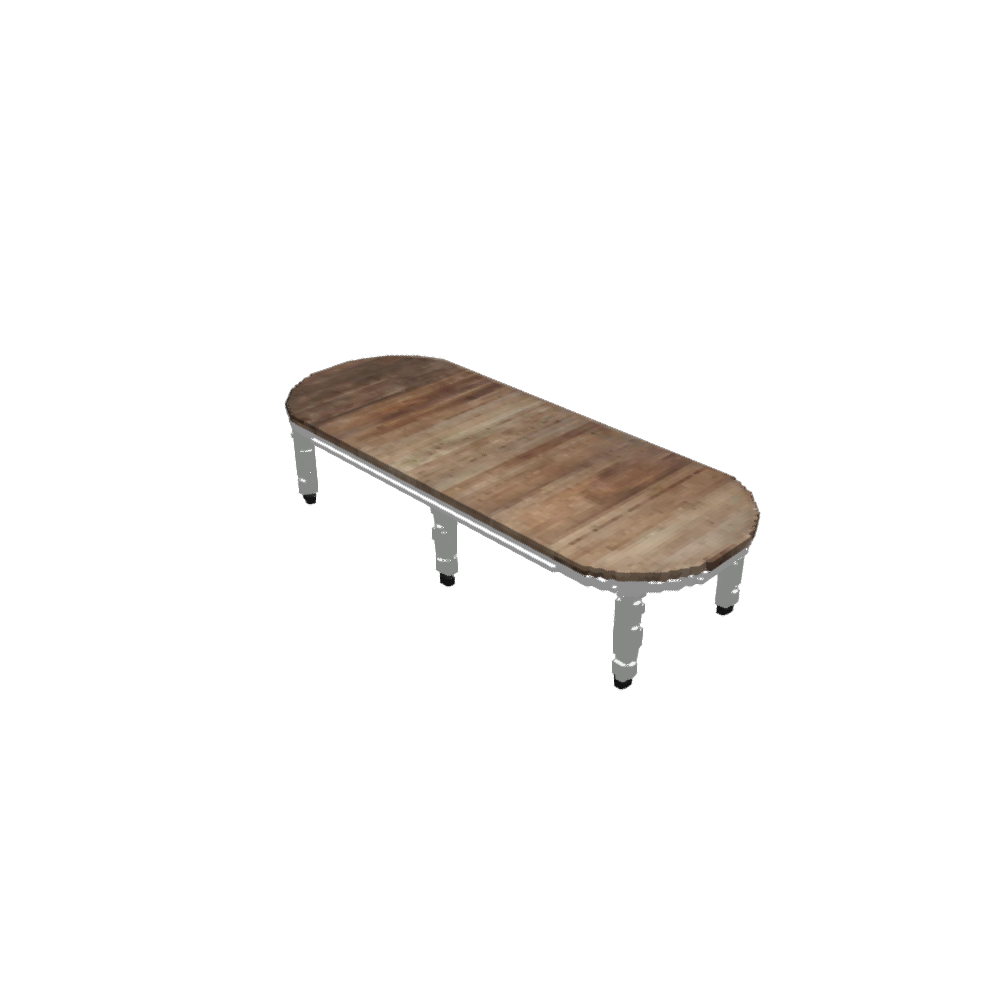}
        \caption*{\textcolor{red}{\textbf{0.91(GT)}}}
    \end{subfigure}
    \hfill    
    \begin{subfigure}[t]{0.09\textwidth}
        \includegraphics[width=\linewidth]{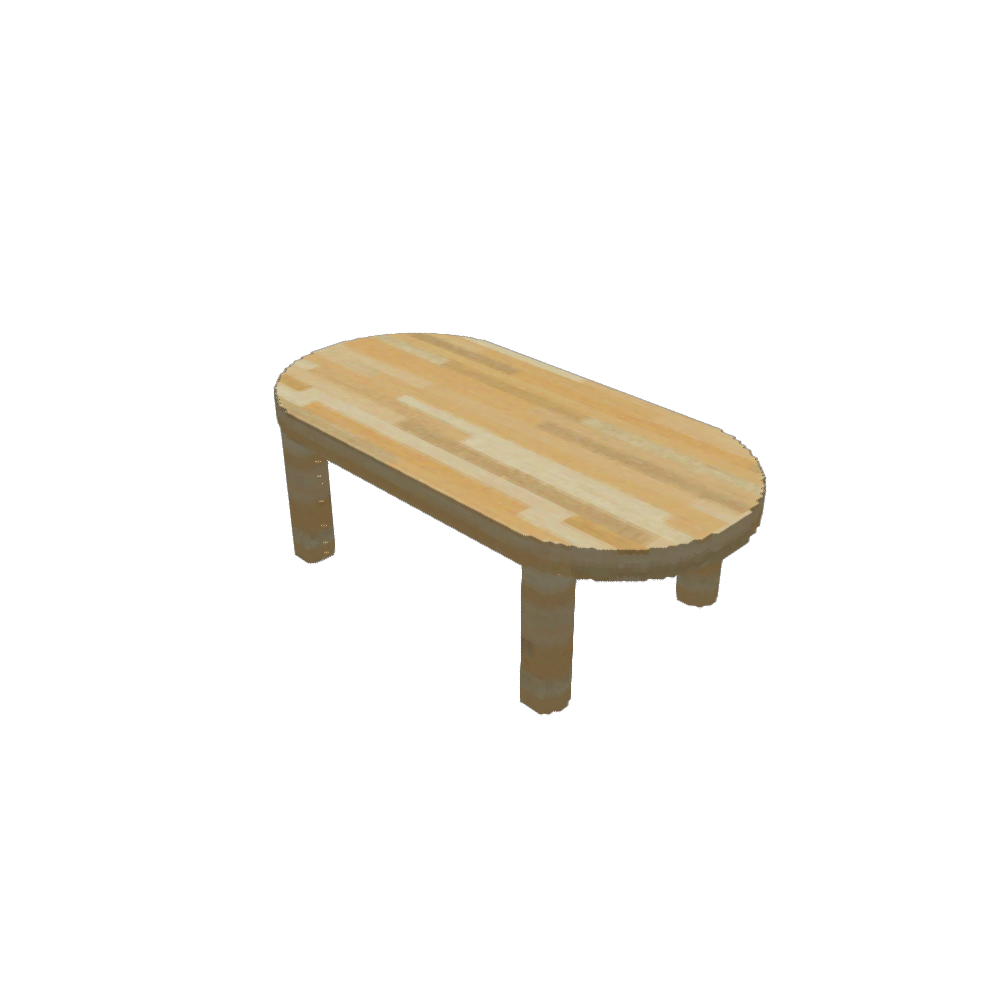}
        \caption*{0.86}
    \end{subfigure}
    \hfill    
    \begin{subfigure}[t]{0.09\textwidth}
        \includegraphics[width=\linewidth]{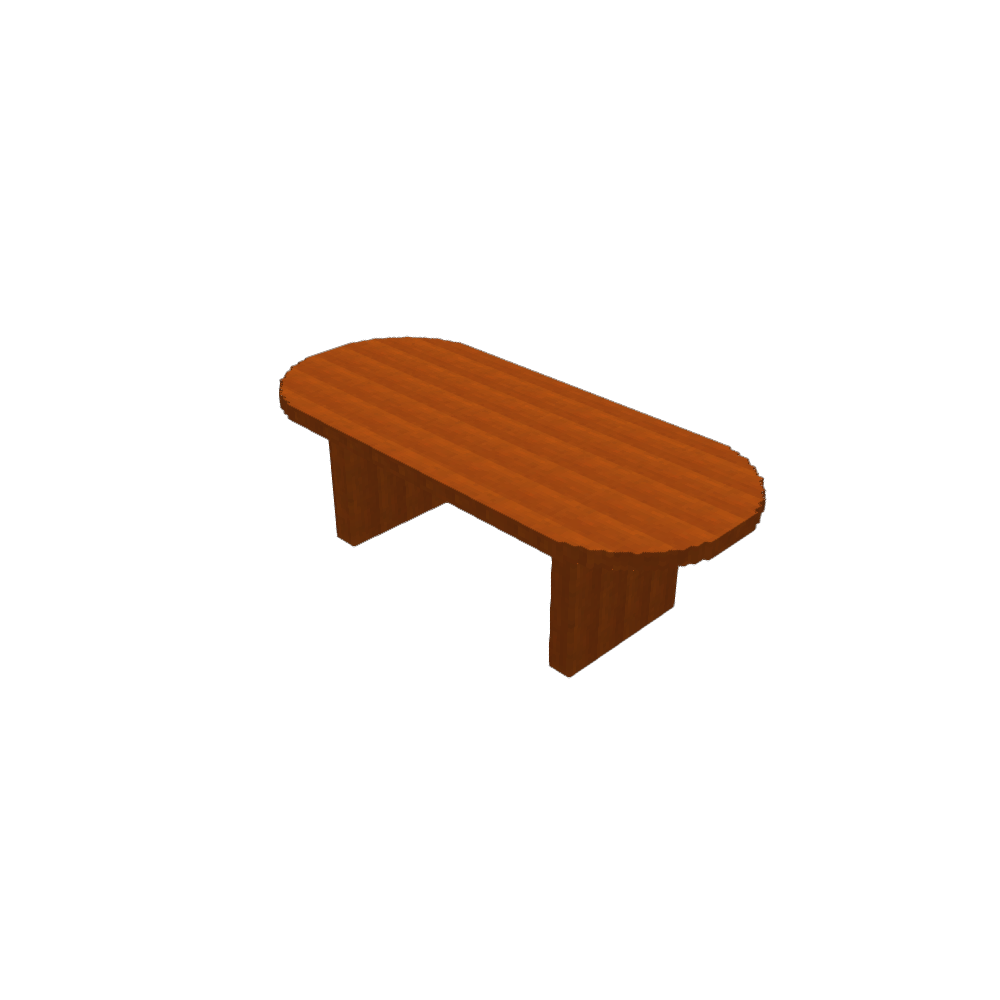}
        \caption*{0.86}
    \end{subfigure}
    \hfill    
    \begin{subfigure}[t]{0.09\textwidth}
        \includegraphics[width=\linewidth]{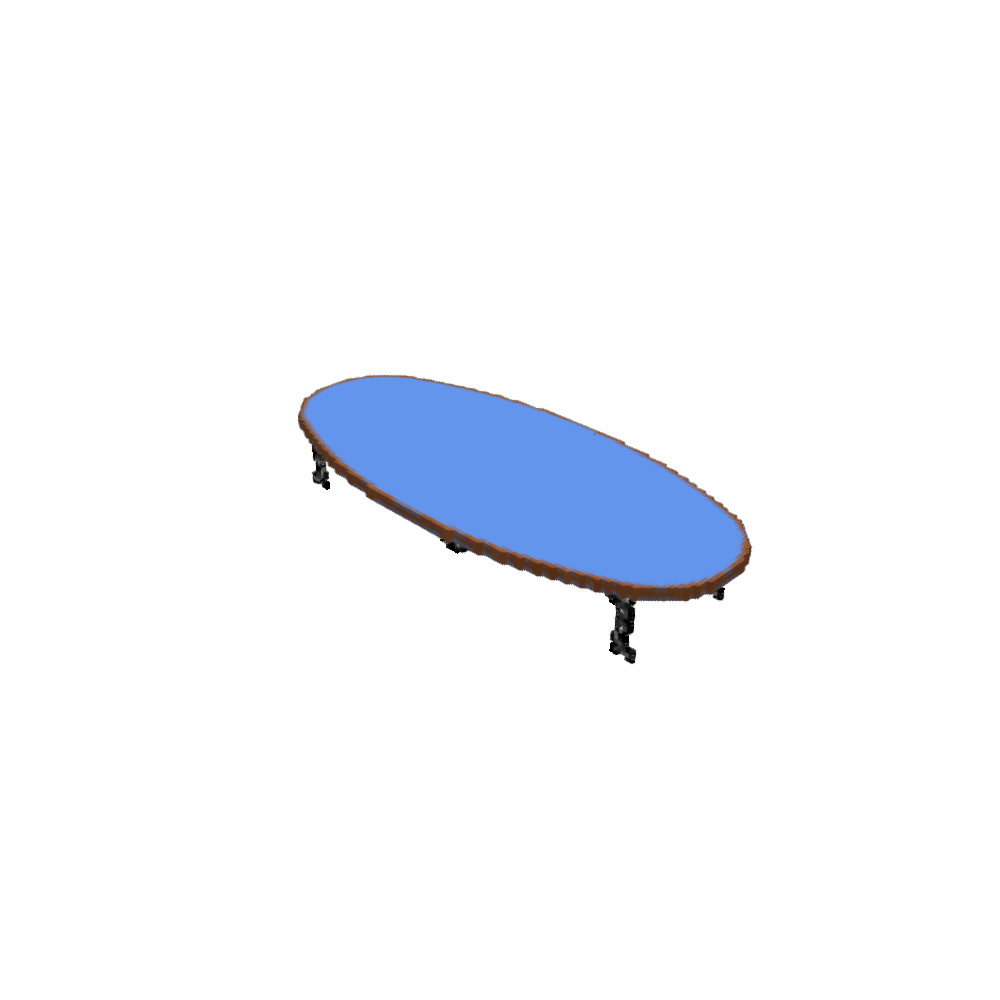}
        \caption*{0.83}
    \end{subfigure}
    \hfill    
    \begin{subfigure}[t]{0.09\textwidth}
        \includegraphics[width=\linewidth]{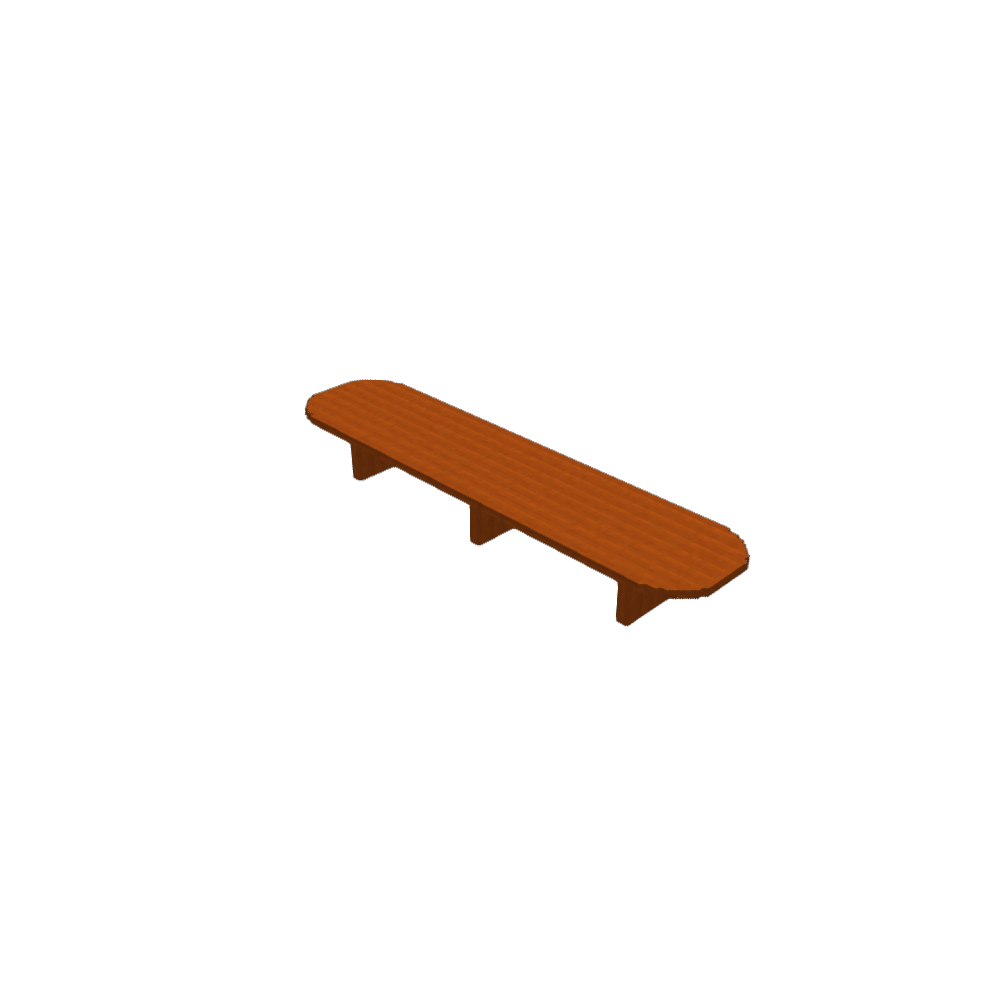}
        \caption*{0.83}
    \end{subfigure}
    
    \rule{\linewidth}{0.4pt}
    
    \begin{minipage}[t]{0.49\textwidth}
        \vspace{-1cm}
        a red sofa that is sitting on a black carpet.  the sofa is round and ovalular. 
    \end{minipage}
    \hfill
    \vline
    \begin{subfigure}[t]{0.09\textwidth}
        \includegraphics[width=\linewidth]{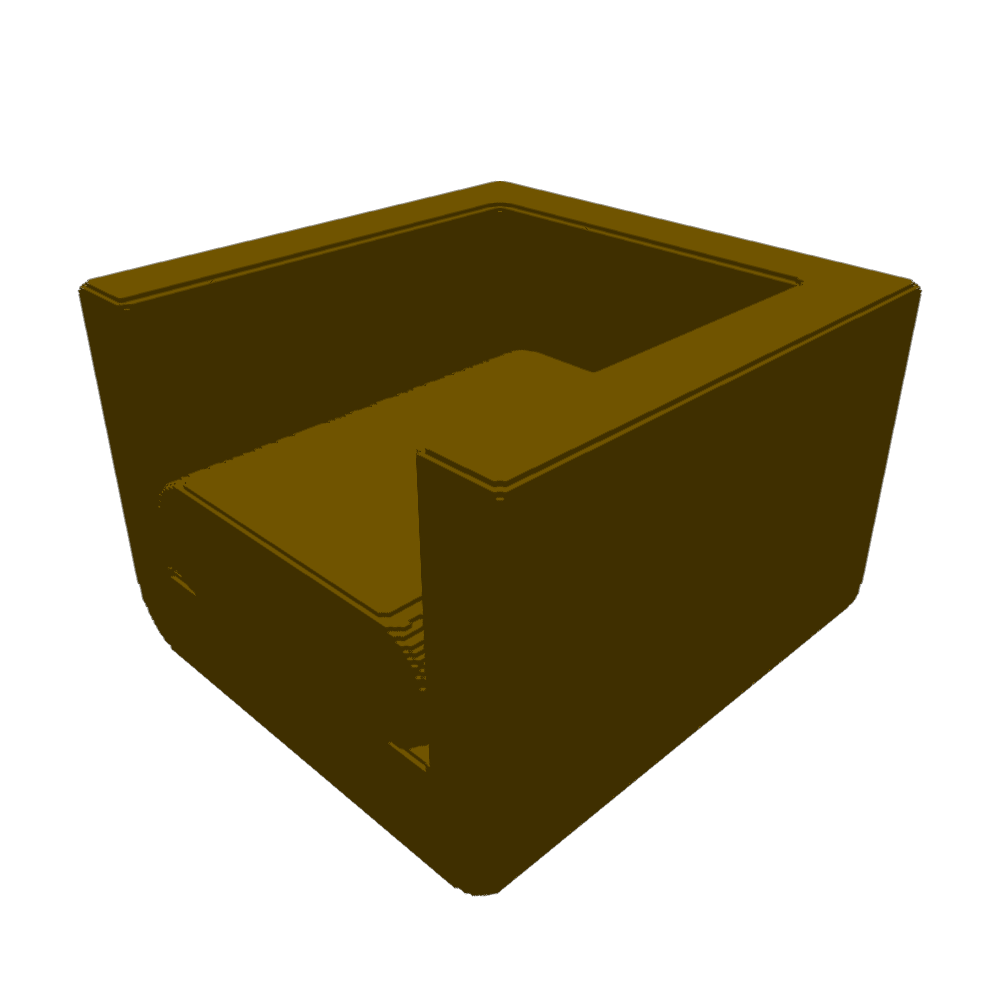}
        \caption*{0.90}
    \end{subfigure}
    \hfill    
    \begin{subfigure}[t]{0.09\textwidth}
        \includegraphics[width=\linewidth]{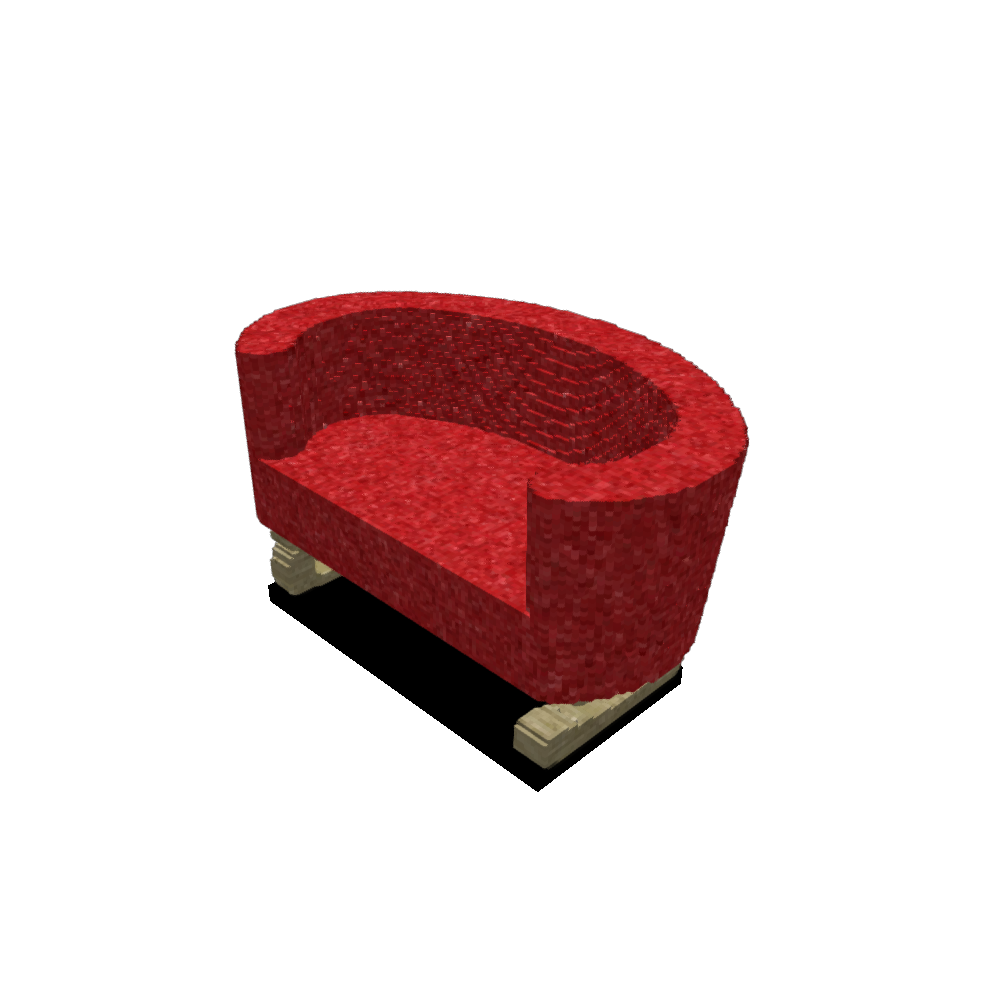}
        \caption*{\textcolor{red}{\textbf{0.90(GT)}}}
    \end{subfigure}
    \hfill    
    \begin{subfigure}[t]{0.09\textwidth}
        \includegraphics[width=\linewidth]{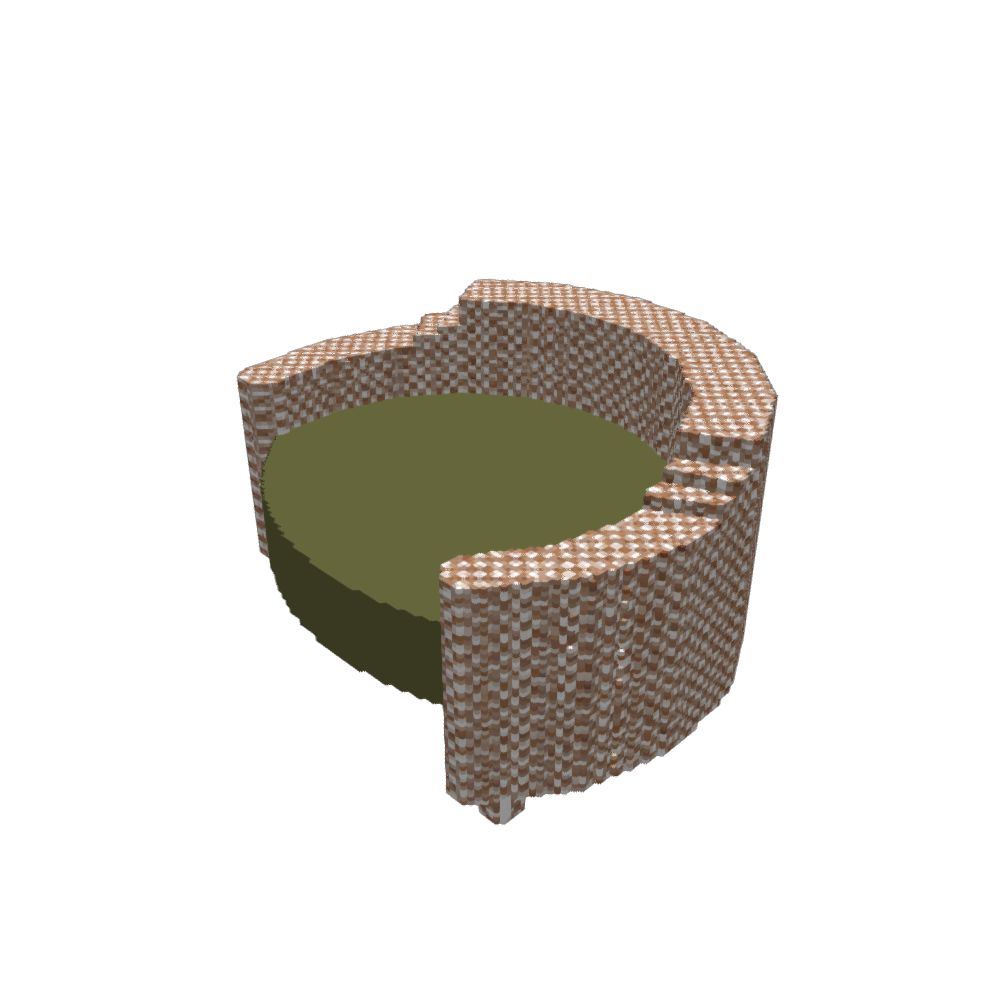}
        \caption*{0.87}
    \end{subfigure}
    \hfill    
    \begin{subfigure}[t]{0.09\textwidth}
        \includegraphics[width=\linewidth]{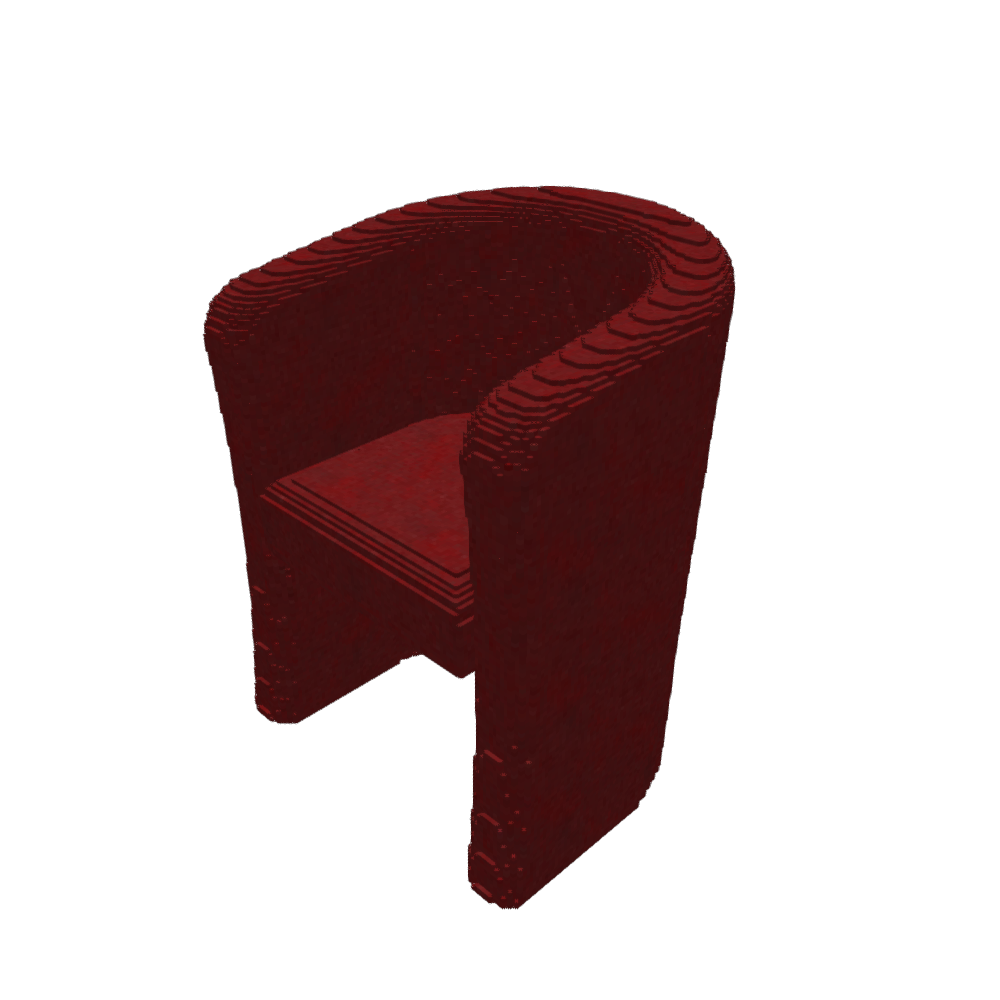}
        \caption*{0.86}
    \end{subfigure}
    \hfill    
    \begin{subfigure}[t]{0.09\textwidth}
        \includegraphics[width=\linewidth]{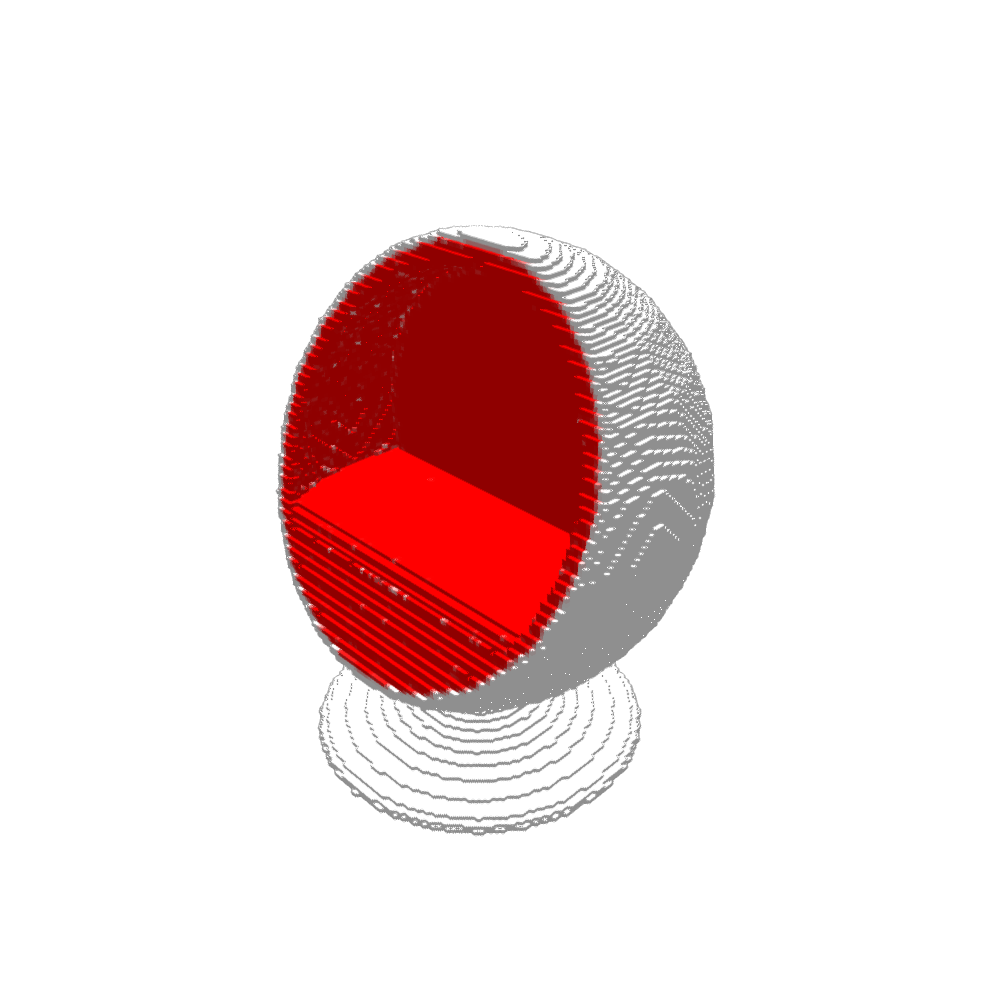}
        \caption*{0.82}
    \end{subfigure}
    
    \rule{\linewidth}{0.4pt}
    
    \begin{minipage}[t]{0.49\textwidth}
        \vspace{-1cm}
        a unique design brown wooden table with white color at top is great for outdoor 
    \end{minipage}
    \hfill
    \vline
    \begin{subfigure}[t]{0.09\textwidth}
        \includegraphics[width=\linewidth]{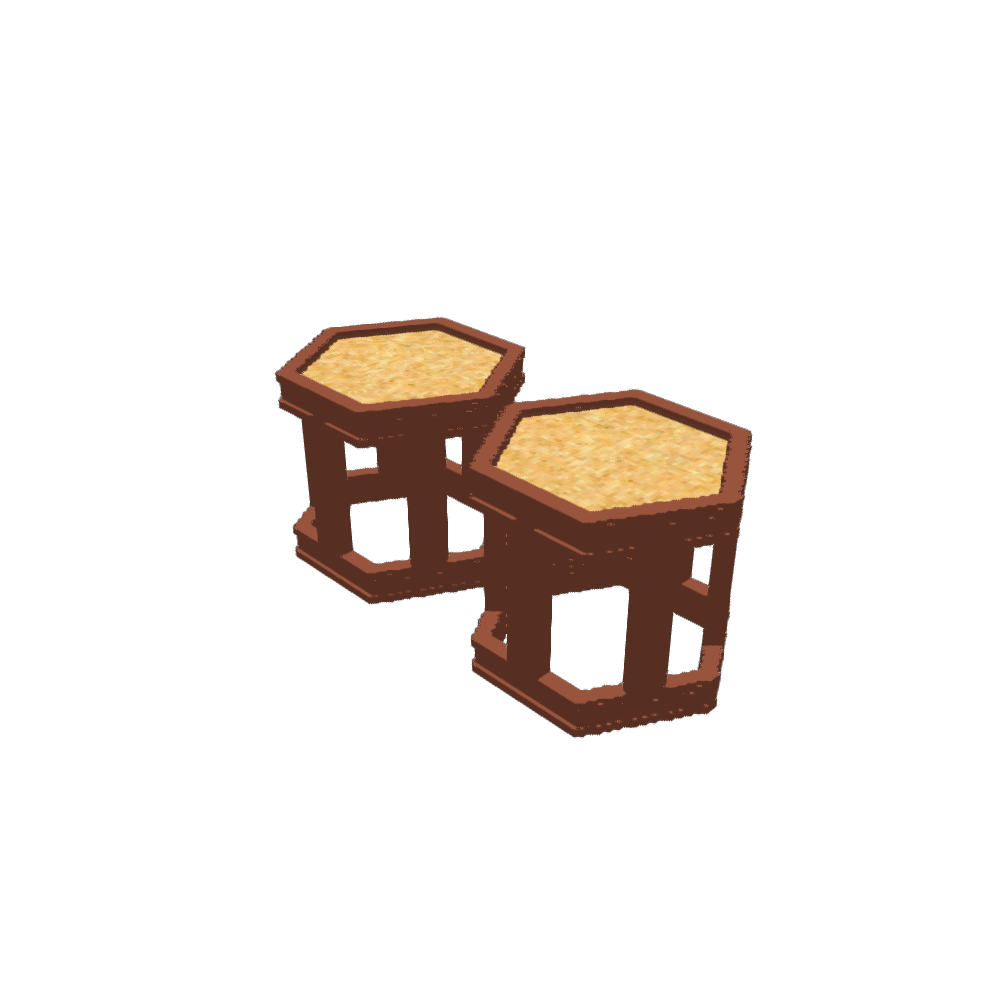}
        \caption*{0.94}
    \end{subfigure}
    \hfill    
    \begin{subfigure}[t]{0.09\textwidth}
        \includegraphics[width=\linewidth]{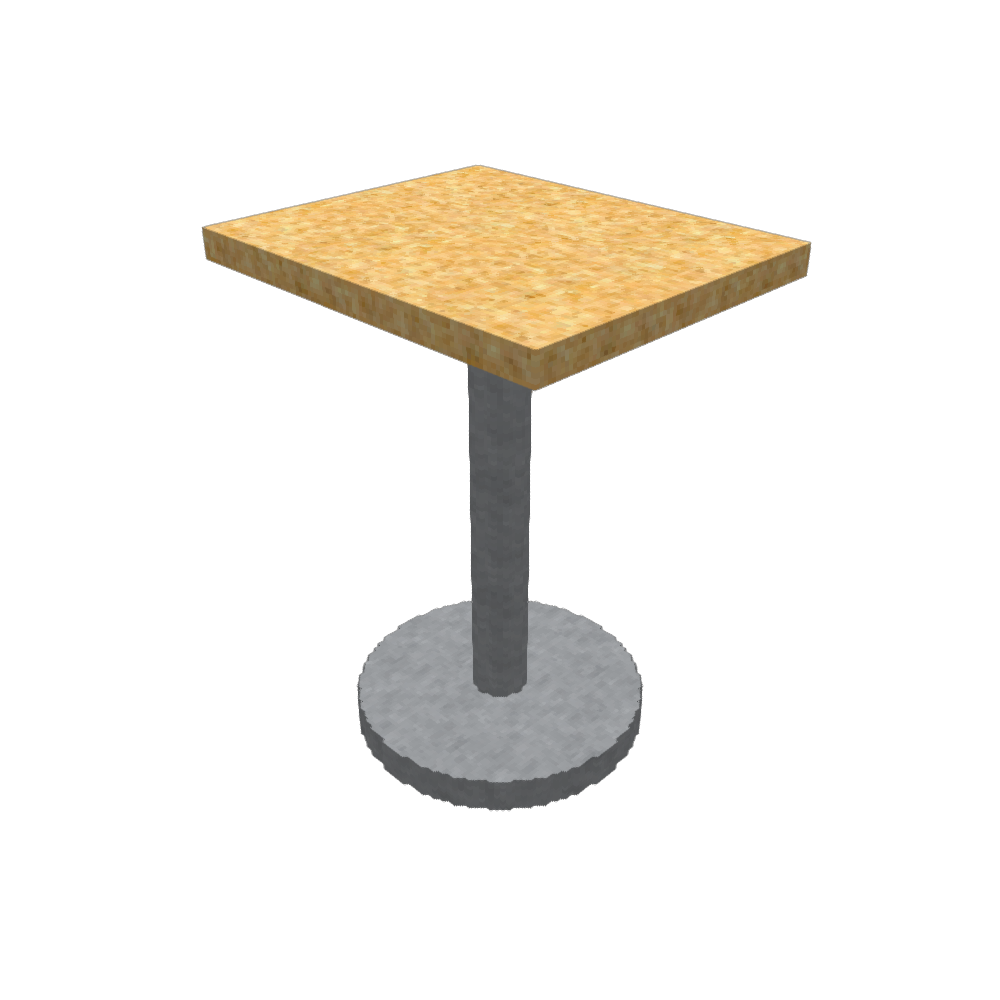}
        \caption*{0.87}
    \end{subfigure}
    \hfill    
    \begin{subfigure}[t]{0.09\textwidth}
        \includegraphics[width=\linewidth]{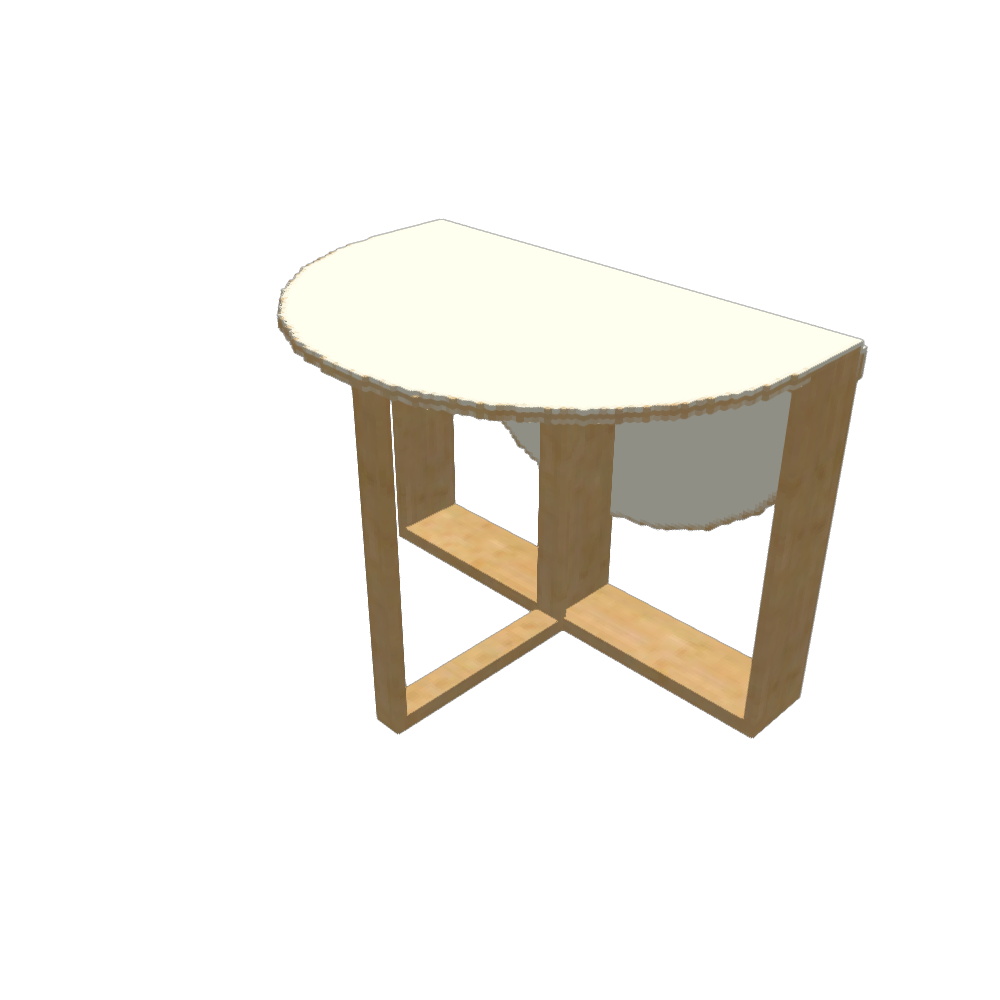}
        \caption*{\textcolor{red}{\textbf{0.83(GT)}}}
    \end{subfigure}
    \hfill    
    \begin{subfigure}[t]{0.09\textwidth}
        \includegraphics[width=\linewidth]{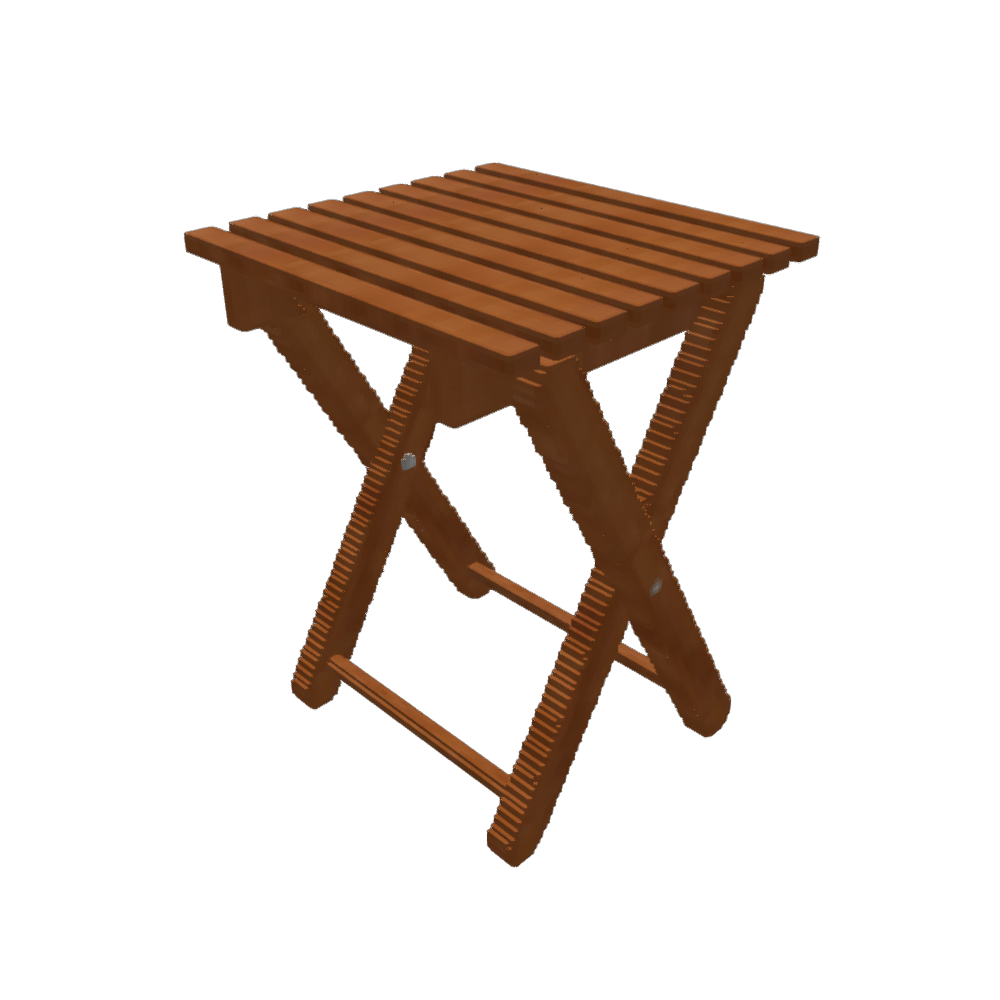}
        \caption*{0.82}
    \end{subfigure}
    \hfill    
    \begin{subfigure}[t]{0.09\textwidth}
        \includegraphics[width=\linewidth]{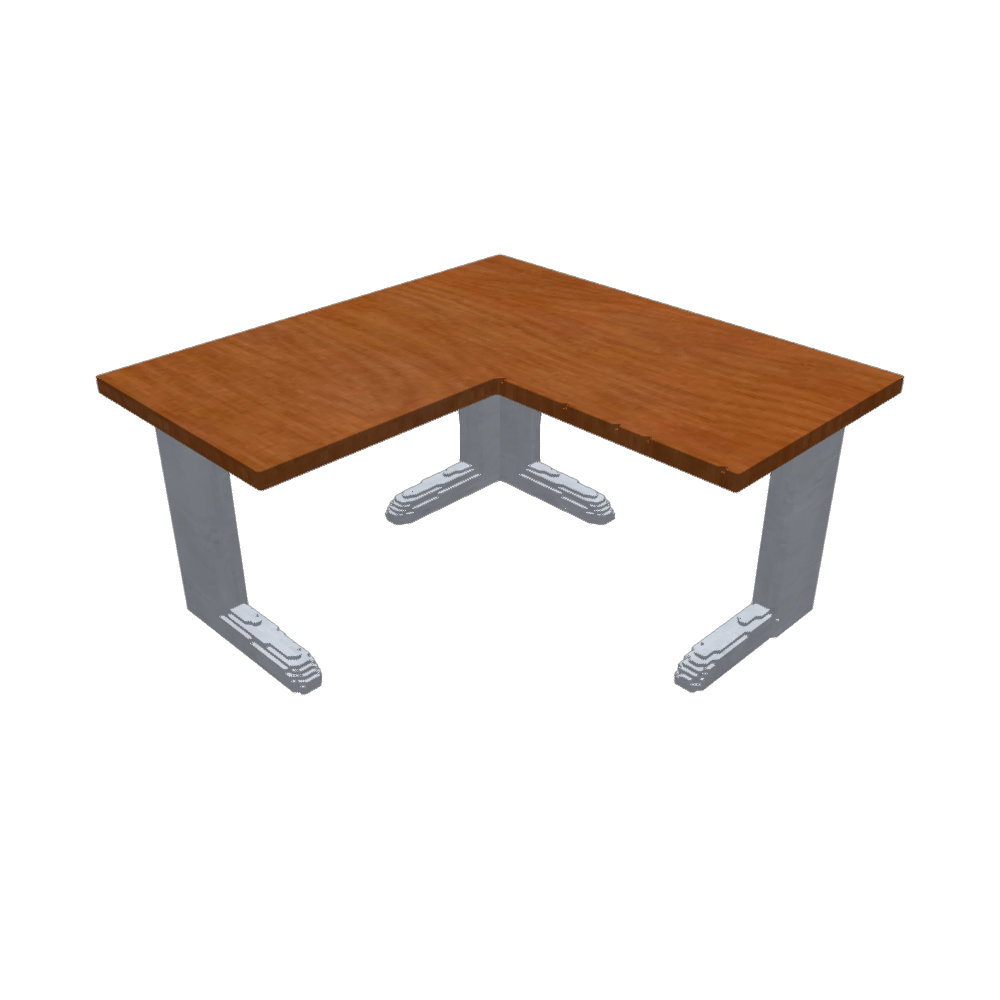}
        \caption*{0.81}
    \end{subfigure}
    \caption{Text-to-Shape Retrieval results on Text2Shape dataset, For each query sentence, we show the top-5 ranked shape, the scores of ground truth shape are marked in \textcolor{red}{red}.}
    \label{fig_text_to_3d}
\end{figure*}

\begin{figure*}[!ht]
    \rule{\linewidth}{0.4pt}
    \caption*{\textcolor{white}{.}\hspace{20pt}Query Shape\hspace{40pt} Retrieval Results\hspace{500pt}.}
    \rule{\linewidth}{0.4pt}
    \begin{subfigure}[t]{0.19\textwidth}
        \includegraphics[width=\linewidth]{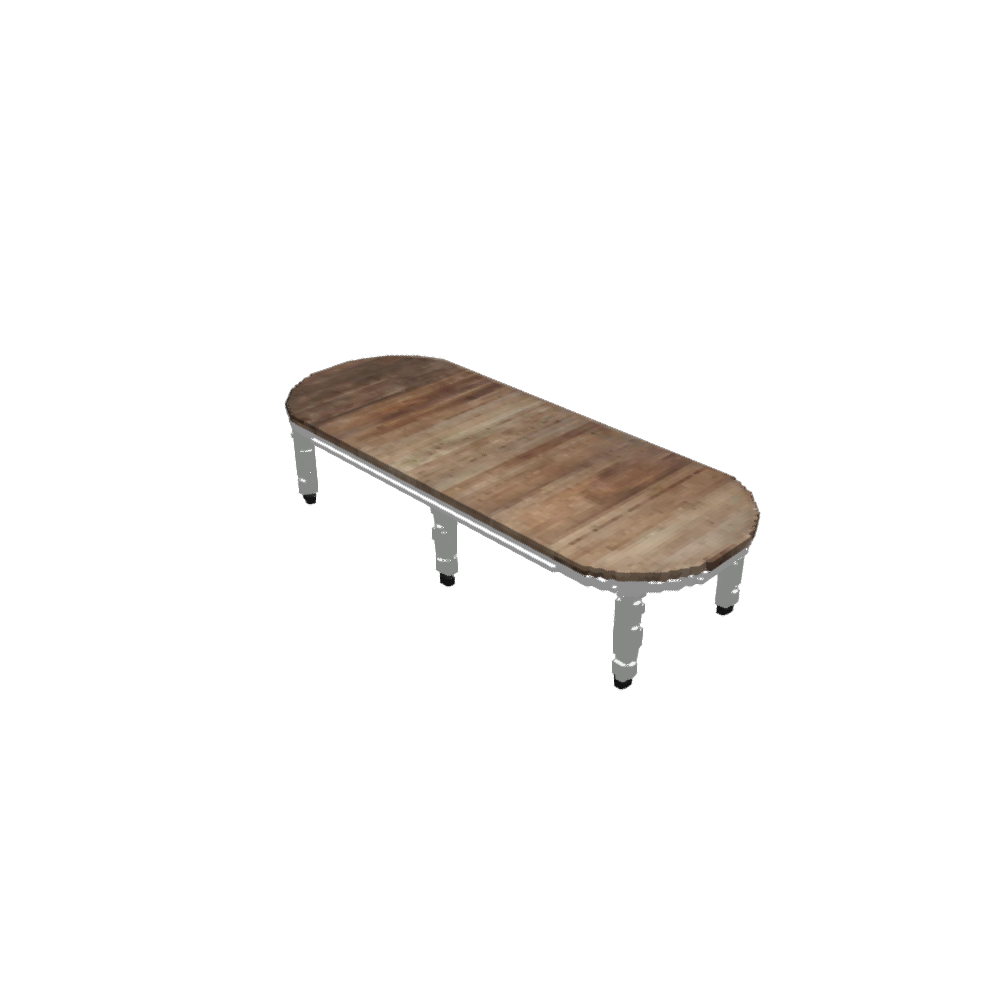}
    \end{subfigure}
    \hfill    
    \vline 
    \hfill    
    \begin{minipage}[t]{0.79\textwidth}
        \vspace{-90pt}        
        \begin{enumerate}
            \item[\textcolor{red}{\textbf{1.}}] \textcolor{red}{\textbf{ it is an oblong table with distressed wooden top and six spindle shaped legs. (Prob: 0.91, GT)}}
            \item[{2.}] elliptical table with brown wooden top and grey straight legs (Prob: 0.88)
            \item[{3.}] a brown oblong wooden topped table with four grey supporting legs (Prob: 0.86)
            \item[{4.}] oval table with shape oval , 4 legs and high qualit wood from alaska that will make you happy (Prob: 0.86)
            \item[{5.}] this is a dining table that is oval with the insert, but could collapse down to a circle table. it has 4 legs. (Prob: 0.85)
        \end{enumerate}
    \end{minipage}
    
    \rule{\linewidth}{0.4pt}
    
    \begin{subfigure}[t]{0.19\textwidth}
        \includegraphics[width=\linewidth]{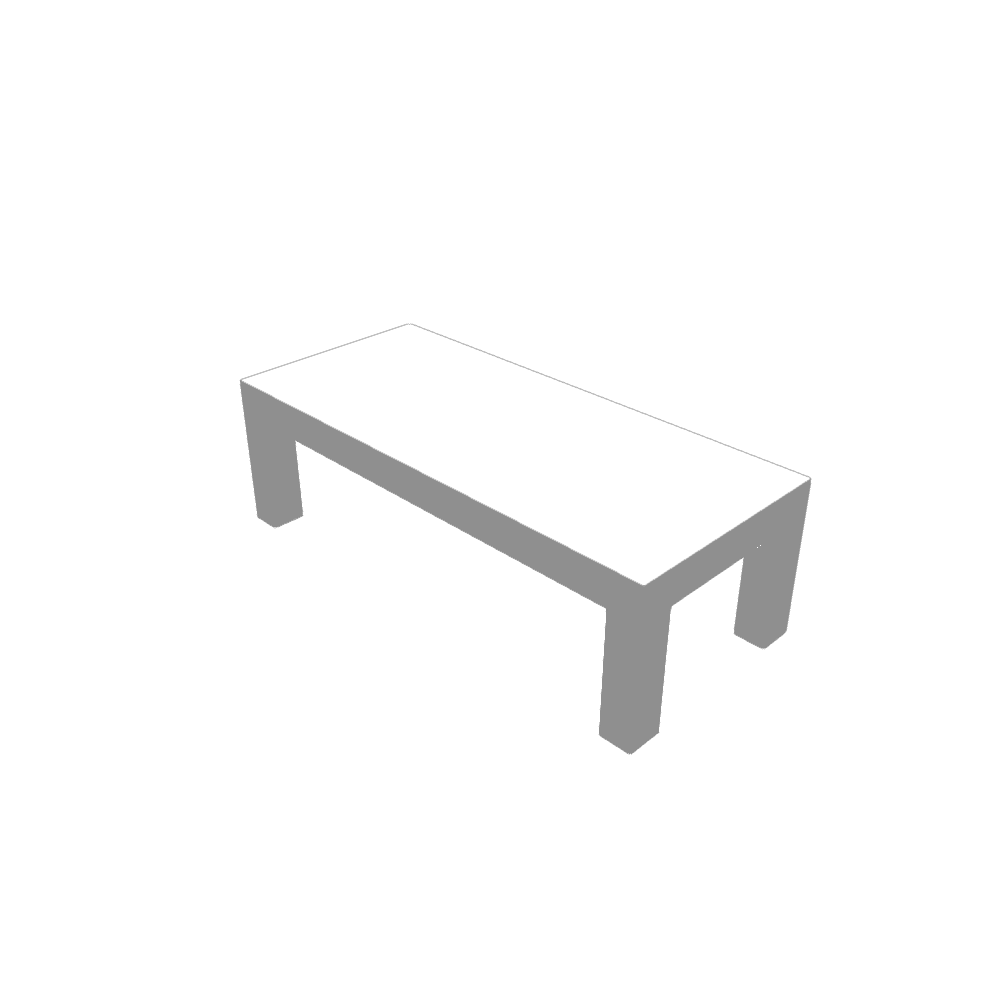}
    \end{subfigure}
    \hfill    
    \vline
    \hfill    
    \begin{minipage}[t]{0.79\textwidth}
        \vspace{-90pt}        
        \begin{enumerate}
            \item a grey rectangular shaped wooden table with four short legs. (Prob: 0.91)
            \textcolor{red}{\textbf{\item grey colored, wooden table. four short solid legs with rectangular top. (Prob: 0.89, GT)}}
            \item a grey rectangular short table with four short grey legs. (Prob: 0.89)
            \item a white colored rectangular table which has rectangular top painted in white and has four short legs colored in black. (Prob: 0.89)
            \textcolor{red}{\textbf{\item a low and long grey table with four legs. (Prob: 0.89, GT)}}
        \end{enumerate}
    \end{minipage}
    \hfill
    
    \rule{\linewidth}{0.4pt}
    
    \begin{subfigure}[t]{0.19\textwidth}
        \includegraphics[width=\linewidth]{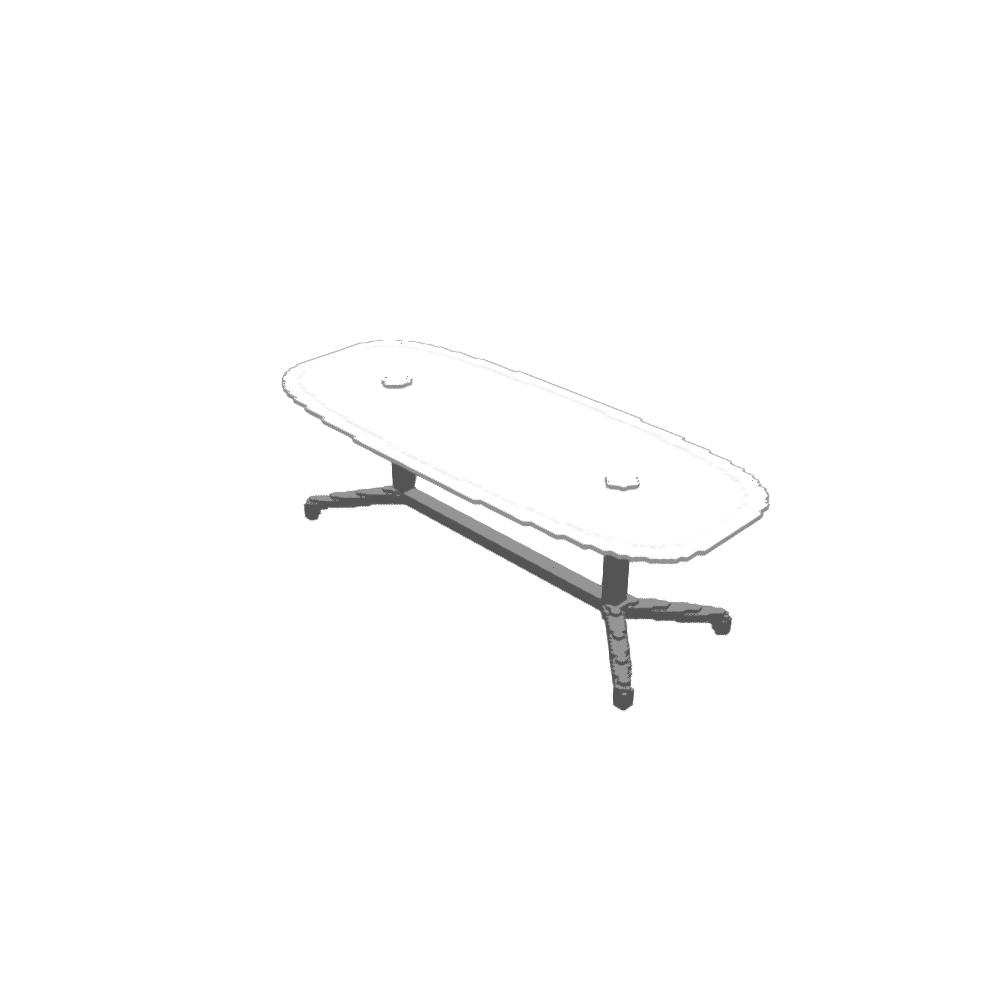}
    \end{subfigure}
    \hfill    
    \vline
    \hfill    
    \begin{minipage}[t]{0.79\textwidth}
        \vspace{-90pt}        
        \begin{enumerate}
            \item[\textcolor{red}{\textbf{1.}}] \textcolor{red}{\textbf{a white conference table with legs (Prob: 0.92, GT)}}
            \item[{2.}] a table with a white colored oval type top and four grey colored plate type legs (Prob: 0.88)
            \item[{3.}] simple white table. lunch room table. 4 legs. metal legs. formica top. wide. (Prob: 0.87)
            \item[{4.}] an ash colored oval shaped steel coffee table which has skinny rectangular shaped long four legs. (Prob: 0.87)
            \item[{5.}] outdoor table, wooden, gray, oval shape, with four legs. (Prob: 0.86)
        \end{enumerate}
    \end{minipage}
    \hfill
    
    \rule{\linewidth}{0.4pt}
    
    \begin{subfigure}[t]{0.19\textwidth}
        \includegraphics[width=\linewidth]{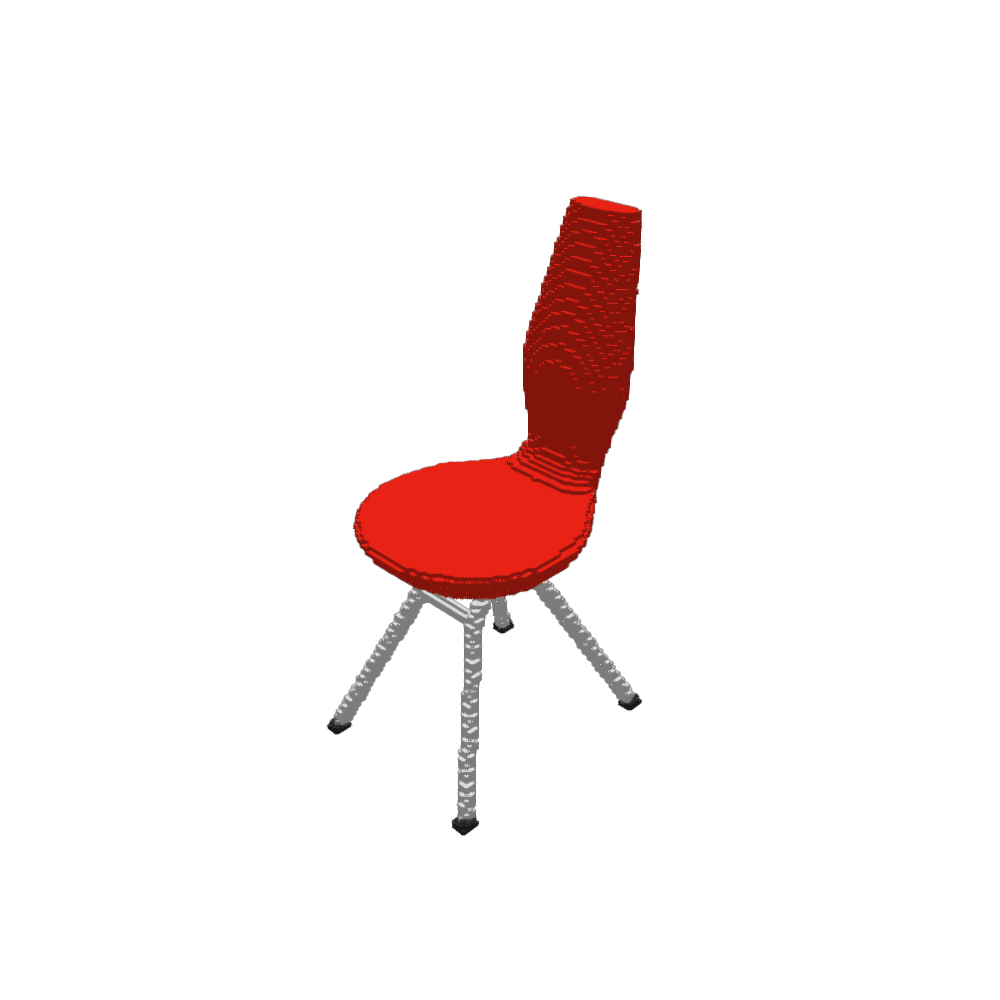}
    \end{subfigure}
    \hfill    
    \vline
    \hfill    
    \begin{minipage}[t]{0.79\textwidth}
        \vspace{-90pt}        
        \begin{enumerate}
            \item[\textcolor{red}{\textbf{1.}}] \textcolor{red}{\textbf{red colour plastic chair with u shape iron legs and chair was looking variety (Prob: 0.90, GT)}}
            \item[{2.}] a red chair with curved back legs.  probably made of plastic. (Prob: 0.89)
            \item[{3.}] a basket backed, red seated high bar stool with thin metal legs (Prob: 0.88)
            \item[\textcolor{red}{\textbf{4.}}] \textcolor{red}{\textbf{red high back chair made of plastic. four legs are made of metal. (Prob: 0.88, GT)}}
            \item[\textcolor{red}{\textbf{5.}}] \textcolor{red}{\textbf{this is a red molded chair with back and no arms. the chair has 4 metal/plastic legs. (Prob: 0.87, GT)}}
        \end{enumerate}
    \end{minipage}
    \hfill
    \caption{Shape-to-Text Retrieval results on Text2Shape dataset, For each query shape, we show the top-5 ranked sentences, the ground truth sentences are marked in \textcolor{red}{red}.}
    \label{fig_3d_to_text}
\end{figure*}

We show text-to-3D and 3D-to-text retrieval results in Figure~\ref{fig_text_to_3d} and Figure~\ref{fig_3d_to_text} respectively. From the two figures, our Uni3DL model learns satisfying text-3D feature alignments and produces satisfying cross-modal retrieval results. 

\section{Limitation and Future Work}

In this study, we introduced Uni3DL, a novel unified model for understanding both 3D structures and language, operating directly on raw point clouds. This approach marks a departure from conventional 3D vision-language models that predominantly rely on projected multi-view images. While these projection-based methods are limited by their handling of geometric information, their integration with powerful 2D pretrained foundation models, such as CLIP, has yielded promising results.

To leverage the benefits of both point-based and projection-based techniques, our future work will focus on a hybrid approach. This strategy aims to concurrently learn joint 2D and 3D features, integrating insights from 2D foundation models. This advancement is expected to significantly enhance the sophistication and accuracy of 3D language understanding in upcoming versions of our model.

\end{document}